\documentclass[runningheads]{llncs}
\pdfoutput=1
\usepackage{graphicx}

\usepackage{tikz}
\usepackage{comment}
\usepackage{amsmath,amssymb} 
\usepackage{color}
\usepackage{array}
\usepackage{booktabs}

\newcommand\blfootnote[1]{%
\begingroup
\renewcommand\thefootnote{}\footnote{#1}%
\addtocounter{footnote}{-1}%
\endgroup
}

\usepackage[accsupp]{axessibility}  

\usepackage[width=122mm,left=12mm,paperwidth=146mm,height=193mm,top=12mm,paperheight=217mm]{geometry}

\usepackage{graphicx}
\usepackage{comment}
\usepackage{cases}
\usepackage{amsmath,amssymb} 
\usepackage{color}
\usepackage{subfigure}
\usepackage{csquotes}

\usepackage{multirow}
\usepackage{booktabs}

\newcommand*{\twoelementtable}[3][l]%
{%
    \renewcommand{\arraystretch}{0.8}%
    \begin{tabular}[t]{@{}#1@{}}%
        #2\tabularnewline
        #3%
    \end{tabular}%
}
\usepackage{tabularx}
\usepackage{romannum}
\usepackage{mathtools}

\newtheorem{Remark}{Remark}

\begin{document}

\pagestyle{headings}
\mainmatter
\def\ECCVSubNumber{3438}  

\title{Domain Adaptive Video Segmentation via Temporal Pseudo Supervision}

\titlerunning{Domain Adaptive Video Segmentation via Temporal Pseudo Supervision}
%
\author{Yun Xing$^{1*}$ \quad
Dayan Guan$^{2*}$ \quad
Jiaxing Huang$^{1}$ \quad
Shijian Lu$^{1\dagger}$}

\institute{$^1$Nanyang Technological University \\ 
$^2$Mohamed bin Zayed University of Artificial Intelligence}

%
%

\maketitle

\begin{abstract}
Video semantic segmentation has achieved great progress under the supervision of large amounts of labelled training data. However, domain adaptive video segmentation, which can mitigate data labelling constraints by adapting from a labelled source domain toward an unlabelled target domain, is largely neglected. We design temporal pseudo supervision (TPS), a simple and effective method that explores the idea of consistency training for learning effective representations from unlabelled target videos. Unlike traditional consistency training that builds consistency in spatial space, we explore consistency training in spatiotemporal space by enforcing model consistency across augmented video frames which helps learn from more diverse target data. Specifically, we design cross-frame pseudo labelling to provide pseudo supervision from previous video frames while learning from the augmented current video frames. The cross-frame pseudo labelling encourages the network to produce high-certainty predictions, which facilitates consistency training with cross-frame augmentation effectively. Extensive experiments over multiple public datasets show that TPS is simpler to implement, much more stable to train, and achieves superior video segmentation accuracy as compared with the state-of-the-art. Code is available at \url{https://github.com/xing0047/TPS}.
\end{abstract}

\blfootnote{ $^{*}$Equal contribution.}
\blfootnote{ $^{\dagger}$Corresponding author.}

\begin{figure}[!ht]
\centering
\subfigure {\includegraphics[width=1.0\linewidth]{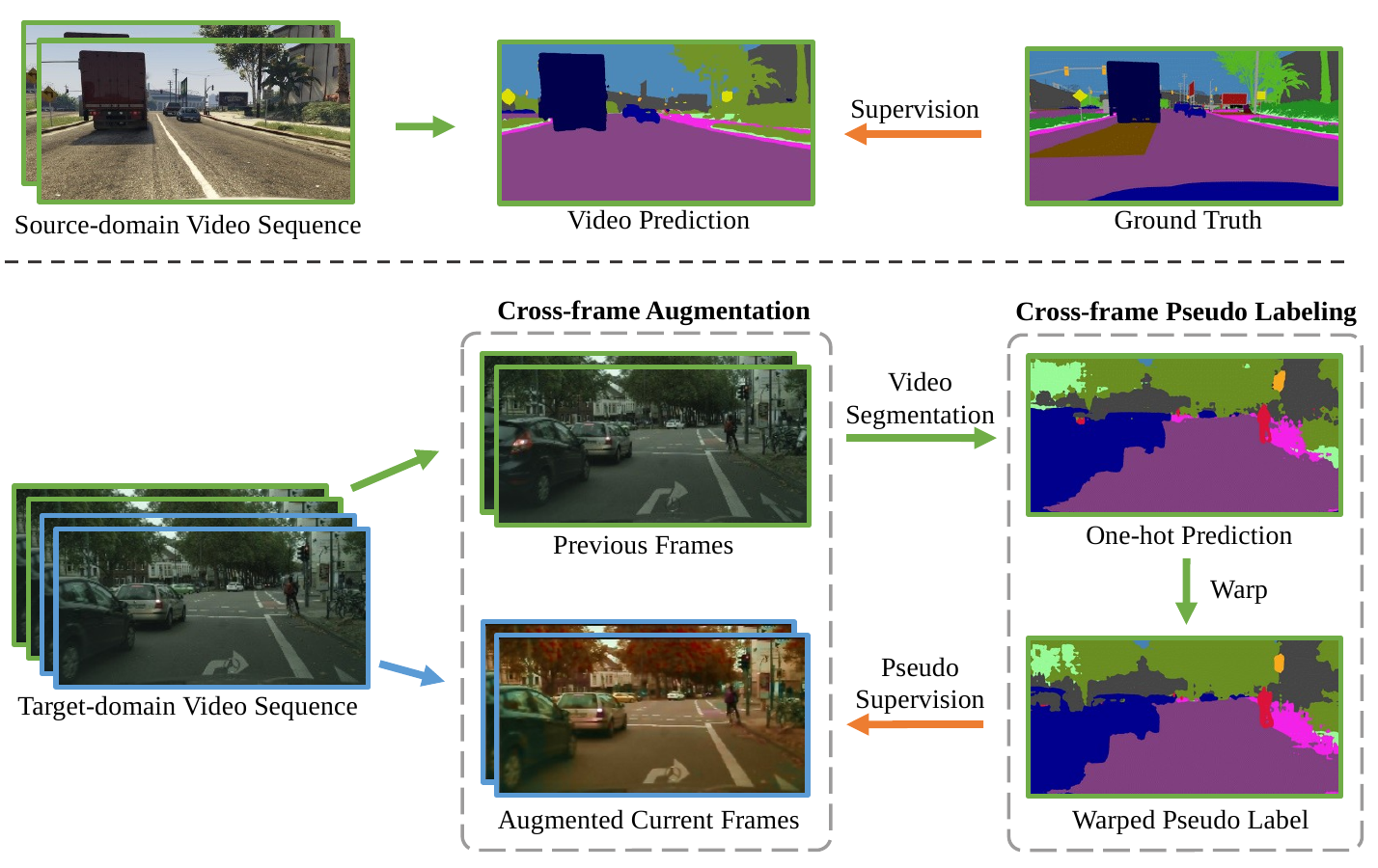}}
\caption{
The proposed temporal pseudo supervision (TPS) handles domain adaptive video segmentation by introducing \textit{\textbf{Cross-frame Augmentation}} and \textit{\textbf{Cross-frame Pseudo Labelling}} for consistency training in target domain. Specifically, the \textit{Cross-frame Pseudo Labelling} obtains one-hot predictions (taken as pseudo labels) for \textit{Previous Frames} and warps the predicted pseudo labels to the current video frames to supervise the learning from the \textit{Augmented Current Frames} that are generated by the \textit{\textbf{Cross-frame Augmentation}}.
}
\label{fig:intro}
\end{figure}

\section{Introduction}

Video semantic segmentation~\cite{floros2012joint,miksik2013efficient,couprie2013causal,liu2015multiclass,patraucean2015spatio}, which aims to predict a semantic label for each pixel in consecutive video frames, is a challenging task in computer vision research. With the advance of deep neural networks in recent years, video semantic segmentation has achieved great progress~\cite{shelhamer2016clockwork,kundu2016feature,gadde2017semantic,huang2018efficient,jain2019accel,kim2020video,hu2020temporally,liu2020efficient} by learning from large-scale and annotated video data~\cite{brostow2008segmentation,cordts2016cityscapes}. However, the annotation in video semantic segmentation involves pixel-level dense labelling which is prohibitively time-consuming and laborious to collect and has become one major constraint in supervised video segmentation. An alternative approach is to resort to synthetic data such as those rendered by game engines where pixel-level annotations are self-generated~\cite{richter2017playing,hernandez2017slanted}. On the other hand, video segmentation models trained with such synthetic data often experience clear performance drops~\cite{guan2021domain} while applied to real videos that usually have different distributions as compared with synthetic data.

Domain adaptive video segmentation aims for bridging distribution shifts across different video domains. Though domain adaptive image segmentation has been studied extensively, domain adaptive video segmentation is largely neglected despite its great values in various practical tasks. To the best of our knowledge, DA-VSN~\cite{guan2021domain} is the only work that explores adversarial learning and temporal consistency regularization to minimize the inter-domain temporal discrepancy and inter-frame discrepancy in target domain. However, DA-VSN relies heavily on adversarial learning which cannot guarantee a low empirical error on unlabelled target data~\cite{kumar2018co,chen2019progressive,zhang2021prototypical}, leading to negative effects on temporal consistency regularization in target domain. Consistency training is a prevalent semi-supervised learning technique that can guarantee a low empirical error on unlabelled data by enforcing model outputs to be invariant to data augmentation~\cite{xie2020unsupervised,sohn2020fixmatch,ouali2020semi}. It has recently been explored in domain adaptation tasks for guaranteeing a low empirical error on unlabelled target data~\cite{araslanov2021self,tranheden2021dacs,melas2021pixmatch}.

Motivated by consistency training in semi-supervised learning, we design a method named temporal pseudo supervision (TPS) that explores consistency training in spatiotemporal space for effective domain adaptive video segmentation. TPS works by enforcing model predictions to be invariant under the presence of cross-frame augmentation that is applied to the unlabelled target-domain video frames as illustrated in Fig.~\ref{fig:intro}. Specifically, TPS introduces cross-frame pseudo labelling that predicts pseudo labels for previous video frames. The predicted pseudo labels are then warped to the current video frames to enforce consistency with the prediction of the augmented current frames. Meanwhile, they also provide pseudo supervision for the domain adaptation model for learning from the augmented current frames. Compared with DA-VSN involving unstable adversarial learning, TPS is simpler to implement, more stable to train and achieve superior video segmentation performance consistently across multiple public datasets. 

The major contributions of this work can be summarized in three aspects. First, we introduce a domain adaptive video segmentation framework that addresses the challenge of absent target annotations from a perspective of consistency training. Second, we design an innovative consistency training method that constructs consistency in spatiotemporal space between the prediction of the augmented current video frames and the warped prediction of previous video frames. Third, we demonstrate that the proposed method achieves superior video segmentation performance consistently across multiple public datasets.

\section{Related works}
\subsection{Video Semantic Segmentation}
Video Semantic Segmentation is the challenging task of assigning a human-defined category to each pixel in each frame of a given video sequence. To tackle this challenge, the most natural and straightforward solution is to directly apply image segmentation approaches to each frame individually, in which way the model tends to ignore the temporal continuity in the video while training. A great many works explore on leveraging temporal consistency across frames by optical-flow guided feature fusion~\cite{zhu2017deep,gadde2017semantic,jain2019accel,liu2020efficient}, sequential network based representation aggregation~\cite{nilsson2018semantic} or joint learning of segmentation as well as optical flow estimation~\cite{hur2016joint,ding2020every}. 

Although video semantic segmentation has achieved great success under a supervised learning paradigm given a large amount of annotated data, pixel-wise video annotations are laborious and usually deficient to train a well-behaved network. Semi-supervised video segmentation aims at exploiting sparsely annotated video frames for segmenting unannotated frames of the same video. To make better use of unannotated data, a stream of work investigates on learning video segmentation network under annotation efficient settings by exploiting optical flow~\cite{mustikovela2016can,zhu2017deep,nilsson2018semantic,ding2020every}, patch matching~\cite{badrinarayanan2010label,budvytis2017large}, motion cues~\cite{tokmakov2016weakly,zhu2019improving}, pseudo-labeling~\cite{chen2020naive}, or self-supervised learning~\cite{wang2019learning,lai2020mast,jabri2020walk}. 

To further ease the burden of annotating, a popular line of study explores training segmentation network for real scene with synthetic data that can be automatically annotated, by either adversarial learning~\cite{tsai2018learning,huang2020contextual,vu2019advent,huang2021rda,pan2020unsupervised,huang2022multi,guan2021domain} or self-training~\cite{zou2019confidence,li2019bidirectional,lian2019constructing,chen2019domain,yang2020fda,kim2020learning,mei2020instance,huang2021model,huang2021cross,zheng2021rectifying,huang2022category,melas2021pixmatch}, which is known as domain adaptation. For domain adaptive video segmentation, DA-VSN~\cite{guan2021domain} is the only work that addresses the problem by incorporating adversarial learning to bridge the domain gap in temporal consistency. However, DA-VSN is largely constrained by adversarial learning that is unstable during training with high empirical risk. Different from adversarial learning~\cite{hoffman2017cycada,tsai2019domain,luo2019significance,guan2021uncertainty,xiao2022transfer,guan2021scale}, consistency training~\cite{xie2020unsupervised,sohn2020fixmatch,melas2021pixmatch,araslanov2021self} is widely explored in semi-supervised learning and domain adaptation recently with the benefits of its higher training stability and lower empirical risk. In this work, we propose to address the domain adaptive video segmentation by introducing consistency training across frames.

\subsection{Consistency Training}
Consistency training is a prevalent semi-supervised learning scheme that regularizes network predictions to be invariant to input perturbations~\cite{xie2020unsupervised,sohn2020fixmatch,ouali2020semi,guan2022unbiased,chen2021semi}. 
It intuitively makes sense as the model is supposed to be robust to small changes on inputs. Recent studies that focus on consistency training differ in how and where to set up perturbation. A great many works introduce random perturbation by Gaussian noise~\cite{french2017self}, stochastic regularization~\cite{sajjadi2016regularization,laine2016temporal} or adversarial noise~\cite{miyato2018virtual} at input level to enhance consistency training by enlarging sample space. More recently, it has been shown that stronger image augmentation~\cite{xie2020unsupervised,berthelot2019remixmatch,sohn2020fixmatch} can better improve the consistency training. Conceptually, the strong augmentation on images enriches the sample space of data, which can benefit the semi-supervised learning significantly.

Aside from the effectiveness of consistency training in semi-supervised learning, a line of recent studies explore adapting the strategy in domain adaptation tasks~\cite{araslanov2021self,tranheden2021dacs,melas2021pixmatch}. SAC~\cite{araslanov2021self} tackles domain adaptive segmentation by ensuring consistency between predictions from different augmented views. DACS~\cite{tranheden2021dacs} performs augmentation by mixing image patches from the two domains with swapping labels and pseudo labels accordingly. Derived from FixMatch~\cite{sohn2020fixmatch} which performs consistency training under the scenario of image classification, PixMatch~\cite{melas2021pixmatch} explores on various image augmentation strategies for domain adaptive image segmentation task. Unlike the aforementioned works involving consistency training in spatial space, we adopt consistency training in spatiotemporal space by enforcing model outputs invariant to cross-frame augmentation at the input level, which is devised to enrich the augmentation set and thus benefit the consistency training on unlabeled target videos.

\section{Method}
\subsection{Background}
Consistency training is a prevalent semi-supervised learning technique that enforces consistency between predictions on unlabeled images and the corresponding perturbed ones. Motivated by consistency training in semi-supervised learning, PixMatch~\cite{melas2021pixmatch} presents strong performance on domain adaptive segmentation by exploiting effective data augmentation on unlabeled target images. The idea is based on the assumption that a well-performed model should predict similarly when fed with strongly distorted inputs for unlabeled target data. Specifically, PixMatch performs pseudo labeling to provide pseudo supervision from original images for model training fed with augmented counterparts. As in FixMatch~\cite{sohn2020fixmatch}, the use of a hard label for consistency training in PixMatch encourage the model to obtain predictions with not only augmentation robustness but also high certainty on unlabeled data. Given a source-domain image $x^{\mathbb{S}}$ and its corresponding ground truth $y^{\mathbb{S}}$, together with an unannotated image $x^{\mathbb{T}}$ from the target domain, the training objective of PixMatch can be formulated as follows:
\begin{equation}
    \mathcal{L}_{\text{PixMatch}} = \mathcal{L}_{ce}(\mathcal{F}(x^{\mathbb{S}}),{y}^{\mathbb{S}}) + \lambda_{T}\mathcal{L}_{ce}(\mathcal{F}(\mathcal{A}(x^{\mathbb{T}})),\mathcal{P}(\mathcal{F}(x^{\mathbb{T}}), \tau)).
\label{eq:pixmatch}
\end{equation}
where $\mathcal{L}_{ce}$ is the cross-entropy loss, $\mathcal{F}$ and $\mathcal{A}$ denote the segmentation network and the transformation function for image augmentation, respectively. $\mathcal{P}$ represents the operation that selects pseudo labels given a confidence threshold of $\tau$. $\lambda_{T}$ is a hyperparameter that controls the trade-off between source and target losses while training. 

\subsection{Temporal Pseudo Supervision}
This work focus on the task of domain adaptive video segmentation. Different from PixMatch~\cite{melas2021pixmatch} that explored consistency training in spatial space for image-level domain adaptation, we propose a Temporal Pseudo Supervision (TPS) method to tackle the video-level domain adaptation by exploring spatio-temporal consistency training. Specifically, TPS introduces cross-frame augmentation for spatio-temporal consistency training to expand the diversity of image augmentation designed for spatial consistency training~\cite{melas2021pixmatch}. For the video-specific domain adaptation problem, we take adjacent frames as a whole in the form of $X_{k} = \mathcal{S}(x_{k-1}, x_{k})$, where $\mathcal{S}$ is a notation for stack operation. 

As for cross-frame augmentation in TPS, we apply image augmentation $\mathcal{A}$ defined in Eq.~\ref{eq:pixmatch} on the current frames $X_{k}^{\mathbb{T}}$ and such process is treated as performing cross-frame augmentation $\mathcal{A}^{cf}$ on previous frames $X_{k-\eta}^{\mathbb{T}}$, where $\eta$ is referred to as propagation interval which measures the temporal distance between the previous frames and the current frames. In this way, TPS can construct consistency training in spatiotemporal space by enforcing consistency between predictions on $\mathcal{A}^{cf}(X_{k-\eta}^{\mathbb{T}})$ and $X_{k-\eta}^{\mathbb{T}}$, which is different from PixMatch~\cite{melas2021pixmatch} that enforces spatial consistency between predictions on $\mathcal{A}(x^{\mathbb{T}})$ and $x^{\mathbb{T}}$ (as in Eq.~\ref{eq:pixmatch}). Formally, the cross-frame augmentation $\mathcal{A}^{cf}$ is defined as:
\begin{equation}
    \mathcal{A}^{cf}(X_{k-\eta}^{\mathbb{T}}) =  \mathcal{S}( \mathcal{A}(x_{k-1}^{\mathbb{T}}), \mathcal{A}(x_{k}^{\mathbb{T}})).
\label{eq:aug}
\end{equation}

\begin{Remark}
It is worth highlighting that the image augmentation $\mathcal{A}$ plays a crucial role in consistency training by strongly perturbing inputs to construct unseen views. As for the augmentation set $\mathcal{A}$, there have been studies~\cite{xie2020unsupervised,berthelot2019remixmatch,sohn2020fixmatch} presenting that stronger augmentation can benefit the consistency training more. To expand the diversity in image augmentation for the video task, we take the temporal deviation in video as a new  kind of data augmentation for the video task and combine it with $\mathcal{A}$, noted as $\mathcal{A}^{cf}$. To validate the effectiveness of cross-frame augmentation, we empirically compare TPS (using $\mathcal{A}^{cf}$) with PixMatch~\cite{melas2021pixmatch} (using $\mathcal{A}$) in Table~1 and~2.
\end{Remark}

With the constructed spatio-temporal space from cross-frame augmentation, TPS performs cross-frame pseudo labelling to provide pseudo supervision from previous video frames for network training fed with augmented current video frames. The cross-frame pseudo labelling has two roles: 1) facilitate the cross-frame consistency training that applies data augmentations across frames; 2) encourage the network to output video predictions with high certainty on unlabeled frames. 

Given a video sequence in target domain, we first forward previous video frames ${X}_{k-\eta}^{\mathbb{T}}$ through a video segmentation network $\mathcal{F}$ to obtain the previous frame prediction, and use FlowNet~\cite{dosovitskiy2015flownet} to produce the optical flow $o_{k-\eta\rightarrow{}k}$ estimated from the previous frame ${x}_{k-\eta}^{\mathbb{T}}$ and the current frame ${x}_{k}^{\mathbb{T}}$. Subsequently, the obtained previous frame prediction is warped using the estimated optical flow $o_{k-\eta\rightarrow{}k}$ to ensure the warped prediction is in line with the current frame temporally. We then perform pseudo labeling by utilizing a confidence threshold $\tau$ to filter out warped predictions with low confidence. In a nutshell, the process of cross-frame pseudo labelling can be formulated as:
\begin{equation}
    \mathcal{P}^{cf}(\mathcal{F}(X_{k-\eta}^{\mathbb{T}}), o_{k-\eta\rightarrow{}k}, \tau) = \mathcal{P}(\mathcal{W}(\mathcal{F}(X_{k-\eta}^{\mathbb{T}}), {o}_{{k-\eta}\rightarrow{k}}), \tau).
\label{eq:pl}
\end{equation}

\begin{Remark}
we would like to note that the confidence threshold $\tau$ is set to pick out high-confident predictions as pseudo labels for consistency training. There exist hard-to-transfer classes in the domain adaptive segmentation task (e.g. light, sign and rider in SYNTHIA-Seq~$\rightarrow$~Cityscapes-Seq) that tend to produce low confidence scores as compared to dominant classes, thus more possibly being ignored in pseudo labelling. To retain the pseudo label of hard-to-transfer classes as much as possible, we take 0 as the threshold $\tau$ for our experiments and further discussion about the effect of $\tau$ in Table~3. 
\end{Remark}

The training objective of TPS resembles Eq.~\ref{eq:pixmatch} in both source and target domain except that: 1) instead of feeding single images to the model, TPS takes adjacent video frames as inputs for video segmentation; 2) TPS replaces $\mathcal{A}$ in Eq.~\ref{eq:pixmatch} with a more diverse version $\mathcal{A}^{cf}$ to enrich the augmentation set by incorporating cross-frame augmentation; 3) in lieu of the straightforward pseudo labeling in Eq.~\ref{eq:pixmatch}, TPS resorts to cross-frame pseudo labeling that propagates video prediction from previous frames across optical flow $o_{k-\eta\rightarrow{}k}$ before further step. In a nutshell, given source-domain video frames $X^{\mathbb{S}}$ along with the target-domain video sequence, we formulate our TPS as: 
\begin{equation}
    \begin{split}
        \mathcal{L}_{\text{TPS}} = & \mathcal{L}_{ce}(\mathcal{F}(X^{\mathbb{S}}), y^{\mathbb{S}}) + \lambda_T\mathcal{L}_{ce}(\mathcal{F}(\mathcal{A}^{cf}(X_{k-\eta}^{\mathbb{T}})), \mathcal{P}^{cf}(\mathcal{F}(X_{k-\eta}^{\mathbb{T}}),  o_{k-\eta\rightarrow{}k}, \tau)).
    \end{split}
\label{eq:tps}
\end{equation}
    
\begin{Remark}
We should point out that $\lambda_{T}$ is set to balance the training between source and target domain as in DA-VSN. In spite of the effectiveness of DA-VSN on domain adaptive video segmentation task, the training process of adversarial learning is inherently unstable with feeding complex or irrelevant cues to the discriminator while training~\cite{luo2019significance}. To alleviate the effect, DA-VSN set $\lambda_{T}$ to 0.001 to stabilize the training process whereas compromise the domain adaptation performance. Different from the previous work, we leverage the inherent stability of consistency training and naturally set $\lambda_{T}$ to 1.0 for our TPS to treat learning of source and target equally. We further make comparison on the stability of training process between DA-VSN and TPS by visualization in Fig.~\ref{fig:loss} and explore on the effect of $\lambda_{T}$ on the performance 
in Table~5.
\end{Remark}

\section{Experiments}

\subsection{Experimental Setting}
\paragraph{\textbf{Datasets.}} 

To validate our method, we conduct comprehensive experiments under two challenging synthetic-to-real benchmarks for domain adaptive video segmentation: SYNTHIA-Seq~\cite{ros2016synthia} $\rightarrow$ Cityscapes-Seq~\cite{cordts2016cityscapes} and VIPER~\cite{richter2017playing} $\rightarrow$ Cityscapes-Seq. As in~\cite{guan2021domain}, we treat either SYNTHIA-Seq or VIPER as source-domain data and take Cityscapes-Seq as the target-domain data. 

\paragraph{\textbf{Implementation details.}} 
As in~\cite{guan2021domain}, we take ACCEL~\cite{jain2019accel} as the video segmentation framework, which is composed of double segmentation branches and an optical flow estimation branch, together with a fusion layer at the output level. Specifically, both branches for segmentation forward a single video frame through Deeplab~\cite{chen2017deeplab}. Meanwhile, the branch of optical flow estimation~\cite{dosovitskiy2015flownet} produces the corresponding optical flow of the adjacent video frames, which can be further used in a score fusion layer to integrate frame prediction from different views. As regard to the training process, we use SGD as the optimizer with momentum and weight decay set to $0.9$ and $5\times10^{-4}$ respectively. The model is trained with a learning rate of $2.5\times10^{-4}$ for 40k iterations. As in~\cite{sohn2020fixmatch,melas2021pixmatch}, we incorporate multiple augmentations in our experiments, including gaussian blur, color jitter and random scaling. The mean intersection-over-union (mIoU) is used to evaluate all methods. For the efficiency of training and inference, we apply bicubic interpolation to resize every video frame in Cityscapes-Seq and VIPER to $512\times1024$, $720\times1280$, respectively. All the experiments are implemented on a single GPU with 11~GB memory.

\renewcommand\arraystretch{1.36}
\begin{table}[!ht]
\label{tab:synthia}
\centering
\caption{Quantitative comparisons over the benchmark of SYNTHIA-Seq~$\rightarrow$~Cityscapes-Seq: TPS outperforms multiple domain adaptation methods by large margins. These methods include the only domain adaptive video segmentation method~\cite{guan2021domain}, the most related domain adaptive segmentation method~\cite{melas2021pixmatch} and other domain adaptive segmentation approaches~\cite{vu2019advent,zou2018unsupervised,pan2020unsupervised,zou2019confidence,huang2020contextual,huang2021rda,yang2020fda} which serve as baselines. Note that ``Source only'' denotes the network trained with source-domain data solely}
\begin{scriptsize}
\begin{tabular}{p{2cm}|*{11}{p{0.7cm}}|p{0.7cm}}
 \toprule
 \multicolumn{13}{c}{\textbf{SYNTHIA-Seq~$\rightarrow$~Cityscapes-Seq}} \\
 \midrule
 Methods  &{road} &{side.} &{buil.} &{{pole}} &{{light}} &{{sign}} &{vege.} &{sky} &{{pers.}} &{{rider}} &{{car}} &mIoU  \\
 \midrule
 Source only &56.3 &26.6 &75.6 &25.5 &5.7 &15.6 &71.0 &58.5 &41.7 &17.1 &27.9 &38.3 \\
 \midrule
 AdvEnt~\cite{vu2019advent}  &85.7 &21.3 &70.9 &21.8 &4.8 &15.3 &59.5 &62.4 &46.8 &16.3 &64.6 &42.7 \\
 CBST~\cite{zou2018unsupervised} &64.1 &30.5 &78.2 &\textbf{28.9} &14.3 & 21.3 &75.8 &62.6 &46.9 &20.2 &33.9 &43.3 \\
 IDA~\cite{pan2020unsupervised} &87.0 &23.2 &71.3 &22.1 &4.1 &14.9 &58.8 &67.5 &45.2 &17.0 &73.4 &44.0 \\
 CRST~\cite{zou2019confidence} &70.4 &31.4 &\textbf{79.1} &27.6 &11.5 &20.7 &\textbf{78.0} &67.2 &49.5 &17.1 &39.6 &44.7 \\ 
 CrCDA~\cite{huang2020contextual} &86.5	&26.3	&74.8	&24.5	&5.0	&15.5	&63.5	&64.4	&46.0	&15.8	&72.8	&45.0\\
 RDA~\cite{huang2021rda} &84.7	&26.4	&73.9	&23.8	&7.1	&18.6	&66.7	&68.0	&48.6	&9.3	&68.8	&45.1\\
 FDA~\cite{yang2020fda} &84.1 & 32.8 & 67.6 & 28.1 & 5.5 &20.3 &61.1 &64.8 &43.1 &19.0 &70.6 &45.2 \\ 
 \midrule
 DA-VSN~\cite{guan2021domain} &89.4 &31.0 &77.4 &26.1 &9.1 &20.4 &75.4 &\textbf{74.6} &42.9 &16.1 &82.4 &49.5 \\
 PixMatch~\cite{melas2021pixmatch} &90.2& 49.9& 75.1& 23.1& 17.4& 34.2& 67.1& 49.9& 55.8& 14.0& 84.3&51.0 \\
 \textbf{TPS (Ours)} &\textbf{91.2}& \textbf{53.7}& 74.9& 24.6& \textbf{17.9}& \textbf{39.3}& 68.1& 59.7& \textbf{57.2}& \textbf{20.3}& \textbf{84.5}& \textbf{53.8} \\
\bottomrule
\end{tabular}
\end{scriptsize}
\end{table}

\subsection{Comparison with State-of-the-art}
\label{sec:comparison}
We compare the proposed TPS mainly with the most related methods DA-VSN~\cite{guan2021domain} and PixMatch~\cite{melas2021pixmatch}, considering the fact that DA-VSN is current state-of-the-art method on domain adaptive video segmentation (the same task as in this work) and PixMatch is the state-of-the-art method on domain adaptive image segmentation using consistency training (the same learning scheme as in this work). Quantitative comparisons are shown 
in Table~1 and~2.
We note that TPS surpasses DA-VSN by a large margin on the benchmark of both SYNTHIA-Seq$\rightarrow$Cityscapes-Seq (4.3\% in mIoU) and VIPER$\rightarrow$Cityscapes-Seq (1.1\% in mIoU), which presents the superiority of consistency training over adversarial learning for domain adaptive video segmentation. Additionally, we highlight that our method TPS outperforms PixMatch on both benchmarks (a mIoU of 2.8\% and 2.2\%, respectively) which corroborates the effectiveness of the cross-frame augmentation for consistency training on video-specific task. 
In addition, we also compare our method with multiple baselines~\cite{vu2019advent,zou2018unsupervised,pan2020unsupervised,zou2019confidence,huang2020contextual,huang2021rda,yang2020fda} which were originally devised for domain adaptive image segmentation. These baselines are based on adversarial learning~\cite{vu2019advent,pan2020unsupervised,huang2020contextual} and self-training~\cite{zou2018unsupervised,zou2019confidence,yang2020fda,huang2021rda}.
As in~\cite{guan2021domain}, We apply these approaches by simply replacing the image segmentation model with our video segmentation backbone and implement domain adaptation similarly. As presented in Table~\ref{tab:synthia} and~\ref{tab:viper}, TPS surpasses all baselines by large margins, demonstrating the advantage of our video-specific approach as compared to image-specific ones. 

\renewcommand\arraystretch{1.36}
\begin{table}[!ht]
\label{tab:viper}
\centering
\caption{Quantitative comparisons over the benchmark of VIPER~$\rightarrow$~Cityscapes-Seq: TPS outperforms multiple domain adaptation methods by large margins
}
\begin{scriptsize}
\begin{tabular}{p{1.8cm}|*{15}{p{0.54cm}}|p{0.6cm}}
 \toprule
 \multicolumn{17}{c}{\textbf{VIPER~$\rightarrow$~Cityscapes-Seq}} \\
 \midrule
 Methods  &{road} &{side.} &{buil.} &{fence} &{light} &{sign} &{vege.} &{terr.} &{sky} &{pers.} &{car} &{truck} &{bus} &{mot.} &{bike} &mIoU \\
 \midrule
 Source only &56.7 &18.7 &78.7 &6.0 &22.0 &15.6 &81.6 &18.3 &80.4 &59.9 &66.3 &4.5 &16.8 &20.4 &10.3 &37.1 \\
  \midrule
 AdvEnt~\cite{vu2019advent} &78.5 &31.0 &81.5 &22.1 &29.2 &26.6 &81.8 &13.7 &80.5 &58.3 &64.0 &6.9 &38.4 &4.6 &1.3 &41.2 \\
 CBST~\cite{zou2018unsupervised} &48.1 &20.2 &\textbf{84.8} &12.0 &20.6 &19.2 &83.8 &18.4 &\textbf{84.9} &59.2 &71.5 &3.2 &38.0 &23.8 &\textbf{37.7} &41.7 \\
 IDA~\cite{pan2020unsupervised} &78.7 &33.9 &82.3 &22.7 &28.5 &26.7 &82.5 &15.6 &79.7 &58.1 &64.2 &6.4 &41.2 &6.2 &3.1 &42.0 \\
 CRST~\cite{zou2019confidence} &56.0 &23.1 &82.1 &11.6 &18.7 &17.2 &\textbf{85.5} &17.5 &82.3 &60.8 &73.6 &3.6 &38.9 &\textbf{30.5} &35.0 &42.4 \\
 CrCDA~\cite{huang2020contextual} &78.1	&33.3	&82.2	&21.3	&29.1	&26.8	&82.9	&28.5	&80.7	&59.0	&73.8	&16.5	&41.4	&7.8	&2.5	&44.3\\
 RDA~\cite{huang2021rda} &72.0	&25.9	&80.8	&15.1	&27.2	&20.3	&82.6	&\textbf{31.4}	&82.2	&56.3	&75.5	&22.8	&48.3	&19.1	&6.7	&44.4\\
 FDA~\cite{yang2020fda} &70.3 &27.7 &81.3 &17.6 &25.8 &20.0 &83.7 &31.3 &82.9 &57.1 &72.2 &22.4 &\textbf{49.0} &17.2 &7.5 &44.4 \\
  \midrule
 PixMatch~\cite{melas2021pixmatch} &79.4& 26.1& 84.6& 16.6& 28.7& 23.0& 85.0& 30.1& 83.7& 58.6& 75.8& 34.2& 45.7& 16.6& 12.4 &46.7 \\
 DA-VSN~\cite{guan2021domain} &\textbf{86.8} &36.7 &83.5 &\textbf{22.9} &\textbf{30.2} &27.7 &83.6 &26.7 &80.3 &60.0 &79.1 &20.3 &47.2 &21.2 &11.4 &47.8 \\
 \textbf{TPS (Ours)} & 82.4& \textbf{36.9}& 79.5& 9.0& 26.3& \textbf{29.4}& 78.5& 28.2& 81.8& \textbf{61.2}& \textbf{80.2}& \textbf{39.8}& 40.3& 28.5& 31.7& \textbf{48.9} \\
\bottomrule
\end{tabular}
\end{scriptsize}
\end{table}

\begin{figure}[!ht]
\centering
\begin{minipage}[c]{0.03\linewidth}
\centering {\rotatebox{90}{Frames}} 
\end{minipage}
\vspace{2pt}
\begin{minipage}[c]{0.31\linewidth}
\centering\includegraphics[width=.99\linewidth]{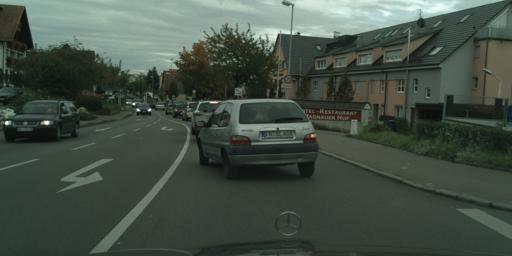}
\end{minipage}
\begin{minipage}[c]{0.31\linewidth}
\centering\includegraphics[width=.99\linewidth]{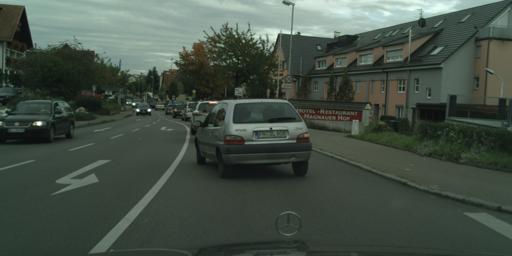}
\end{minipage}
\begin{minipage}[c]{0.31\linewidth}
\centering\includegraphics[width=.99\linewidth]{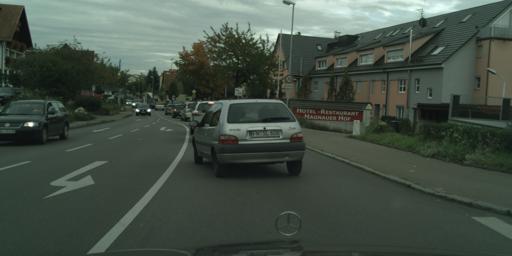}
\end{minipage}
\vspace{2pt}
\centering
\begin{minipage}[c]{0.03\linewidth}
\centering {\rotatebox{90}{GT}} 
\end{minipage}
\begin{minipage}[c]{0.31\linewidth}
\centering\includegraphics[width=.99\linewidth]{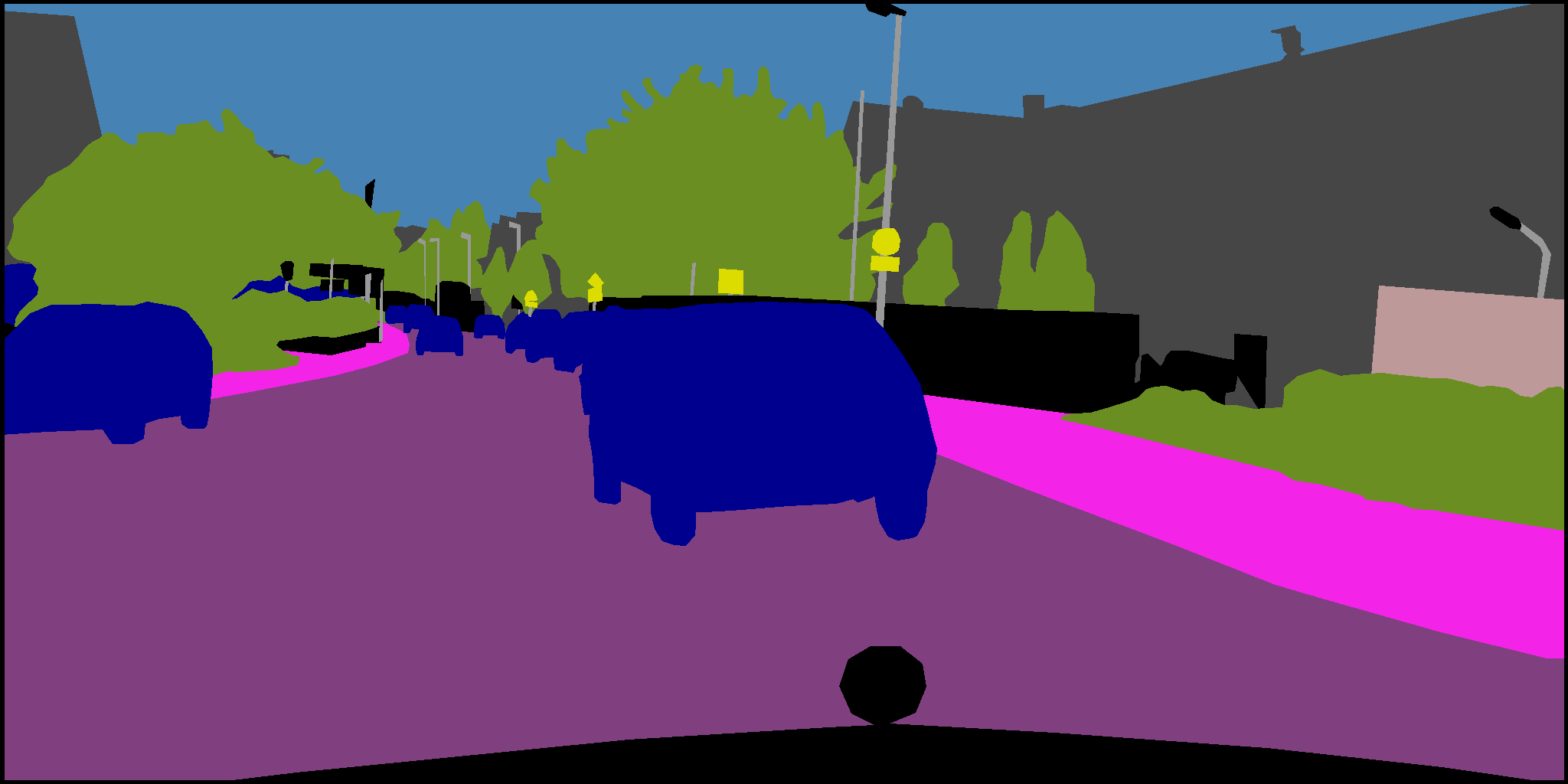}
\end{minipage}
\begin{minipage}[c]{0.31\linewidth}
\centering\includegraphics[width=.99\linewidth]{images/results/lindau_000000_000019/lindau_000000_000019_gt.png}
\end{minipage}
\begin{minipage}[c]{0.31\linewidth}
\centering\includegraphics[width=.99\linewidth]{images/results/lindau_000000_000019/lindau_000000_000019_gt.png}
\end{minipage}
\vspace{2pt}
\centering
\begin{minipage}[c]{0.03\linewidth}
\centering {\rotatebox{90}{Source Only}} 
\end{minipage}
\begin{minipage}[c]{0.31\linewidth}
\centering\includegraphics[width=.99\linewidth]{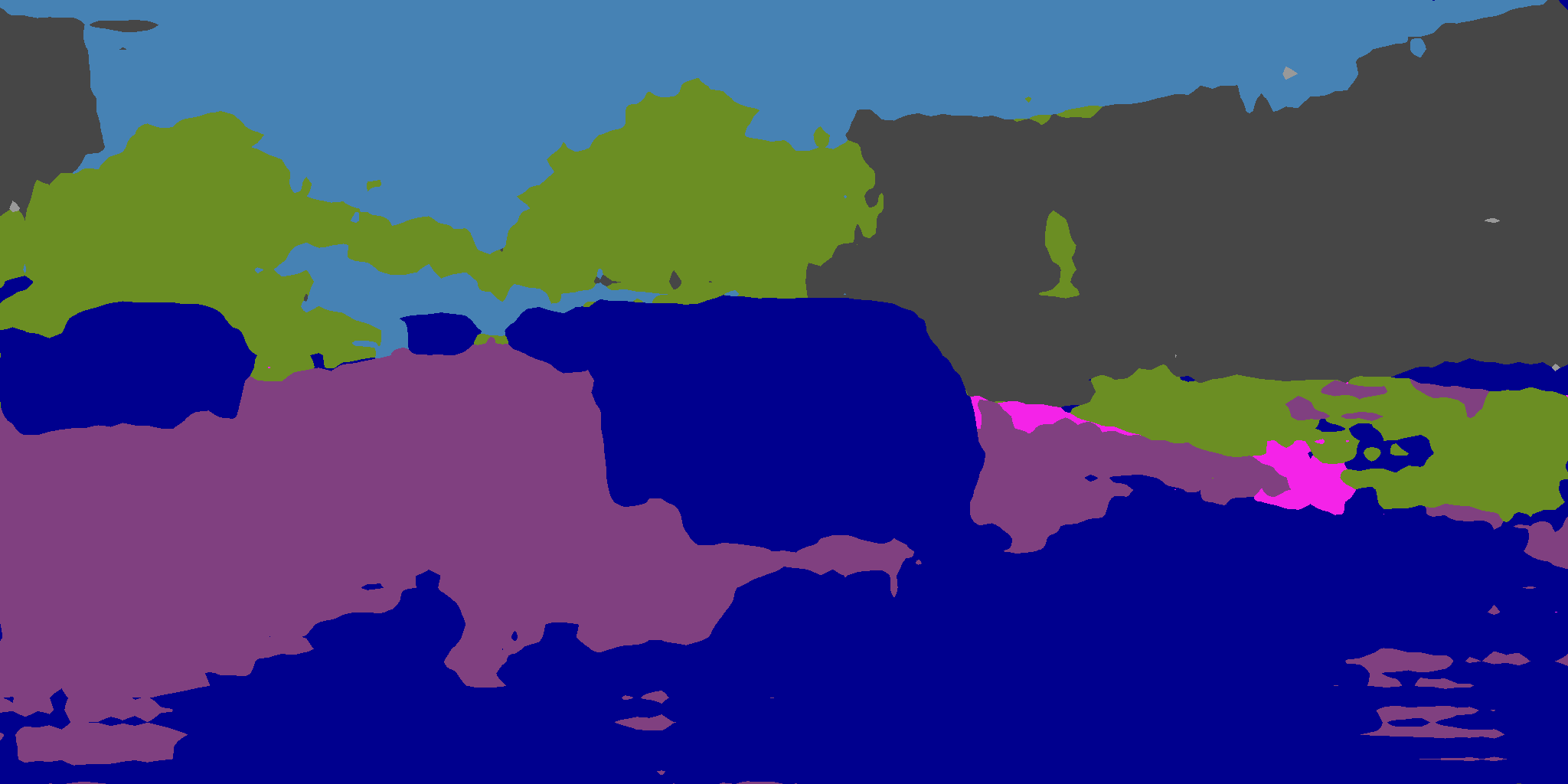}
\end{minipage}
\begin{minipage}[c]{0.31\linewidth}
\centering\includegraphics[width=.99\linewidth]{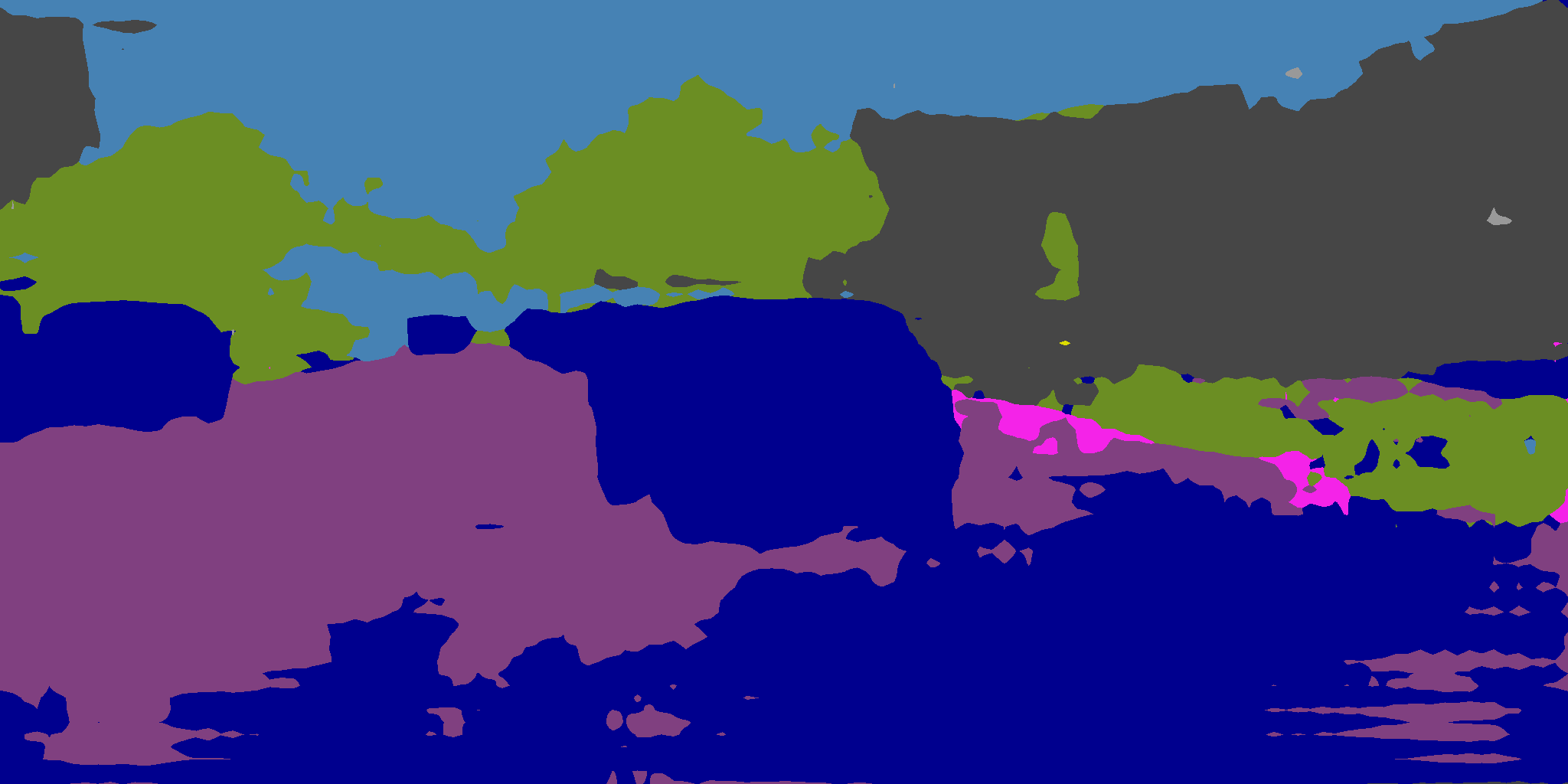}
\end{minipage}
\begin{minipage}[c]{0.31\linewidth}
\centering\includegraphics[width=.99\linewidth]{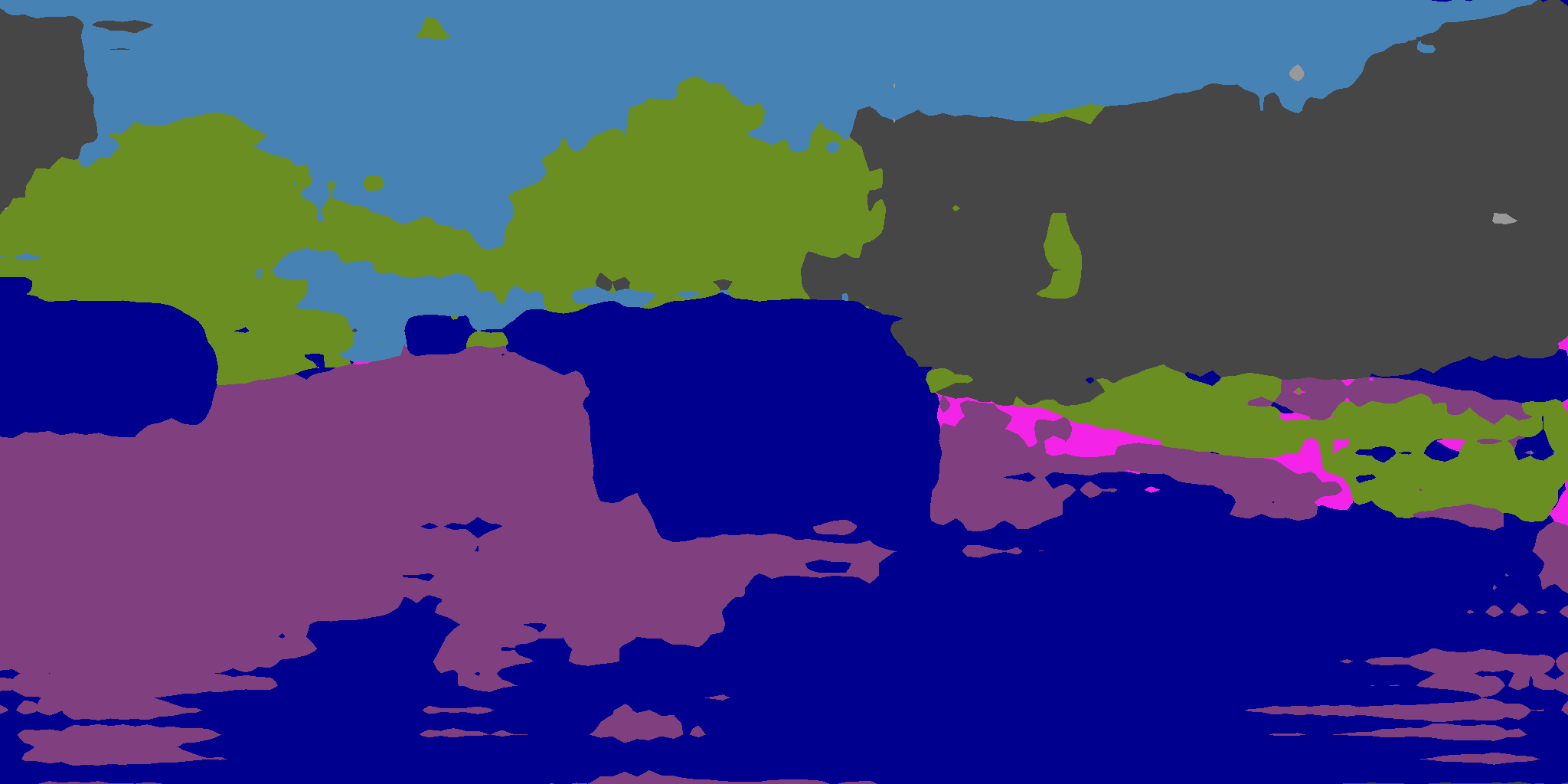}
\end{minipage}
\vspace{2pt}
\centering
\begin{minipage}[c]{0.03\linewidth}
\centering {\rotatebox{90}{DA-VSN}} 
\end{minipage}
\begin{minipage}[c]{0.31\linewidth}
\centering\includegraphics[width=.99\linewidth]{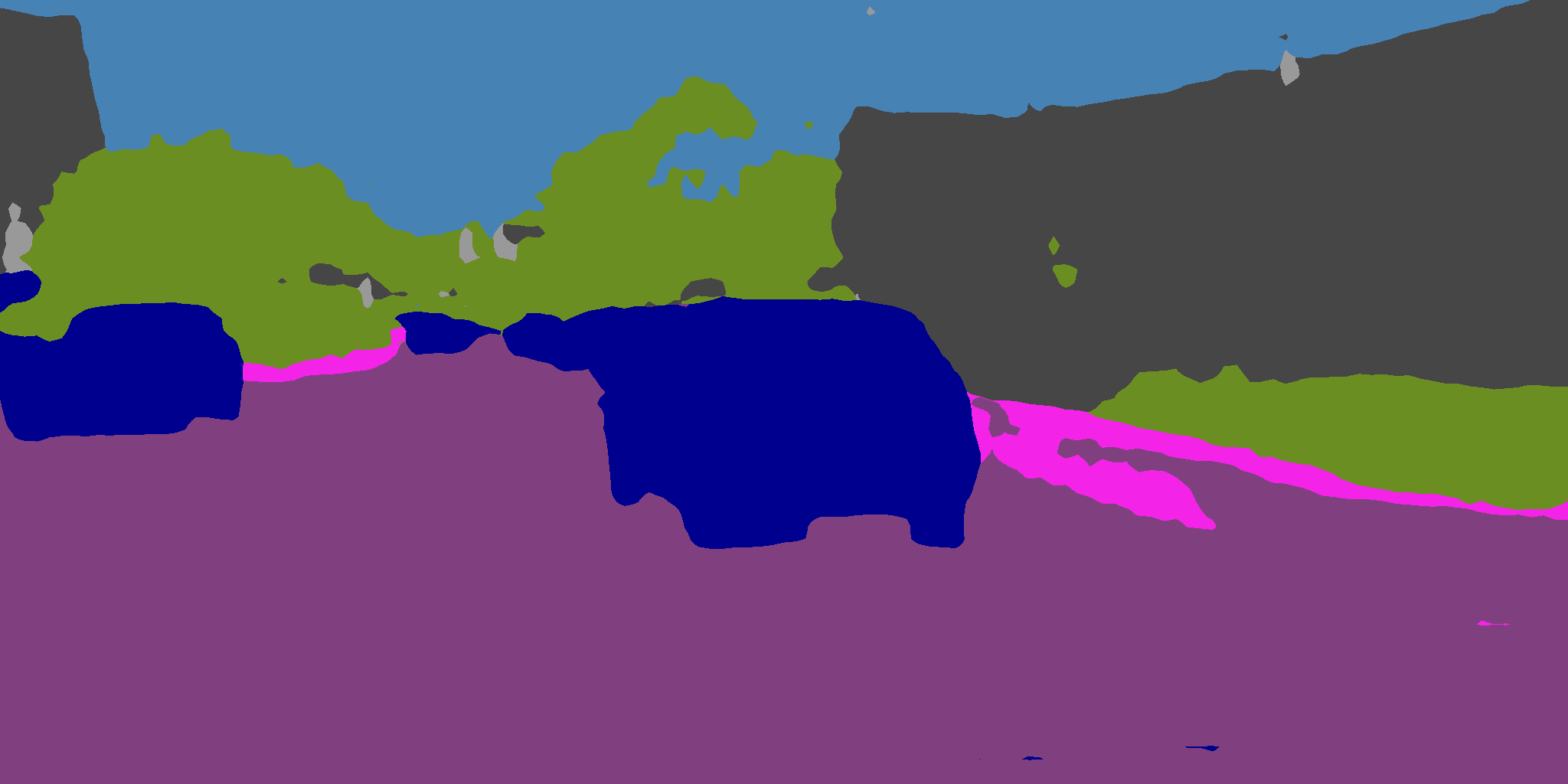}
\end{minipage}
\begin{minipage}[c]{0.31\linewidth}
\centering\includegraphics[width=.99\linewidth]{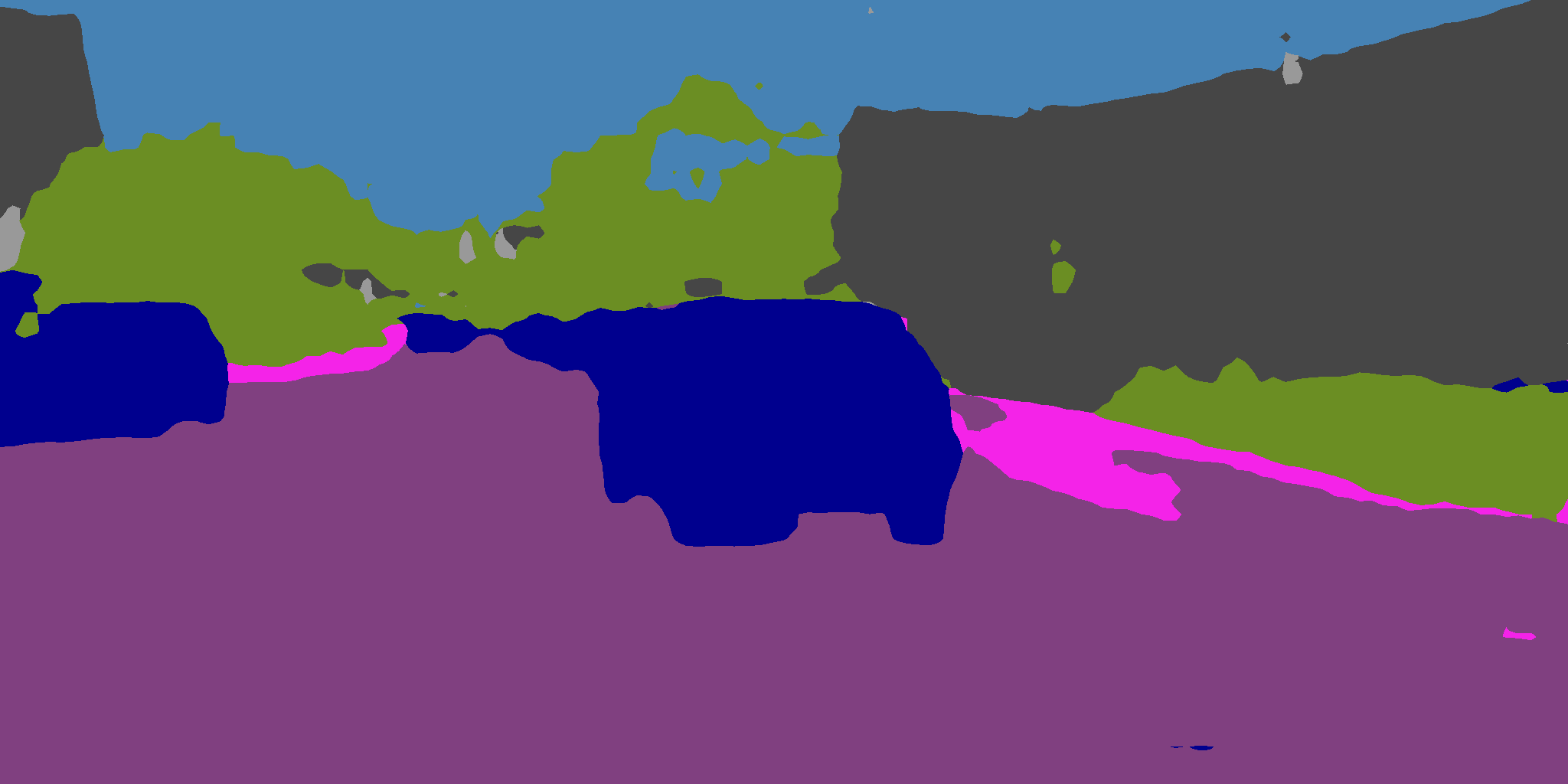}
\end{minipage}
\begin{minipage}[c]{0.31\linewidth}
\centering\includegraphics[width=.99\linewidth]{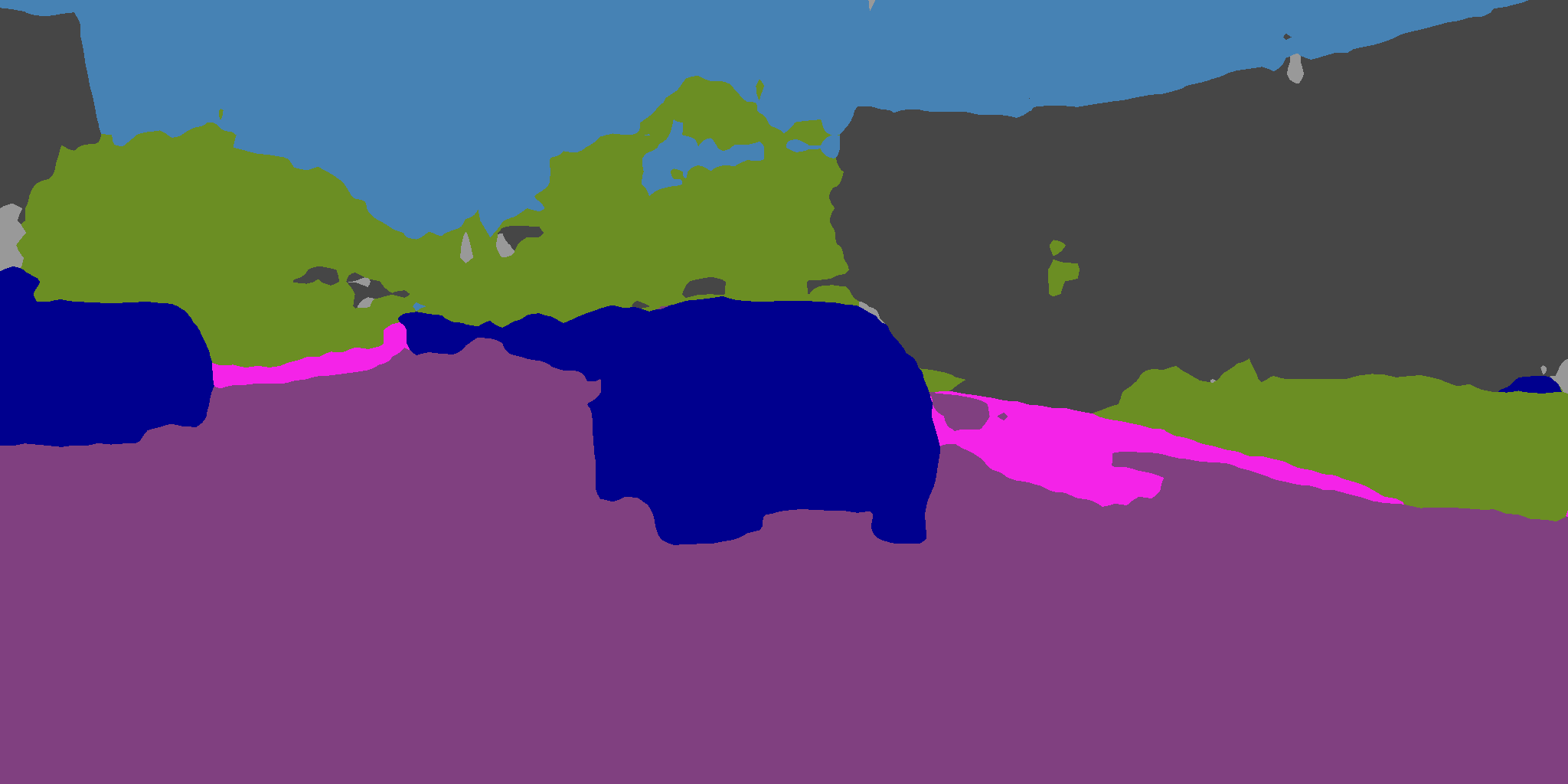}
\end{minipage}
\vspace{2pt}
\centering
\begin{minipage}[c]{0.03\linewidth}
\centering {\rotatebox{90}{PixMatch}} 
\end{minipage}
\begin{minipage}[c]{0.31\linewidth}
\centering\includegraphics[width=.99\linewidth]{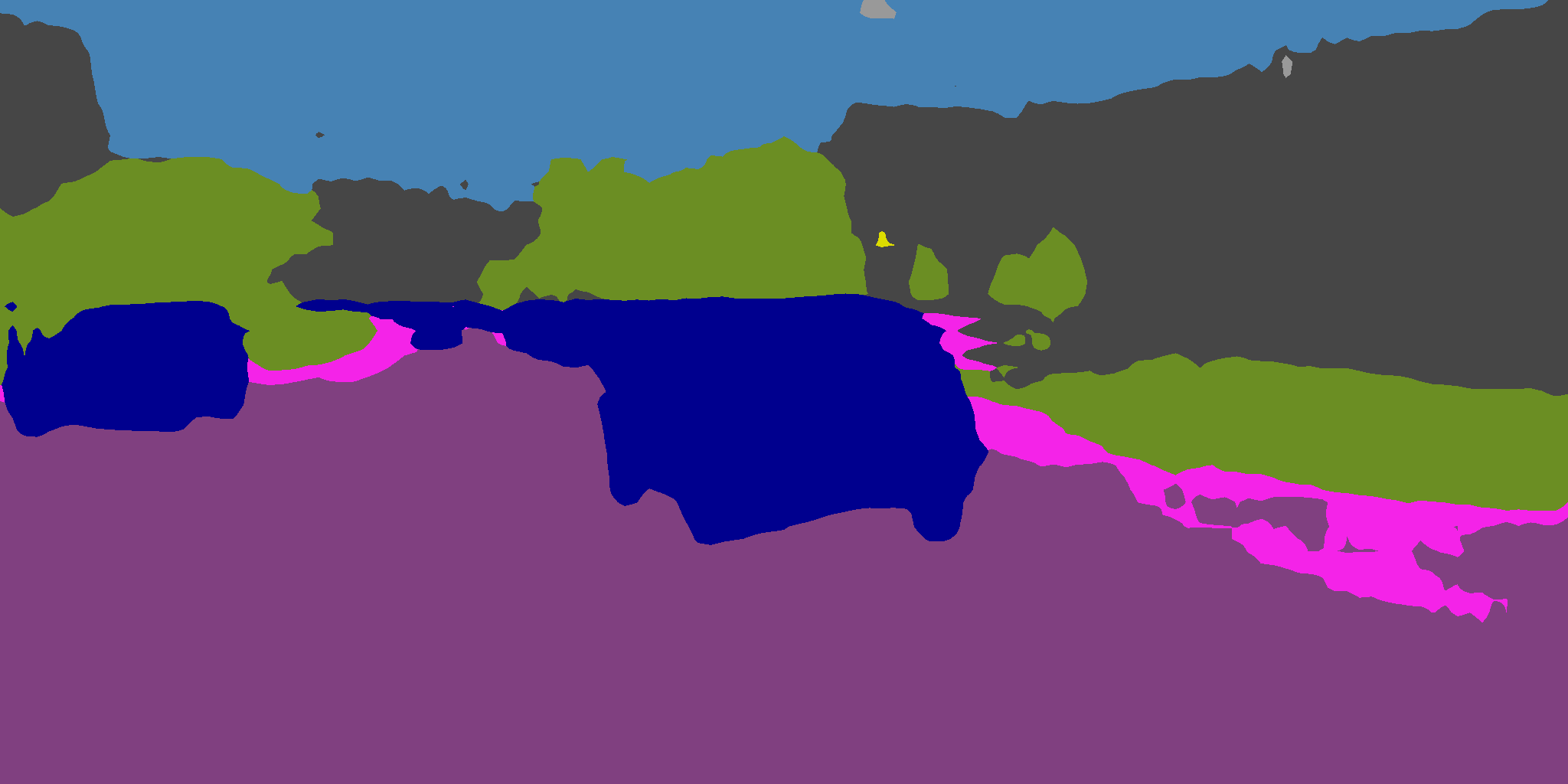}
\end{minipage}
\begin{minipage}[c]{0.31\linewidth}
\centering\includegraphics[width=.99\linewidth]{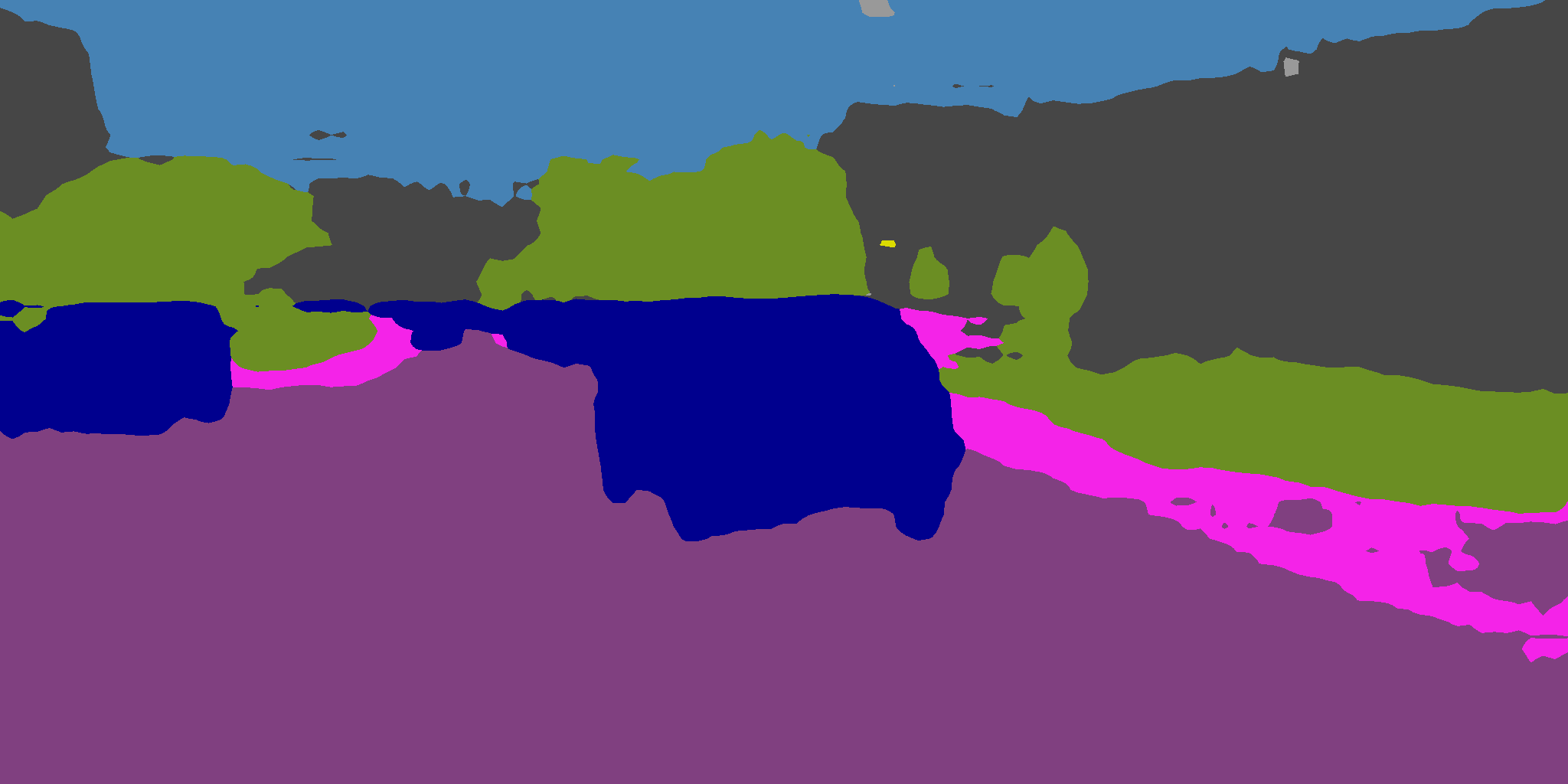}
\end{minipage}
\begin{minipage}[c]{0.31\linewidth}
\centering\includegraphics[width=.99\linewidth]{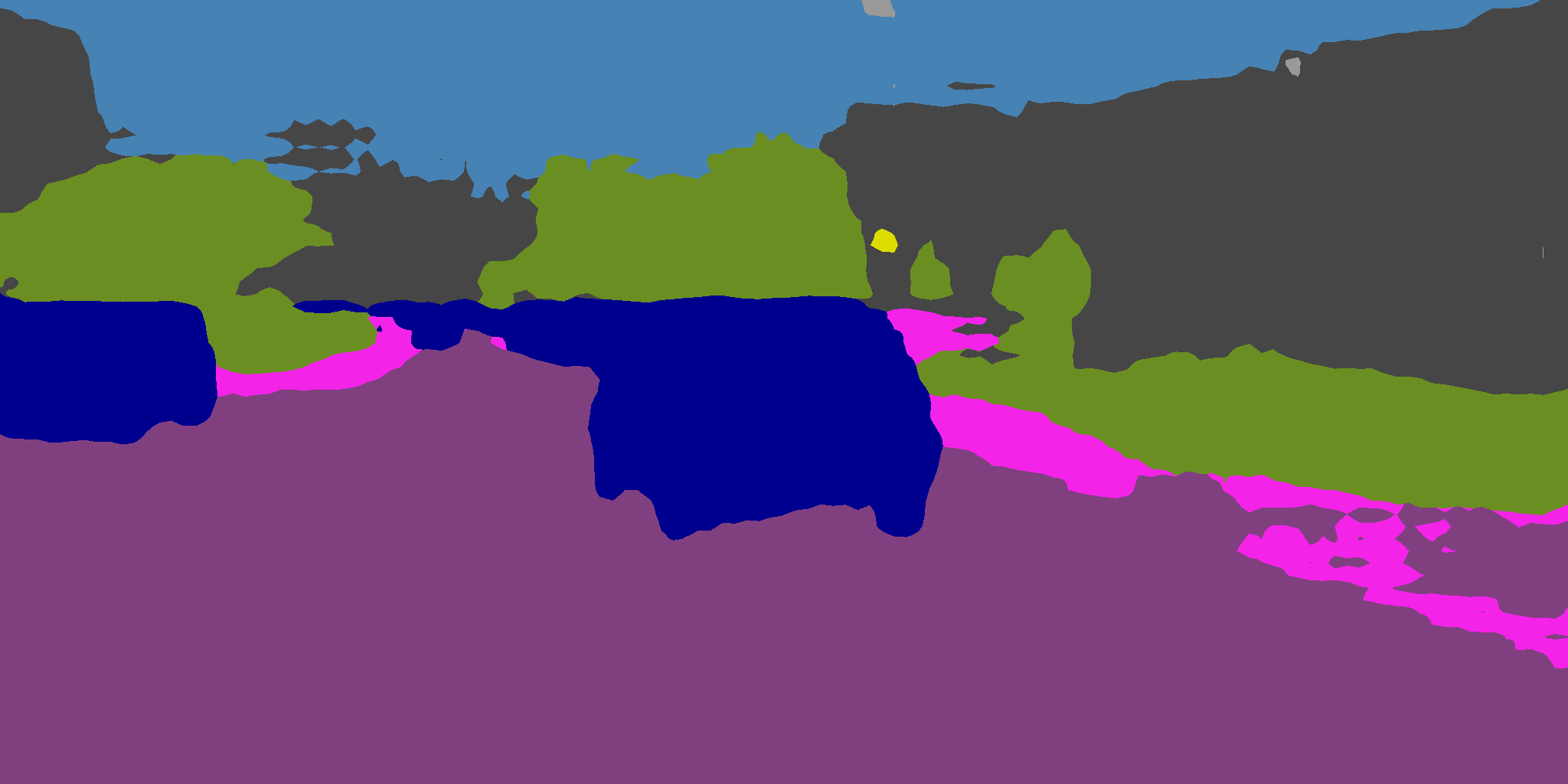}
\end{minipage}
\vspace{2pt}
\centering
\begin{minipage}[c]{0.03\linewidth}
\centering {\rotatebox{90}{\textbf{TPS~(Ours)}}} 
\end{minipage}
\begin{minipage}[c]{0.31\linewidth}
\centering\includegraphics[width=.99\linewidth]{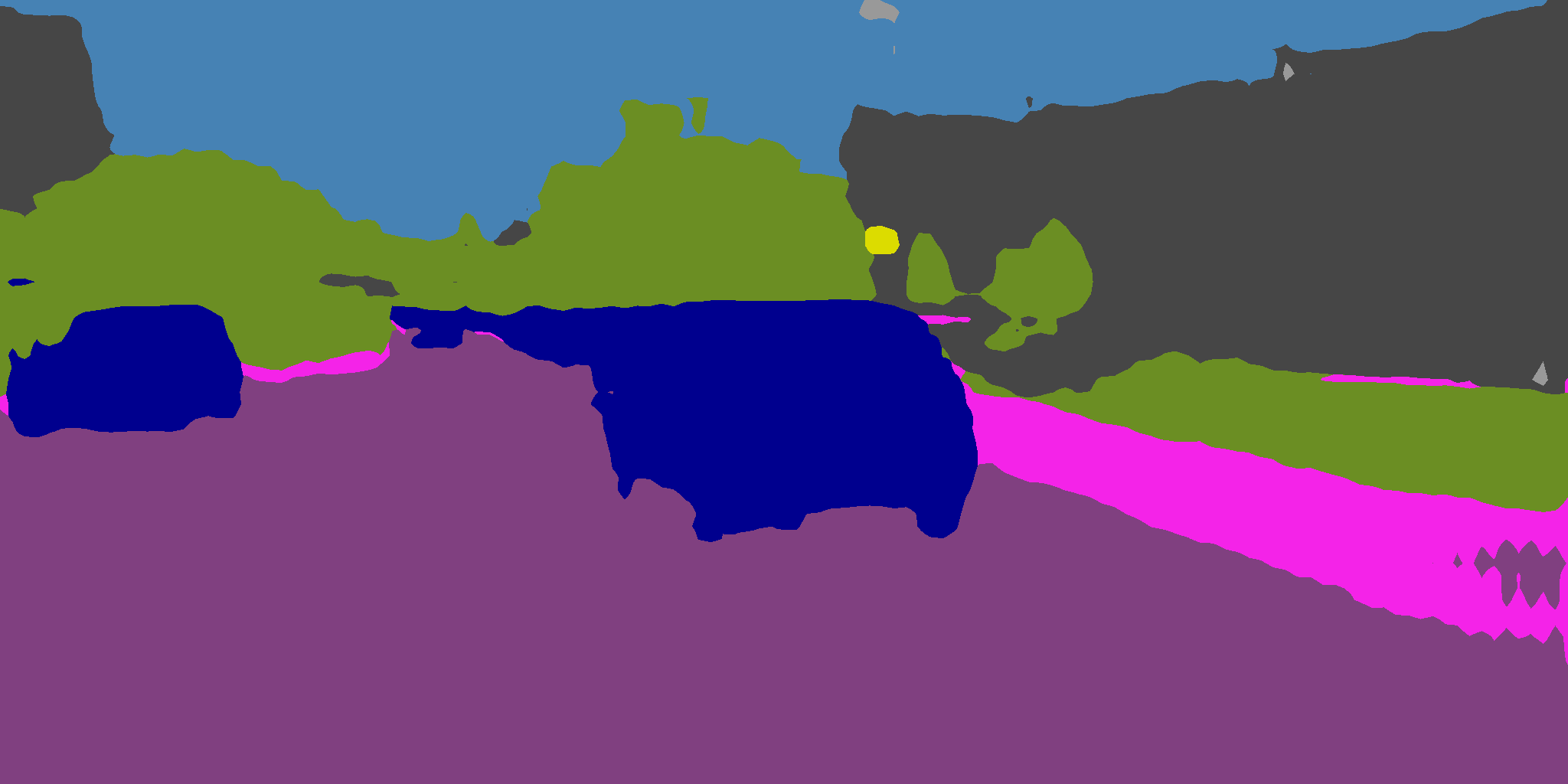}
\end{minipage}
\begin{minipage}[c]{0.31\linewidth}
\centering\includegraphics[width=.99\linewidth]{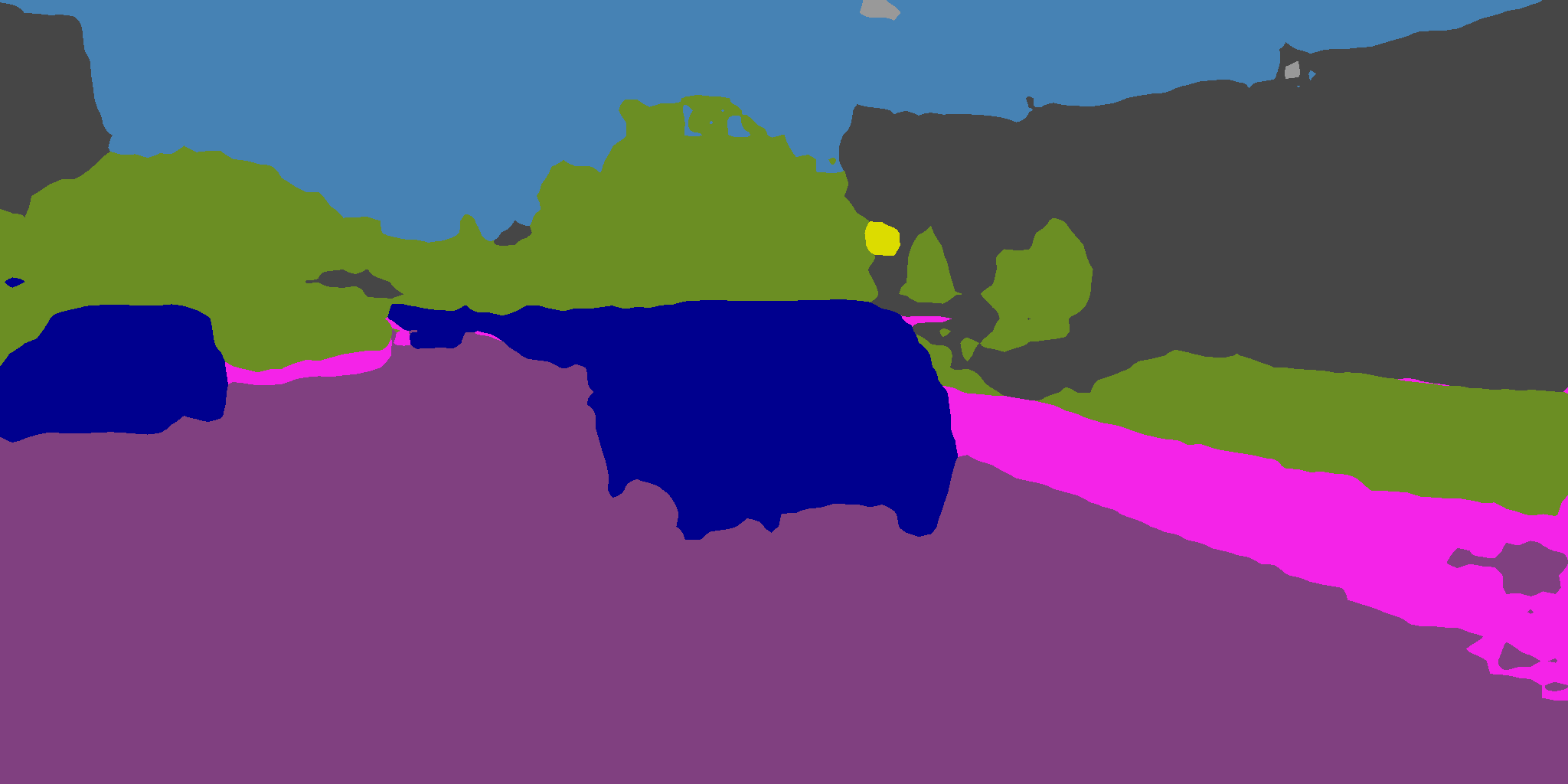}
\end{minipage}
\begin{minipage}[c]{0.31\linewidth}
\centering\includegraphics[width=.99\linewidth]{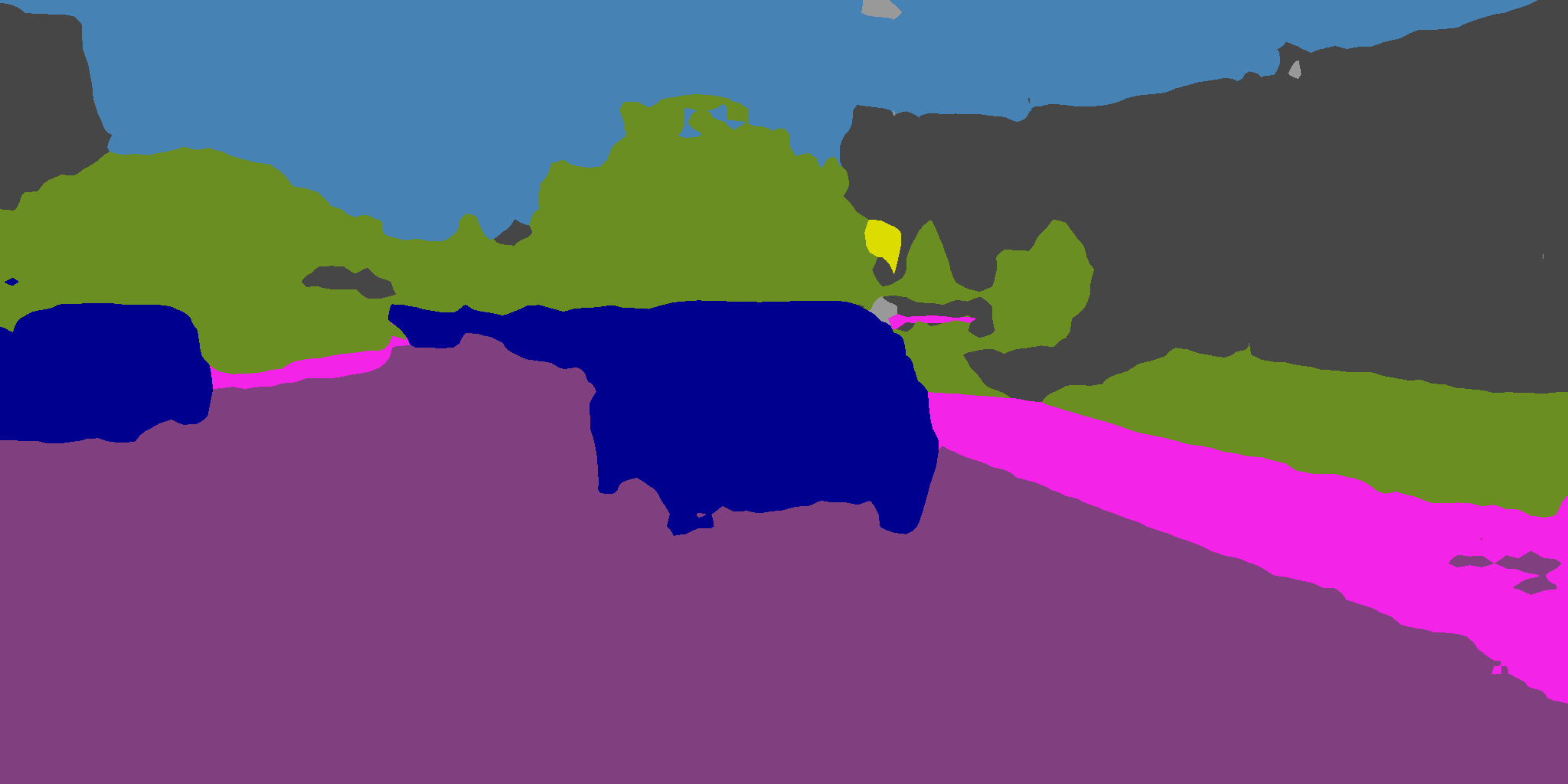}
\end{minipage}
\centering
\caption{Qualitative comparison of TPS with the state-of-the-art over domain adaptive video segmentation benchmark \enquote{SYNTHIA-Seq~$\rightarrow$~Cityscapes-Seq}: TPS produces much more accurate segmentation as compared to ``source only'', indicating the effectiveness of our approach on addressing domain adaptation issue. Moreover, TPS generates better segmentation than PixMatch and DA-VSN as shown in rows 4-5, which is consistent with our quantitative result. Best viewed in color.}
\label{fig:result}
\end{figure}

Furthermore, we present the qualitative result in Fig.~\ref{fig:result} to demonstrate the superiority of our method. We point out that despite the impressive adaptation performance of DA-VSN and PixMatch, both approaches are inferior in video segmentation as compared to TPS. As regard to DA-VSN, in spite of its excellence in retaining temporal consistency, the learnt network using DA-VSN produces less accurate segmentation (e.g. sidewalk in Fig.~\ref{fig:result}). Such outcome demonstrates the superiority of consistency training over adversarial learning in minimizing empirical error. As for PixMatch, we notice that the performance of learnt network with PixMatch is unsatisfying on retaining temporal consistency, which corroborates the necessity of introducing cross-frame augmentation in consistency training. Based on the observation of qualitative results, we conclude that TPS performs better in either keeping temporal consistency or producing accurate segmentation, which is in accordance with the quantitative result 
in Table~1. 

\subsection{Ablation Studies}
We perform extensive ablation studies to better understand why TPS can achieve superior performance on video adaptive semantic segmentation.  All the ablation studies are performed on the benchmark of SynthiaSeq$\rightarrow$Cityscapes, where TPS achieves a mIoU of 53.8\% under the default setting. We present complete ablation results and concrete analysis, including the propagation interval $\eta$ in Eq.~\ref{eq:aug} the confidence threshold $\tau$ in Eq.~\ref{eq:pl}, and the balancing parameter $\lambda_{T}$ in Eq.~\ref{eq:tps}.

\renewcommand\arraystretch{1.36}
\begin{table}[!ht]
\label{tab:interval}
\centering
\caption{Results of TPS with different propagation interval $\eta$: TPS achieves the best performance when $\eta=1$. For the classes of small objects (\textit{e.g.}, pole, light, sign, person and rider), the performance may suffer from warping error while increasing $\eta$
}
\begin{scriptsize}
\begin{tabular}{p{1cm}|*{11}{p{0.8cm}}|p{0.8cm}}
 \toprule
 \multicolumn{13}{c}{\textbf{SYNTHIA-Seq~$\rightarrow$~Cityscapes-Seq}} \\
 \midrule
 $\eta$  &{road} &{side.} &{buil.} &{\textbf{pole}} &{\textbf{light}} &{\textbf{sign}} &{vege.} &{sky} &{\textbf{pers.}} &{\textbf{rider}} &{{car}} &mIoU \\
 \midrule
 3 &88.9 &49.5 &75.4 &23.4 &14.1 &31.6 &73.5 &61.0 &54.3 &15.2 &82.2 &51.7 \\
 2 &91.2 &52.1 &74.9 &19.2 &14.2 &31.7 &71.1 &61.6 &55.9 &19.0 &84.5 &52.3 \\
 1 &91.2 &53.7 &74.9 &\textbf{24.6} &\textbf{17.9} &\textbf{39.3} &68.1 &59.7 &\textbf{57.2} &\textbf{20.3} &{84.5} &\textbf{53.8} \\
\bottomrule
\end{tabular}
\end{scriptsize}
\end{table}

\paragraph{\textbf{Propagation Interval.}} 
The propagation interval $\eta$ in Eq.~\ref{eq:aug} represents temporal variance between previous and current frames in cross-frame augmentation. We note that increasing propagation interval $\eta$ will expand temporal variance and thus enrich cross-frame augmentation. We present our result of the ablation study on propagation interval 
in Table~3. Despite all results surpassing current methods 
in Table~1, we note that the network suffers from a performance drop while increasing propagation interval, especially on the segmentation of small objects, which can be ascribed to the increased warping error caused by propagating video prediction with optical flow.  

\paragraph{\textbf{Confidence Threshold.}} 
The confidence threshold $\tau$ in Eq.~\ref{eq:pl} is closely related to the quality of the produced pseudo labels. A common solution is to set a confidence threshold $\tau\in(0,1)$ to filter out the low-confident predictions while pseudo labelling whereas retains high-confident ones. Despite its potential effectiveness in retaining the quality of pseudo labels, the consistency training in TPS tends to suffer from the inherent class-imbalance distribution in a real-world dataset (target domain), which prevents the network to produce high confidence scores for some hard-to-transfer classes. To explore the effect of the threshold $\tau$ on the performance of TPS, we perform relevant experiments and present our results 
in Table~4. We note that the best result is obtained when $\tau$ is set to 0. We highlight that the segmentation on hard-to-transfer classes in our task (e.g. pole, light, sign and rider) suffers from performance drops as expected while confidence threshold $\tau$ is adopted when pseudo labeling. 

\renewcommand\arraystretch{1.36}
\begin{table}[!ht]
\label{tab:threshold}
\centering
\caption{Results of TPS with different confidence threshold $\tau$: The best result is obtained when $\tau=0$. It can be noticed that the hard-to-transfer classes (\textit{e.g.}, pole, light, sign, rider) experience performance drop while setting $\tau>0$ to filter out low-confident predictions when pseudo labeling }
\begin{scriptsize}
\begin{tabular}{p{1cm}|*{11}{p{0.7cm}}|p{0.7cm}}
 \toprule
 \multicolumn{13}{c}{\textbf{SYNTHIA-Seq~$\rightarrow$~Cityscapes-Seq}} \\
 \midrule
 $\tau$  &{road} &{side.} &{buil.} &{{\textbf{pole}}} &{\textbf{light}} &{\textbf{sign}} &{vege.} &{sky} &{pers.} &{\textbf{rider}} &{car} &mIoU \\
 \midrule
 0.50 &91.1& 54.0& 76.5& 23.7& 14.1& 34.5& 71.7& 59.7& 56.4& 18.5& 84.3 &53.1 \\
 0.25 &88.1& 48.1& 77.2& 21.2& 16.2& 38.5& 74.1& 64.1& 57.6& 17.4& 86.0&53.5 \\
 0.00 & 91.2& 53.7& 74.9& \textbf{24.6}& \textbf{17.9}& \textbf{39.3}& 68.1& 59.7&57.2 & \textbf{20.3} &84.5 & \textbf{53.8} \\
\bottomrule
\end{tabular}
\end{scriptsize}
\end{table}

\renewcommand\arraystretch{1.2}
\begin{table}[!ht]
\label{tab:lambda}
\centering
\caption{Parameter analysis on the balancing weight $\lambda_{T}$. We observe that either prioritizing training process on source or target domain degrades the segmentation performance}
\begin{scriptsize}
\begin{tabular*}{0.7\textwidth}{@{\extracolsep{\fill}}p{2cm}|*{6}{p{1cm}}}
 \toprule
\multicolumn{7}{c}{\textbf{SYNTHIA-Seq~$\rightarrow$~Cityscapes-Seq}}  \\
   \midrule
 $\lambda_{T}$&\multicolumn{1}{m{1cm}<{\centering}}{0.1} &\multicolumn{1}{m{1cm}<{\centering}}{0.2} &\multicolumn{1}{m{1cm}<{\centering}}{0.5} &\multicolumn{1}{m{1cm}<{\centering}}{1.0}  &\multicolumn{1}{m{1cm}<{\centering}}{1.5} &\multicolumn{1}{m{1cm}<{\centering}}{2.0} \\
\midrule
 \textbf{TPS (Ours)} &\multicolumn{1}{m{1cm}<{\centering}}{50.0} &\multicolumn{1}{m{1cm}<{\centering}}{51.2} &\multicolumn{1}{m{1cm}<{\centering}}{52.6} &\multicolumn{1}{m{1cm}<{\centering}}{\textbf{53.8}}  &\multicolumn{1}{m{1cm}<{\centering}}{53.4} &\multicolumn{1}{m{1cm}<{\centering}}{53.3} \\
\bottomrule
\end{tabular*}
\end{scriptsize}
\end{table}

\paragraph{\textbf{Balancing Weight.}} 
The balancing weight $\lambda_{T}$ in Eq.~\ref{eq:tps} contributes to our solution by balancing training process between source and target domain nicely. Both supervised learning in source domain with dense annotations and consistency training in target domain should be taken good care of. We present our result of ablation study on $\lambda_{T}$ in Table~5. As presented in Table~5, the best result is retrieved while $\lambda_{T}$ is set to 1.0. We can observe that all results of various $\lambda_{T}$ surpass the result of previous work DA-VSN (achieved a mIoU of 49.5 in Table~1) on the benchmark of SYNTHIA-Seq$\rightarrow$Cityscapes-Seq, which demonstrates the superiority of consistency training in TPS.

\begin{figure}[ht]
\centering
\begin{minipage}[c]{0.49\linewidth}
\centering\includegraphics[width=1.0\linewidth]{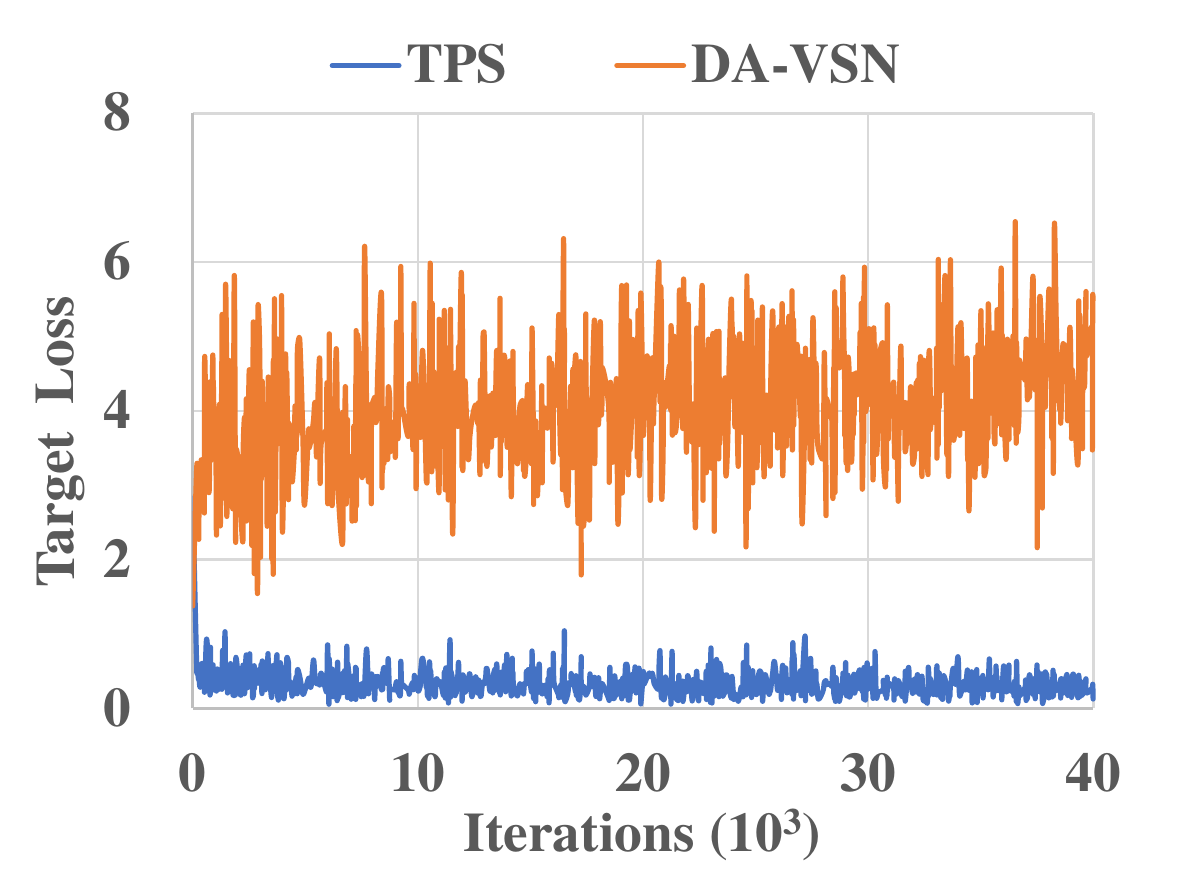}
\end{minipage}
\begin{minipage}[c]{0.49\linewidth}
\centering\includegraphics[width=1.0\linewidth]{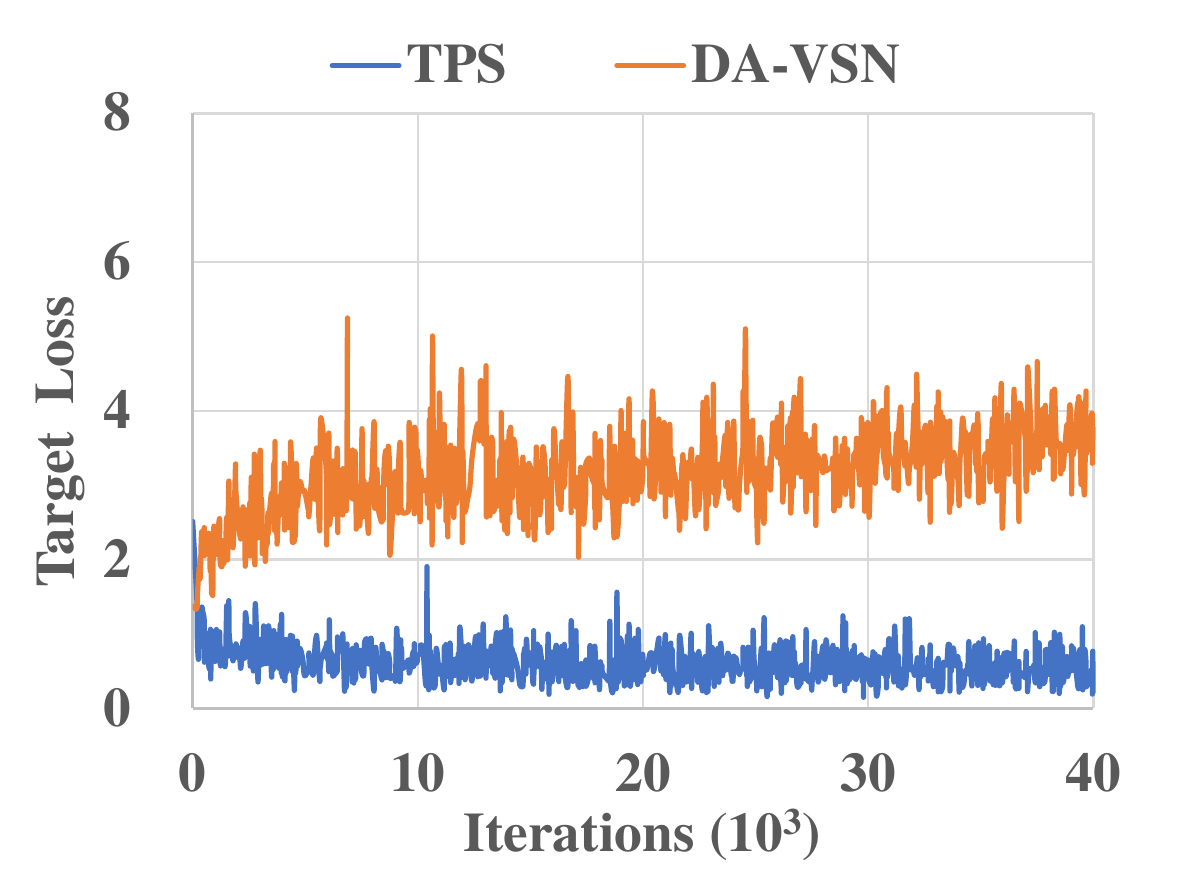}
\end{minipage}
\begin{minipage}[c]{0.49\linewidth}
\centering {(a) SYNTHIA-Seq~$\rightarrow$~Cityscapes-Seq}
\end{minipage}
\begin{minipage}[c]{0.49\linewidth}
\centering {(b) VIPER~$\rightarrow$~Cityscapes-Seq}
\end{minipage}
\caption{Target losses from TPS and DA-VSN for two domain adaptation benchmarks: (a) SYNTHIA-Seq~$\rightarrow$~Cityscapes-Seq and (b) VIPER~$\rightarrow$~Cityscapes-Seq. We point out that the degradation of target loss in TPS is more stable than that in DA-VSN for both two benchmarks. Best viewed in color.}
\label{fig:loss}
\end{figure}

\begin{figure}[ht]
\centering
\begin{minipage}[c]{0.42\linewidth}
\centering {Source Only} 
\end{minipage}
\begin{minipage}[c]{0.42\linewidth}
\centering {PixMatch~\cite{melas2021pixmatch}}
\end{minipage}
\begin{minipage}[c]{0.42\linewidth}
\centering\includegraphics[width=.99\linewidth]{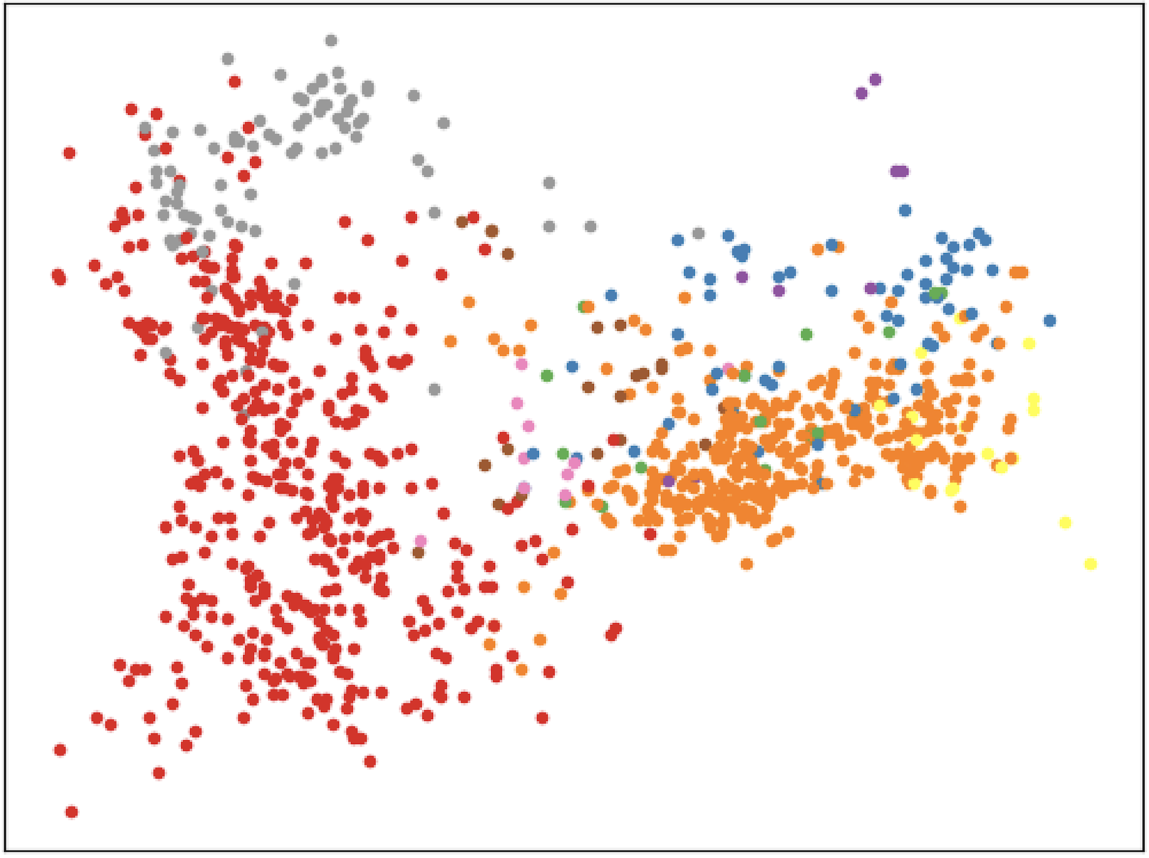}
\end{minipage}\vspace{2pt}
\begin{minipage}[c]{0.42\linewidth}
\centering\includegraphics[width=.99\linewidth]{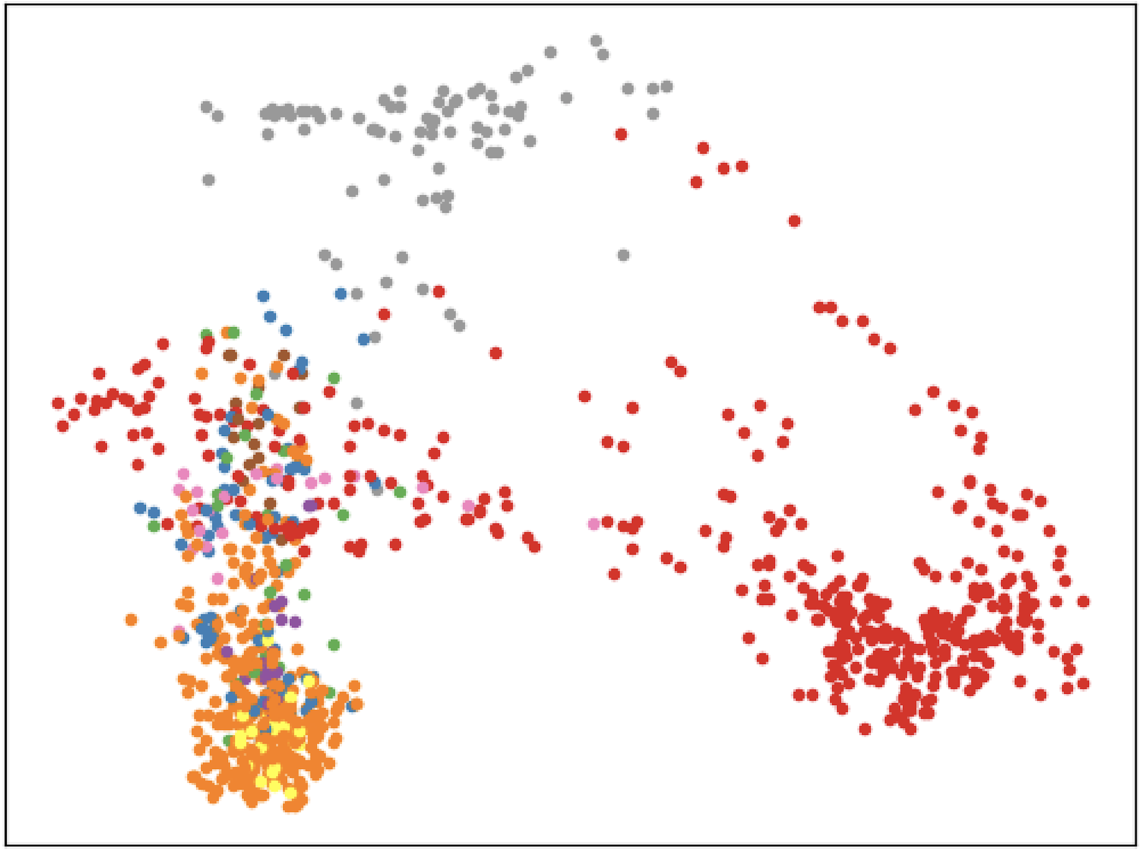}
\end{minipage}
\centering
\begin{minipage}[c]{0.42\linewidth}
\centering {$\sigma_{\text{inter}}^{2}=0.679$, $\sigma_{\text{intra}}^{2}=0.606$}
\end{minipage}
\begin{minipage}[c]{0.42\linewidth}
\centering {$\sigma_{\text{inter}}^{2}=0.719$, $\sigma_{\text{intra}}^{2}=0.541$}
\end{minipage}\vspace{6pt}
\centering
\begin{minipage}[c]{0.42\linewidth}
\centering {DA-VSN~\cite{guan2021domain}} 
\end{minipage}
\begin{minipage}[c]{0.42\linewidth}
\centering {\textbf{TPS (Ours)}}
\end{minipage}
\begin{minipage}[c]{0.42\linewidth}
\centering\includegraphics[width=.99\linewidth]{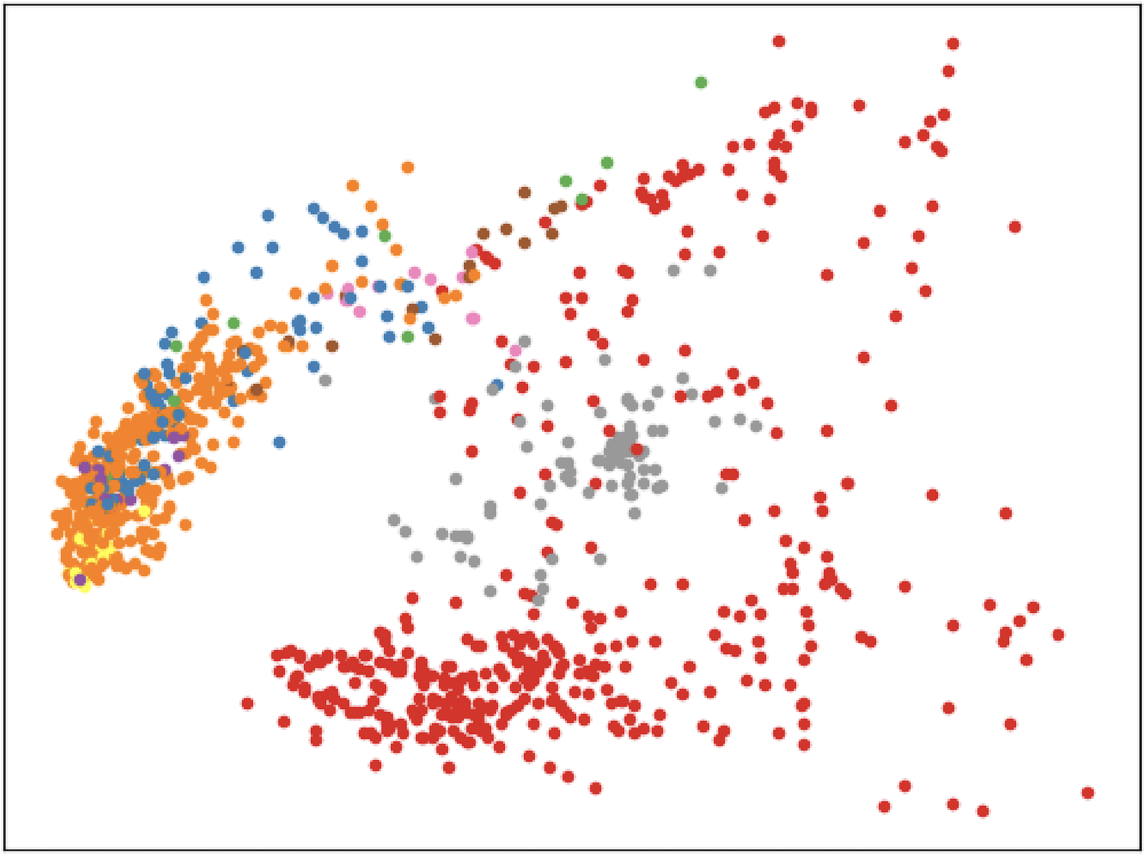}
\end{minipage}
\begin{minipage}[c]{0.42\linewidth}
\centering\includegraphics[width=.99\linewidth]{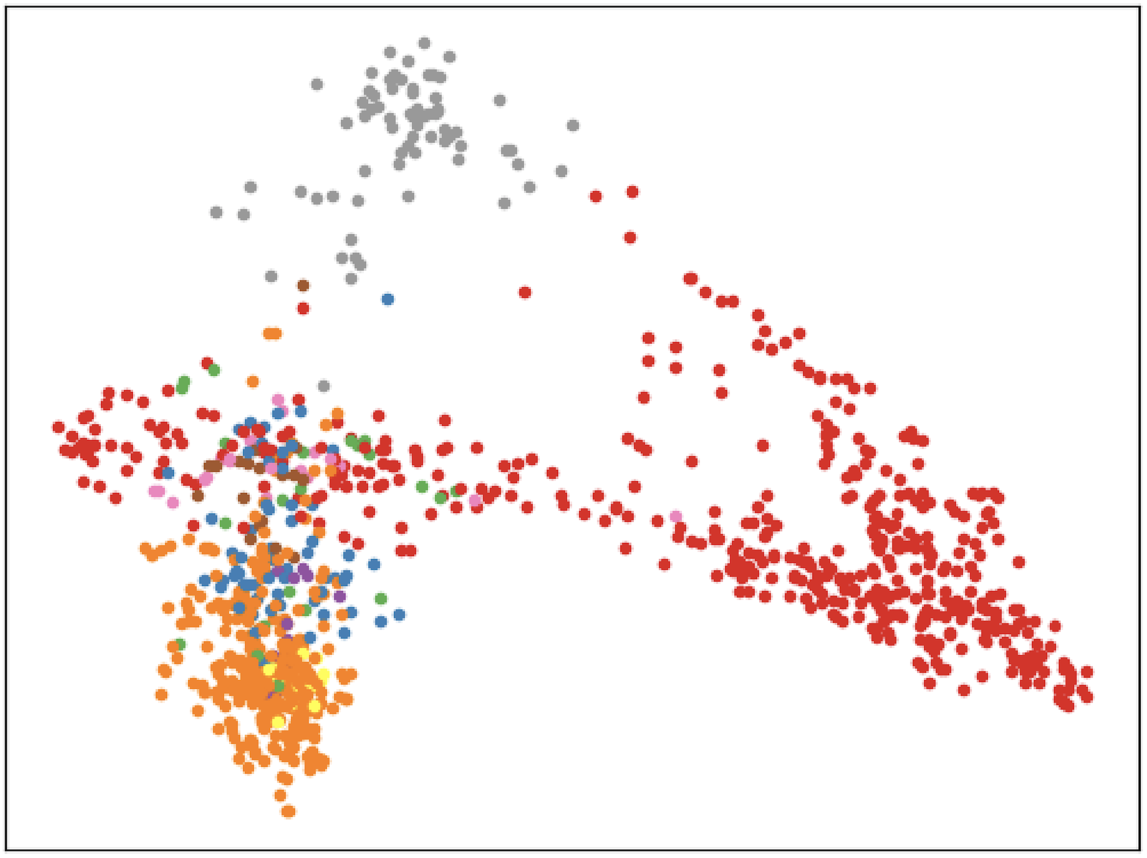}
\end{minipage}
\centering
\begin{minipage}[c]{0.42\linewidth}
\centering {$\sigma_{\text{inter}}^{2}=0.700$, $\sigma_{\text{intra}}^{2}=0.584$}
\end{minipage}
\begin{minipage}[c]{0.42\linewidth}
\centering {$\sigma_{\text{inter}}^{2}=0.740$, $\sigma_{\text{intra}}^{2}=0.527$}
\end{minipage}
\caption{Visualization of temporal feature representations in the target domain via t-SNE~\cite{maaten2008visualizing} (different colors represent different categories): the proposed TPS surpasses Source Only, PixMatch~\cite{melas2021pixmatch} and DA-VSN~\cite{guan2021domain} clearly with higher inter-class variance and lower intra-class variance. Note that we obtain the temporal features by stacking features extracted from two consecutive frames as in~\cite{guan2021domain}, and perform PCA with whitening on the obtained temporal features to retrieve principal components with unit component-wise variances. 
The visualization is based on the domain adaptive video segmentation benchmark SYNTHIA-Seq~$\rightarrow$~Cityscapes-Seq. Best viewed in color.}
\label{fig:tsne}
\end{figure}

\subsection{Discussion}
\paragraph{\textbf{Training stability.}} 
To compare the training stability of DA-VSN with TPS on two benchmarks, we visualize the target-domain training processes of both DA-VSN and TPS by calculating the target losses for every 20 iterations. As illustrated in Fig.~\ref{fig:loss}, the decay of target loss with TPS is much less noisy than in DA-VSN, along with lower empirical error on average in target domain on both benchmarks, indicating the effectiveness of consistency training on the domain adaptive video segmentation task. In contrast, the target loss in DA-VSN degrades more unsteadily and harder to converge due to the adversarial learning module in DA-VSN, and such negative effect is stronger under the scenario of SYNTHIA-Seq$\rightarrow$Cityscapes-Seq. The performance differences between benchmarks can be explained by the fact that SYNTHIA-Seq has larger domain gap with Cityscapes-Seq than VIPER, and we also point out that the notable advance on the benchmark of SYNTHIA-Seq$\rightarrow$Cityscapes-Seq brought by TPS further demonstrates the superiority of consistency training over adversarial learning approach on bridging larger domain gap between different video distribution. This merit is important for real-world applications, since real scenarios could be very different from pre-built synthetic environment.

\paragraph{\textbf{Feature Visualization.}} To delve deeper and investigate on the effectiveness of TPS, we visualize the target-domain video representation with t-SNE~\cite{maaten2008visualizing} presented in Fig.~\ref{fig:tsne}, together with visualization for source only, PixMatch and DA-VSN for comparison. We observe that TPS outperforms source-only training by a large margin, which reveals the outstanding adaptation performance of our consistency-training-based approach. Furthermore, we also spot that TPS surpasses the previous works on domain adaptive video segmentation task by achieving largest inter-class variance while keeping smallest intra-class variance, which is a proper indicator that the upstream class-wise representation from TPS are more distinguishable. 

\renewcommand\arraystretch{1.2}
\begin{table}[!ht]
\label{tab:complement}
\centering
\caption{Complementary Study on TPS: the proposed TPS can be easily integrated with the state-of-the-art work DA-VSN~\cite{guan2021domain} with a clear performance gain over two challenging domain adaptation benchmarks for video segmentation }
\begin{scriptsize}
\begin{tabular*}{\textwidth}{@{\extracolsep{\fill}}p{2cm}|*{3}{p{1.5cm}}|*{3}{p{1.5cm}}}
 \toprule
&\multicolumn{3}{c|}{\textbf{SYNTHIA-Seq~$\rightarrow$~Cityscapes-Seq}}&\multicolumn{3}{c}{\textbf{VIPER~$\rightarrow$~Citycapes-Seq}}  \\
   \midrule
 Method &\multicolumn{1}{m{1.5cm}<{\centering}}{Base} &\multicolumn{1}{m{1.5cm}<{\centering}}{+TPS} &\multicolumn{1}{m{1.5cm}<{\centering}|}{Gain} &\multicolumn{1}{m{1.5cm}<{\centering}}{Base}  &\multicolumn{1}{m{1.5cm}<{\centering}}{+TPS} &\multicolumn{1}{m{1.5cm}<{\centering}}{Gain} \\
\midrule
DA-VSN &\multicolumn{1}{m{1.5cm}<{\centering}}{49.5} &\multicolumn{1}{m{1.5cm}<{\centering}}{55.1} &\multicolumn{1}{m{1.5cm}<{\centering}|}{+5.6} &\multicolumn{1}{m{1.5cm}<{\centering}}{47.8}  &\multicolumn{1}{m{1.5cm}<{\centering}}{50.2} &\multicolumn{1}{m{1.5cm}<{\centering}}{+2.4} \\
\bottomrule
\end{tabular*}
\end{scriptsize}
\label{tab:comp}
\end{table}

\paragraph{\textbf{Complementary Study.}} 
We further conduct experiments to explore if TPS complements the domain adaptive video segmentation network DA-VSN~\cite{guan2021domain} by performing additional cross-frame consistency training on target-domain data. The results of our complementary study are summarized in Table~\ref{tab:comp}. It can be observed that the integration of TPS improves the performance of DA-VSN by a large margin over two benchmarks, indicating that consistency training in TPS complements the adversarial learning in DA-VSN productively. Moreover, TPS complements with DA-VSN~\cite{guan2021domain} by surpassing ``TPS only" (achieved a mIoU of 53.8 and 48.9 
in Table~1 and~2 respectively), which proves that the effects of adversarial learning and consistency training on the domain adaptive video segmentation task are orthogonal.

\section{Conclusion}
This paper proposes a temporal pseudo supervision method that introduces cross-frame augmentation and cross-frame pseudo labeling to address domain adaptive video segmentation from the perspective of consistency training. Specifically, cross-frame augmentation is designed to expand the diversity of image augmentation in traditional consistency training and thus effectively exploit unlabeled target videos. To facilitate consistency training with cross-frame augmentation, cross-frame pseudo labelling provides pseudo supervision from previous video frames for network training fed with augmented current video frames, where the introduction of pseudo labeling encourages the network to output video predictions with high certainty. Comprehensive experiments demonstrate the effectiveness of our method in domain adaption for video segmentation. In the future, we will investigate how the idea of temporal pseudo supervision perform in other video-specific tasks with unlabeled data, such as semi-supervised video segmentation and domain adaptive action recognition.

\section*{Acknowledgement}
This study is supported under the RIE2020 Industry Alignment Fund – Industry Collaboration Projects (IAF-ICP) Funding Initiative, as well as cash and in-kind contribution from Singapore Telecommunications Limited (Singtel), through Singtel Cognitive and Artificial Intelligence Lab for Enterprises.

\clearpage
\bibliographystyle{splncs04}
\bibliography{egbib}

\newpage

\vspace{10pt}
\noindent \textbf{A. More Dataset Details}
\vspace{10pt}

$\bullet$ Cityscapes-Seq~\cite{cordts2016cityscapes} is a widely used real dataset that contains 2,975 and 500 video sequences for training and evaluation, respectively. Specifically, each sequence involves 30 consecutive frames with resolution of $1024\times2048$, while only one frame among the sequence is fully annotated.

\vspace{3pt}
$\bullet$ SYNTHIA-Seq~\cite{ros2016synthia} consists of 8,000 simulated video frames with the resolution of $760\times1280$ and pixel-level annotations automatically produced by game engine. Similar to~\cite{guan2021domain}, we evaluate on the 11 classes in common with the Cityscapes-Seq.

\vspace{3pt}
$\bullet$ VIPER~\cite{richter2017playing} contains 133,670 synthesized video frames with the resolution of $1080\times1920$. The full annotations in VIPER are available for all frames, which are collected by a virtual moving object in diverse ambient conditions. Following the setup of~\cite{guan2021domain}, we use the 15 classes in line with Cityscapes-Seq.

\vspace{10pt}
\noindent \textbf{B. More Implementation Details}
\vspace{10pt}

We provide more details here for the image augmentations we use in our experiments. The combination of augmentations for each training sample is selected randomly from the augmentation set, including color jitter (i.e. brightness, contrast, saturation and hue), gaussian blur, random flipping and scaling. For completeness, we listed the detail of the transformations in Table~\ref{tab:augmentation}.

\renewcommand\arraystretch{1.2}
\begin{table}[!ht]
\centering
\caption{List of Data Transformations
}
\begin{scriptsize}
\begin{tabular*}{\textwidth}{@{\extracolsep{\fill}}p{2cm}|p{6cm}|p{1.5cm}}
 \toprule
  \multicolumn{1}{c}{Transformation} & 
  \multicolumn{1}{c}{Description} &
  \multicolumn{1}{c}{Range} \\
  \midrule
 \multicolumn{1}{m{2cm}<{\centering}}{Brightness} & \multicolumn{1}{m{6cm}<{\centering}}{Adjust the brightness of the image} & \multicolumn{1}{m{1.5cm}<{\centering}}{[0.2, 1.8]}\\
 \multicolumn{1}{m{2cm}<{\centering}}{Contrast} & \multicolumn{1}{m{6cm}<{\centering}}{Control the contrast of the image} & \multicolumn{1}{m{1.5cm}<{\centering}}{[0.2, 1.8]}\\
 \multicolumn{1}{m{2cm}<{\centering}}{Saturation} & \multicolumn{1}{m{6cm}<{\centering}}{Adjust the saturation of the image} & \multicolumn{1}{m{1.5cm}<{\centering}}{[0.2, 1.8]}\\
 \multicolumn{1}{m{2cm}<{\centering}}{Hue} & \multicolumn{1}{m{6cm}<{\centering}}{Adjust hue of image by shifting RGB channels } & \multicolumn{1}{m{1.5cm}<{\centering}}{[0.8, 1.2]}\\
 \multicolumn{1}{m{2cm}<{\centering}}{Gaussian Blur} & \multicolumn{1}{m{6cm}<{\centering}}{Adapt Gaussian Blur to the image} & \multicolumn{1}{m{1.5cm}<{\centering}}{\{5, 7, 9\}}\\
 \multicolumn{1}{m{2cm}<{\centering}}{Horizontal Flip} & \multicolumn{1}{m{6cm}<{\centering}}{Flip image and label horizontally} & \multicolumn{1}{m{1.5cm}<{\centering}}{-}\\
 \multicolumn{1}{m{2cm}<{\centering}}{Rescale} & \multicolumn{1}{m{6cm}<{\centering}}{Rescale the size of image} & \multicolumn{1}{m{1.5cm}<{\centering}}{[0.8, 1.2]}\\
\bottomrule
\end{tabular*}
\end{scriptsize}
\label{tab:augmentation}
\end{table}

\begin{figure}[!ht]
\vspace{20pt}
\centering
\begin{minipage}[c]{0.03\linewidth}
\centering {\rotatebox{90}{Frames}} 
\end{minipage}
\vspace{2pt}
\begin{minipage}[c]{0.31\linewidth}
\centering\includegraphics[width=.99\linewidth]{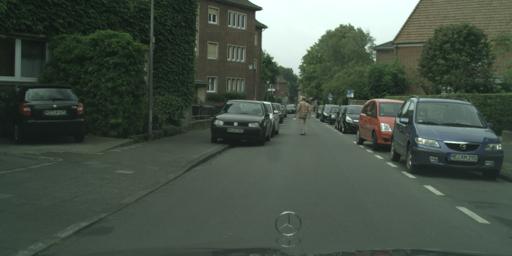}
\end{minipage}
\begin{minipage}[c]{0.31\linewidth}
\centering\includegraphics[width=.99\linewidth]{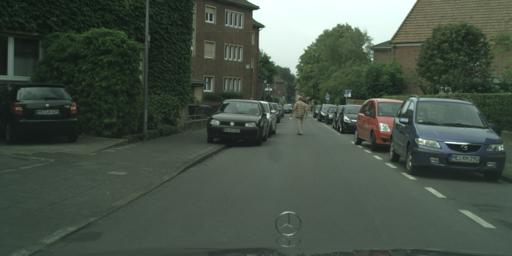}
\end{minipage}
\begin{minipage}[c]{0.31\linewidth}
\centering\includegraphics[width=.99\linewidth]{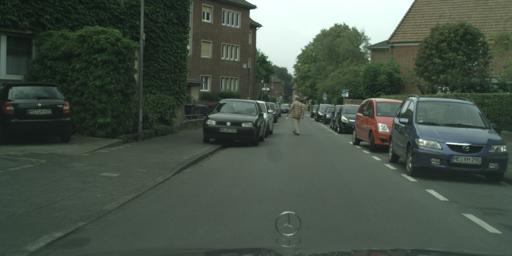}
\end{minipage}
\vspace{2pt}
\centering
\begin{minipage}[c]{0.03\linewidth}
\centering {\rotatebox{90}{GT}} 
\end{minipage}
\begin{minipage}[c]{0.31\linewidth}
\centering\includegraphics[width=.99\linewidth]{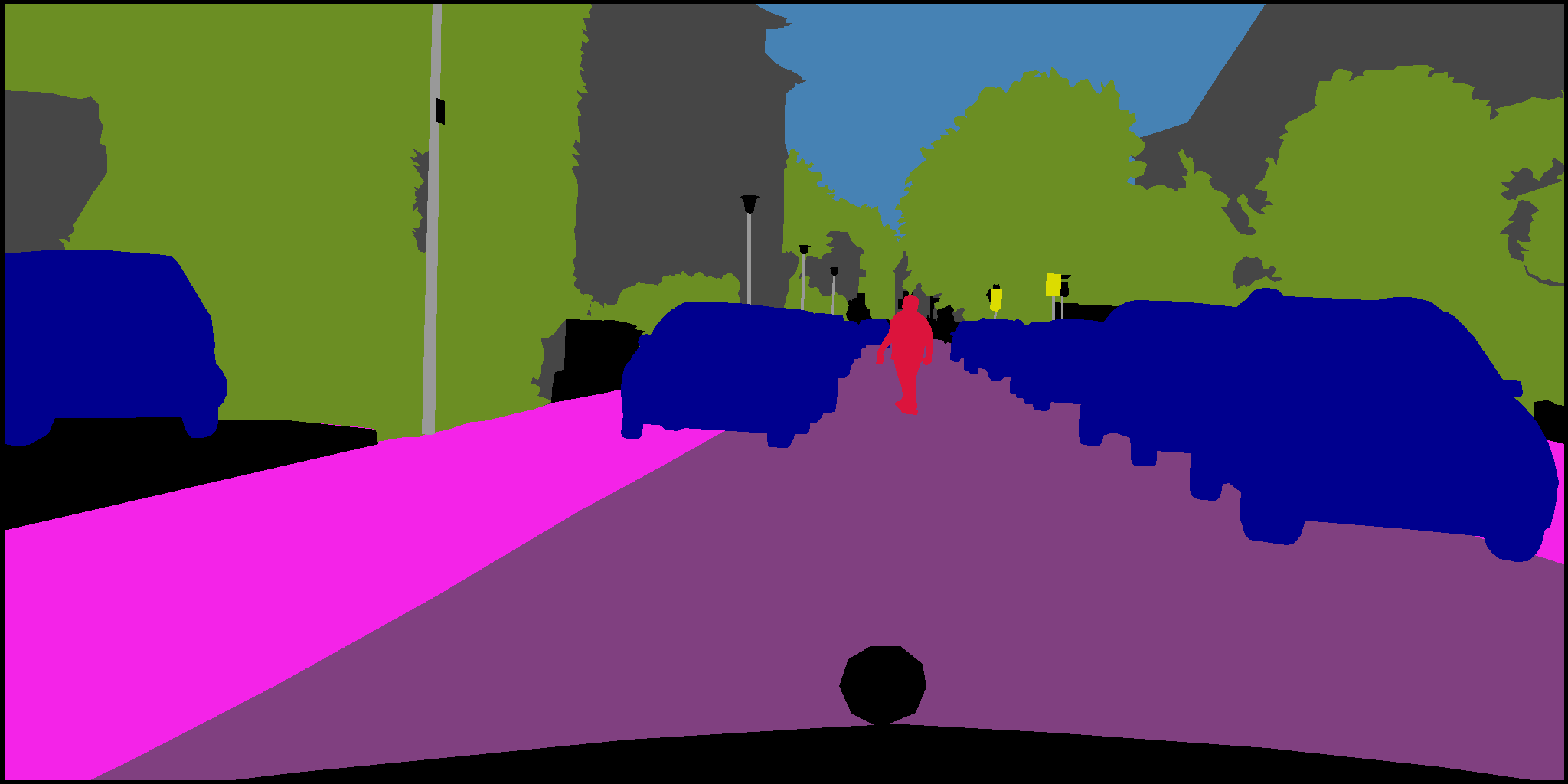}
\end{minipage}
\begin{minipage}[c]{0.31\linewidth}
\centering\includegraphics[width=.99\linewidth]{images/results/munster_000036_000019/munster_000036_000019_gt.png}
\end{minipage}
\begin{minipage}[c]{0.31\linewidth}
\centering\includegraphics[width=.99\linewidth]{images/results/munster_000036_000019/munster_000036_000019_gt.png}
\end{minipage}
\vspace{2pt}
\centering
\begin{minipage}[c]{0.03\linewidth}
\centering {\rotatebox{90}{Source Only}} 
\end{minipage}
\begin{minipage}[c]{0.31\linewidth}
\centering\includegraphics[width=.99\linewidth]{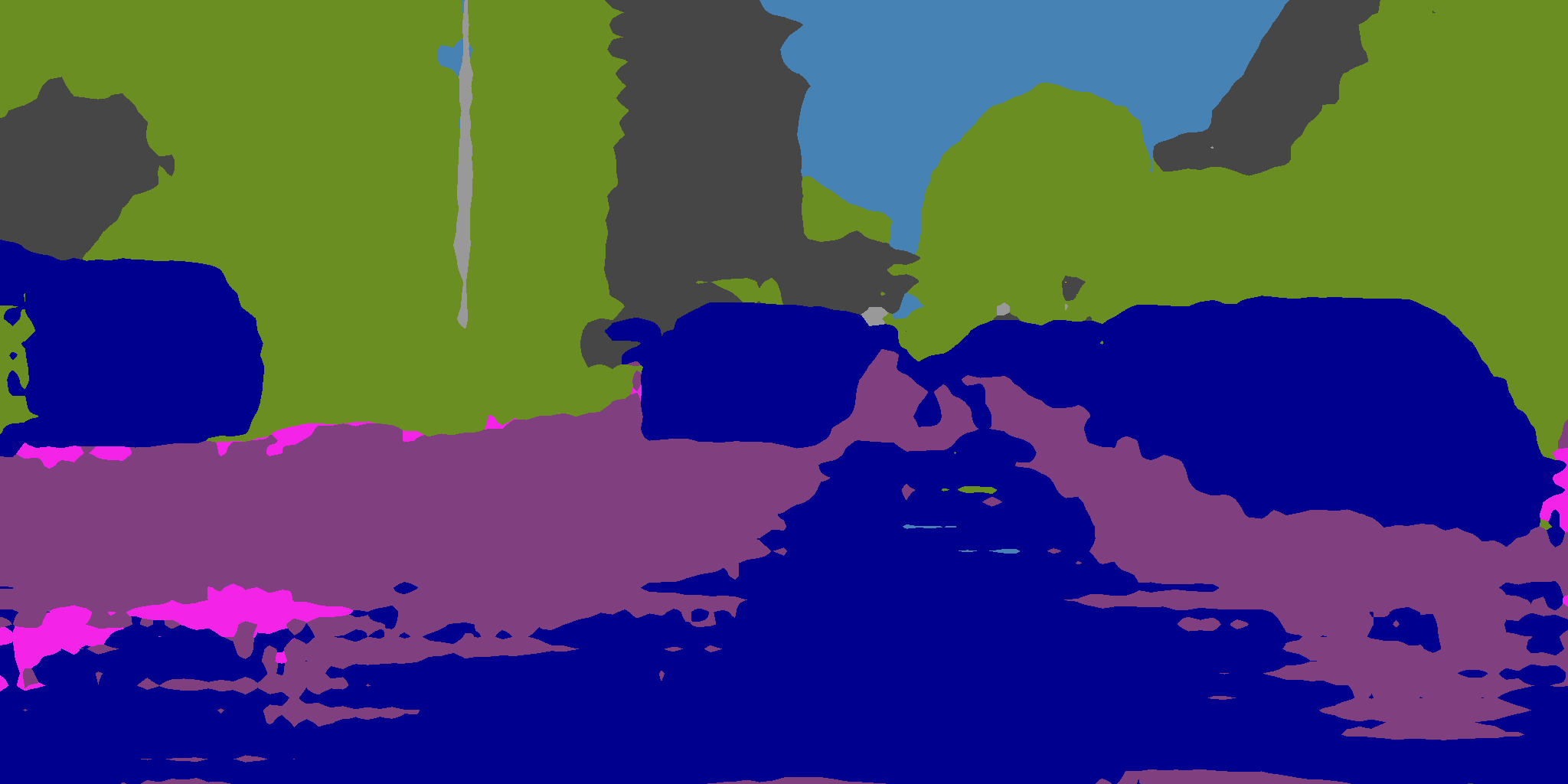}
\end{minipage}
\begin{minipage}[c]{0.31\linewidth}
\centering\includegraphics[width=.99\linewidth]{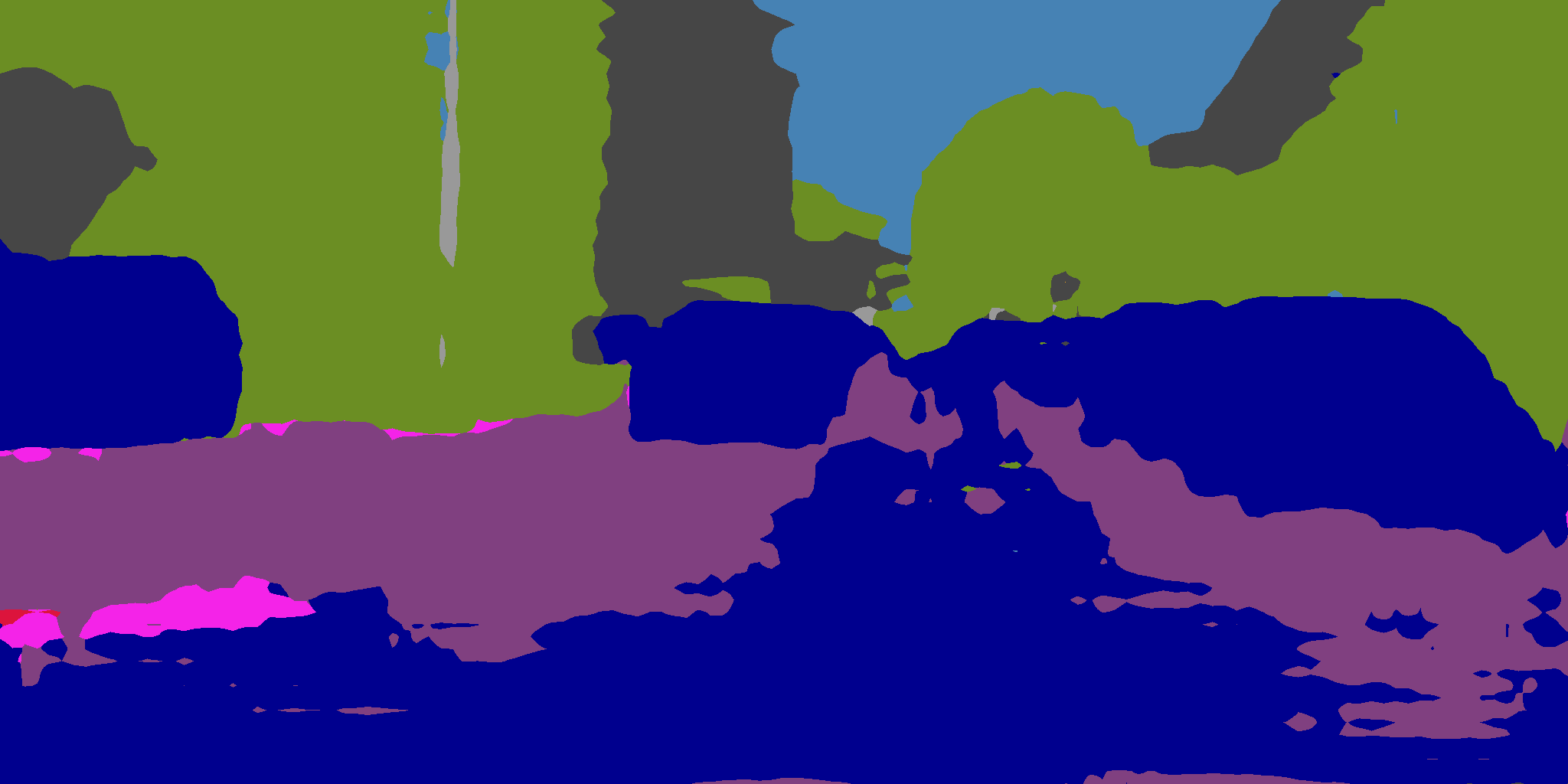}
\end{minipage}
\begin{minipage}[c]{0.31\linewidth}
\centering\includegraphics[width=.99\linewidth]{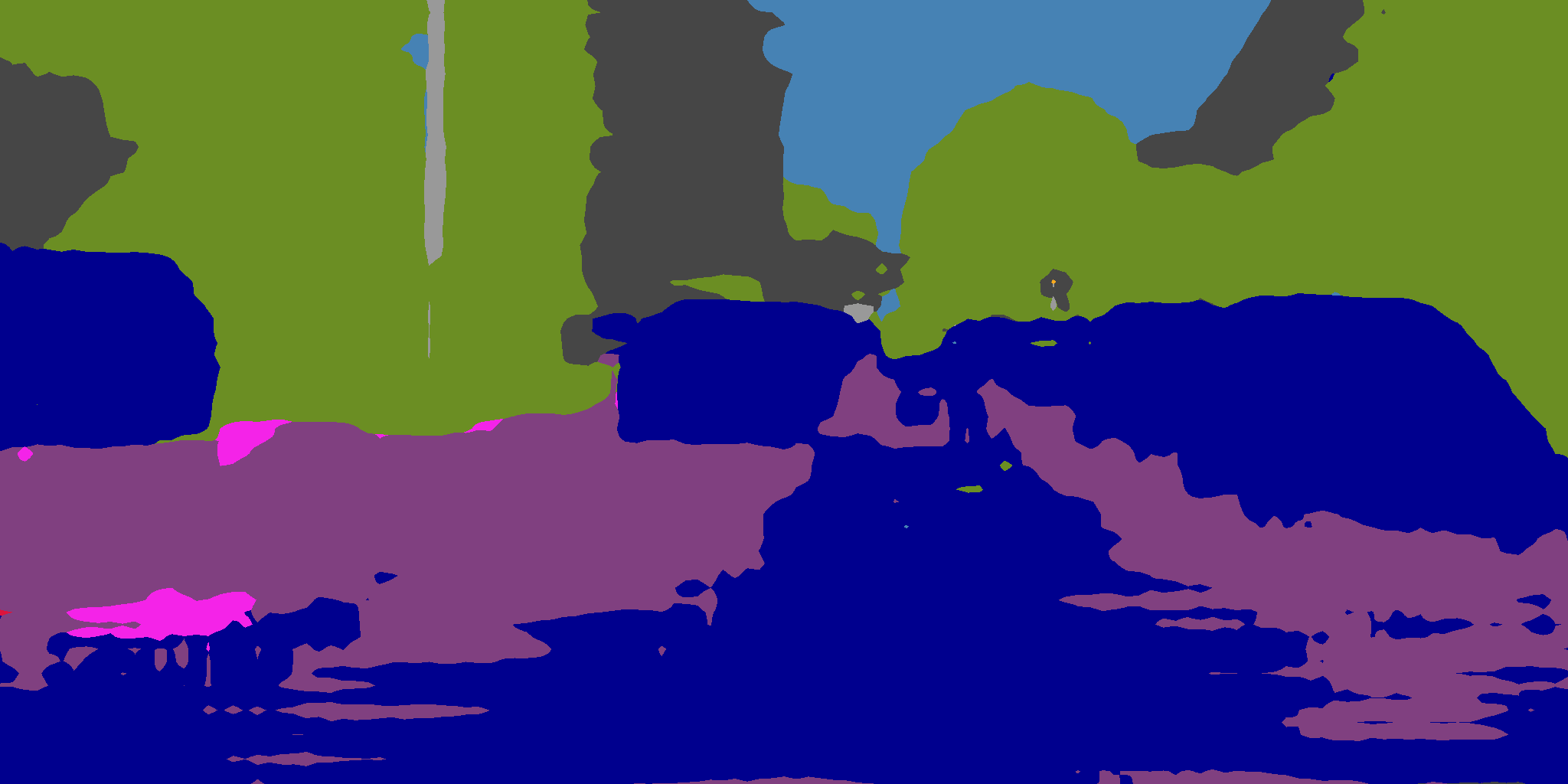}
\end{minipage}
\vspace{2pt}
\centering
\begin{minipage}[c]{0.03\linewidth}
\centering {\rotatebox{90}{DA-VSN}} 
\end{minipage}
\begin{minipage}[c]{0.31\linewidth}
\centering\includegraphics[width=.99\linewidth]{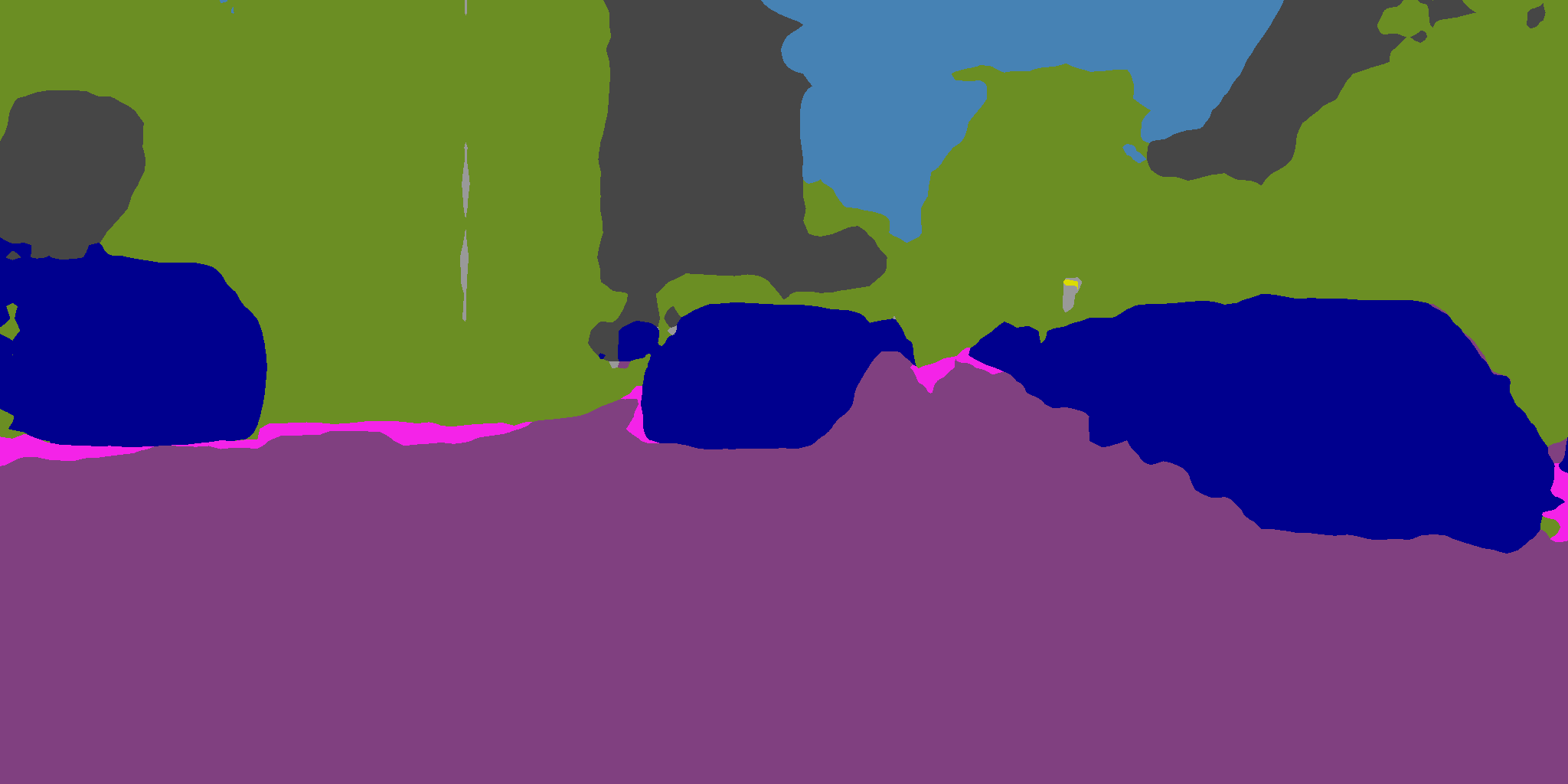}
\end{minipage}
\begin{minipage}[c]{0.31\linewidth}
\centering\includegraphics[width=.99\linewidth]{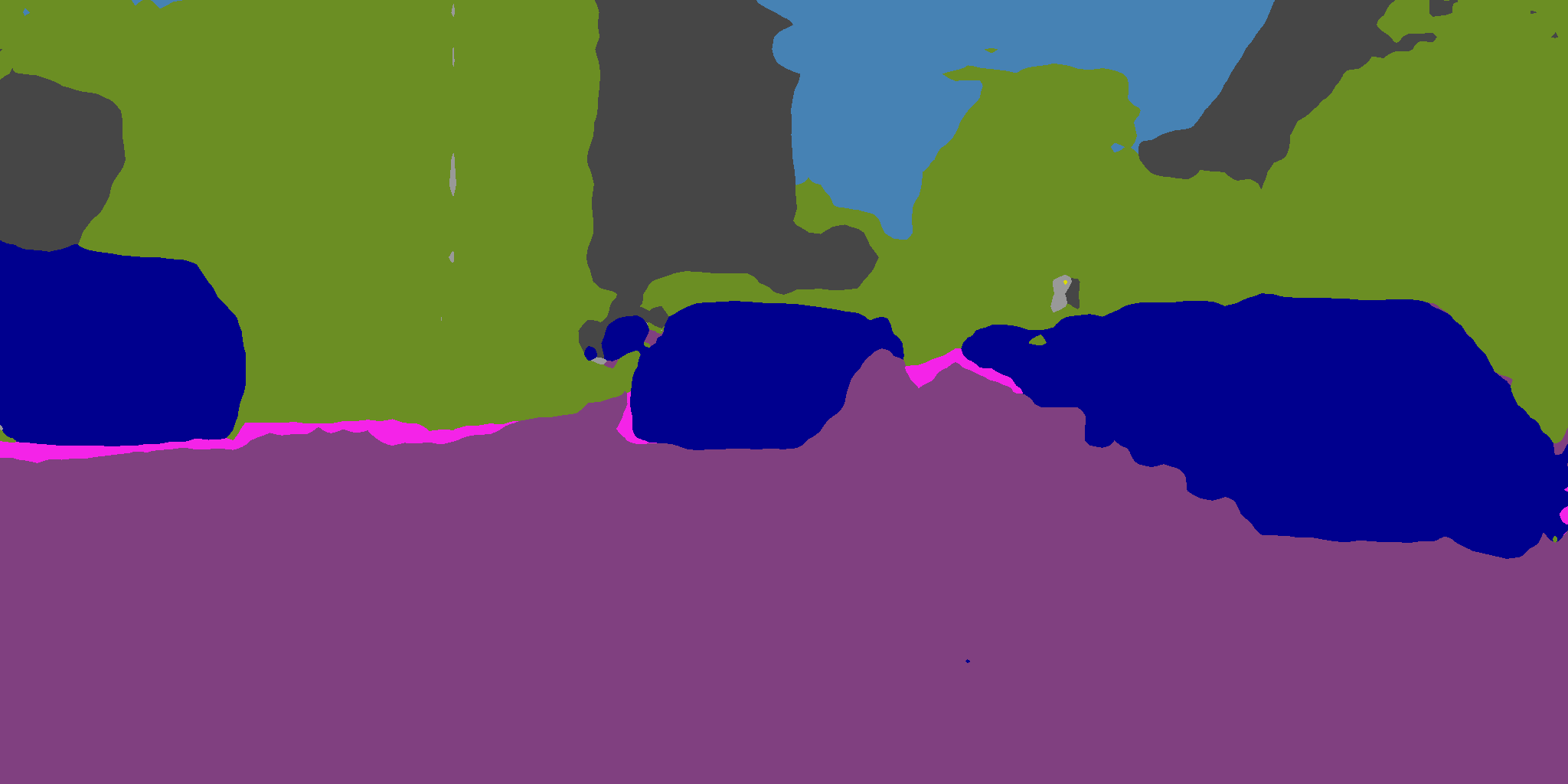}
\end{minipage}
\begin{minipage}[c]{0.31\linewidth}
\centering\includegraphics[width=.99\linewidth]{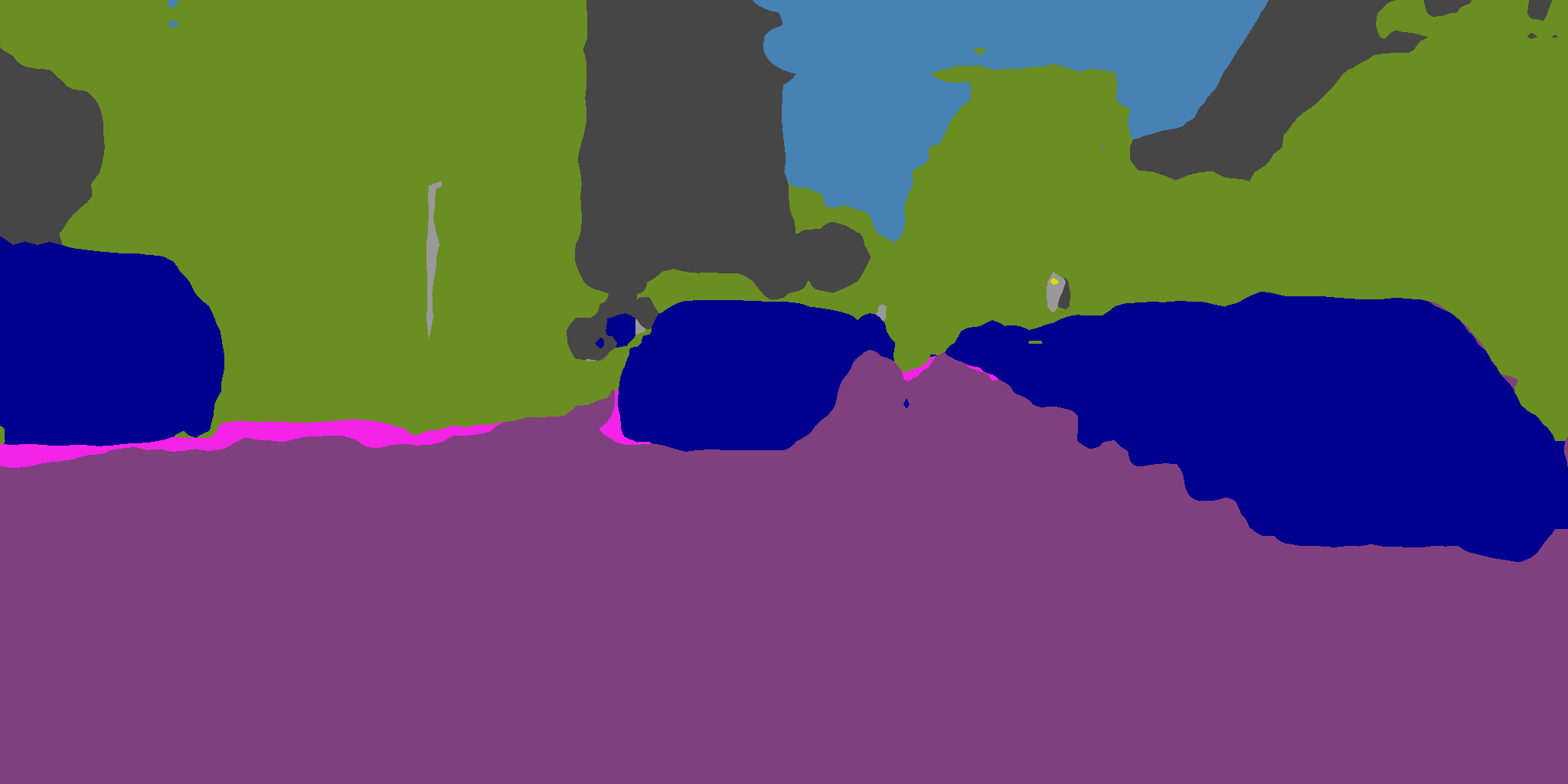}
\end{minipage}
\vspace{2pt}
\centering
\begin{minipage}[c]{0.03\linewidth}
\centering {\rotatebox{90}{PixMatch}} 
\end{minipage}
\begin{minipage}[c]{0.31\linewidth}
\centering\includegraphics[width=.99\linewidth]{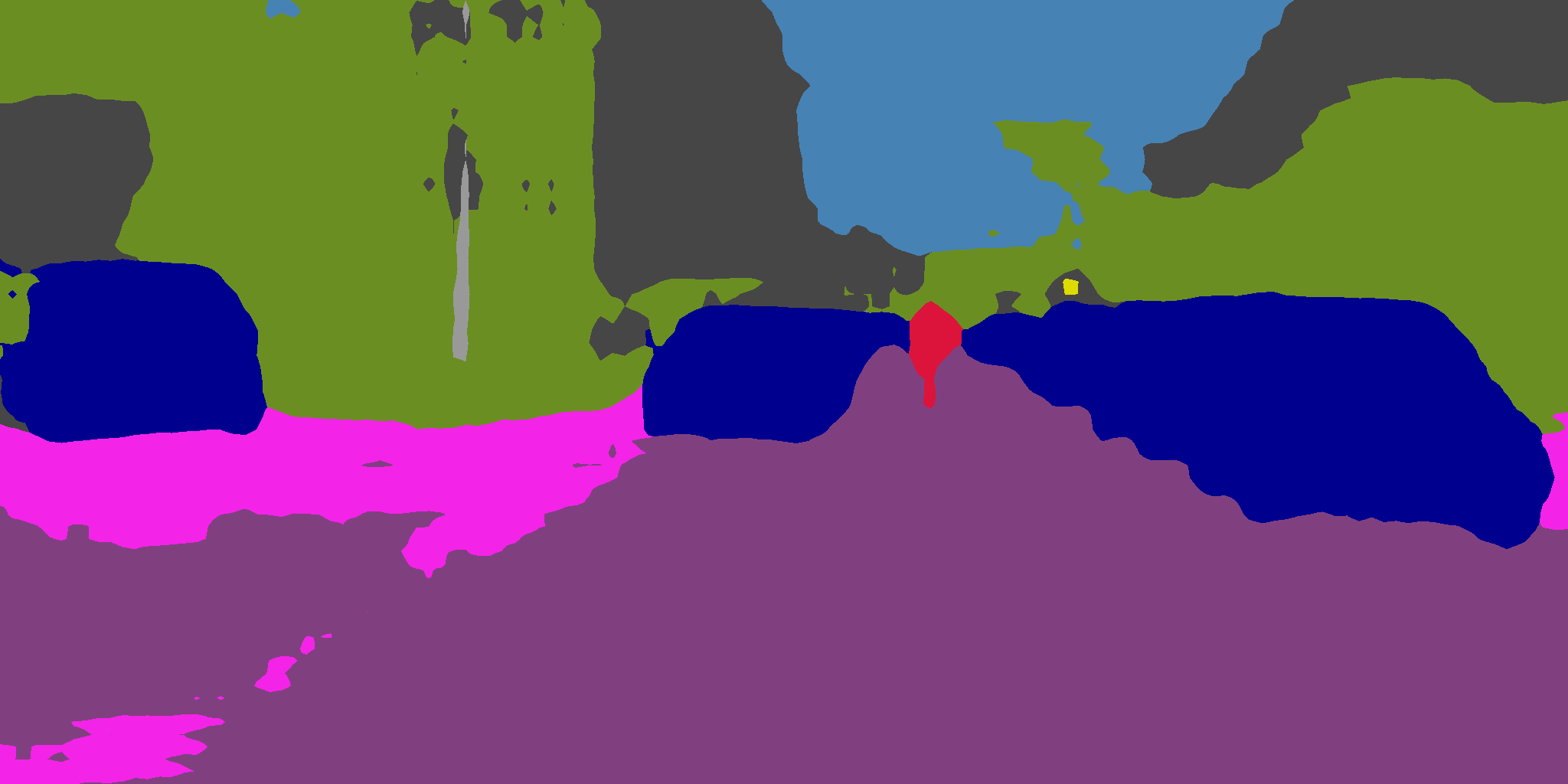}
\end{minipage}
\begin{minipage}[c]{0.31\linewidth}
\centering\includegraphics[width=.99\linewidth]{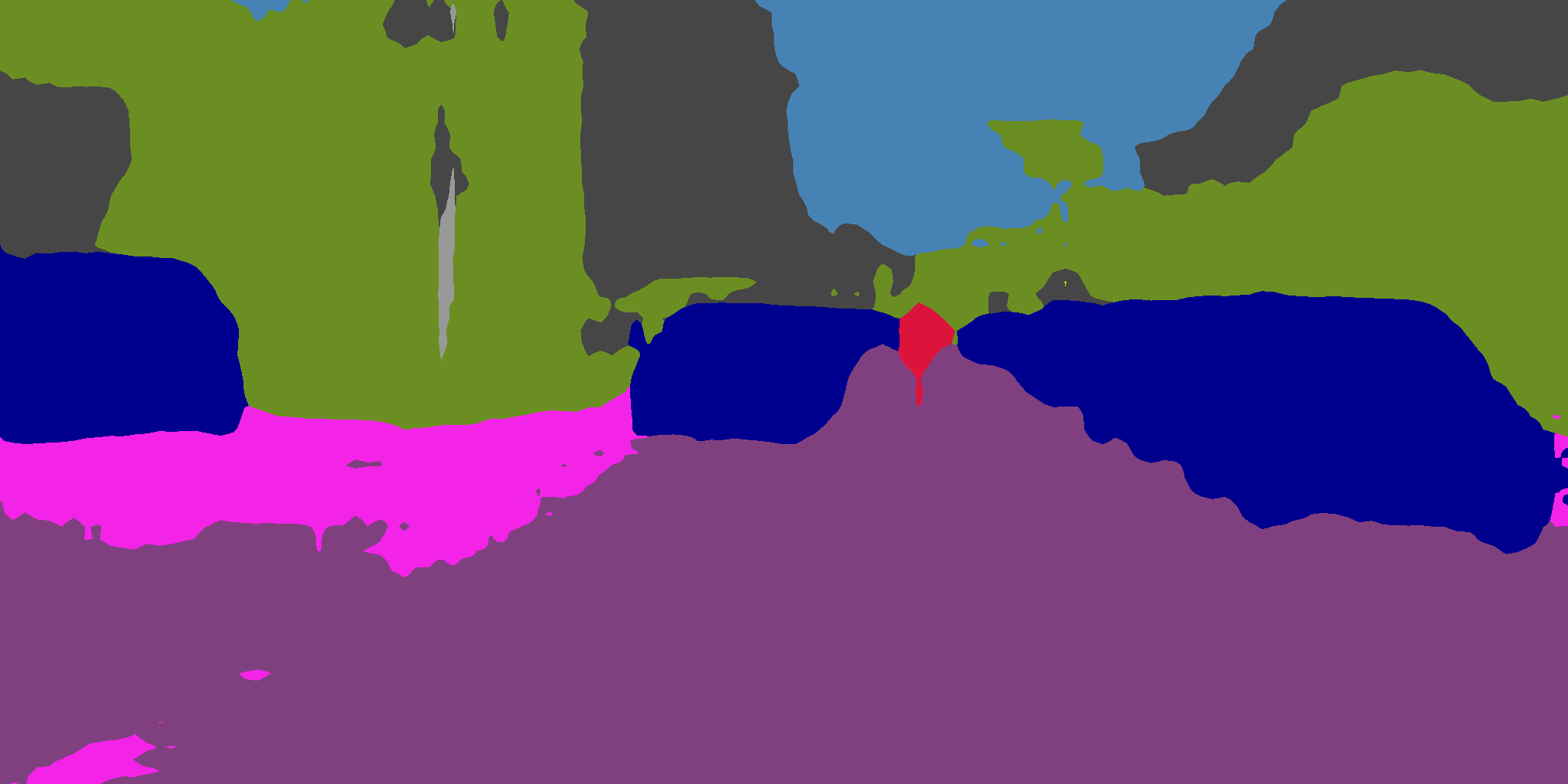}
\end{minipage}
\begin{minipage}[c]{0.31\linewidth}
\centering\includegraphics[width=.99\linewidth]{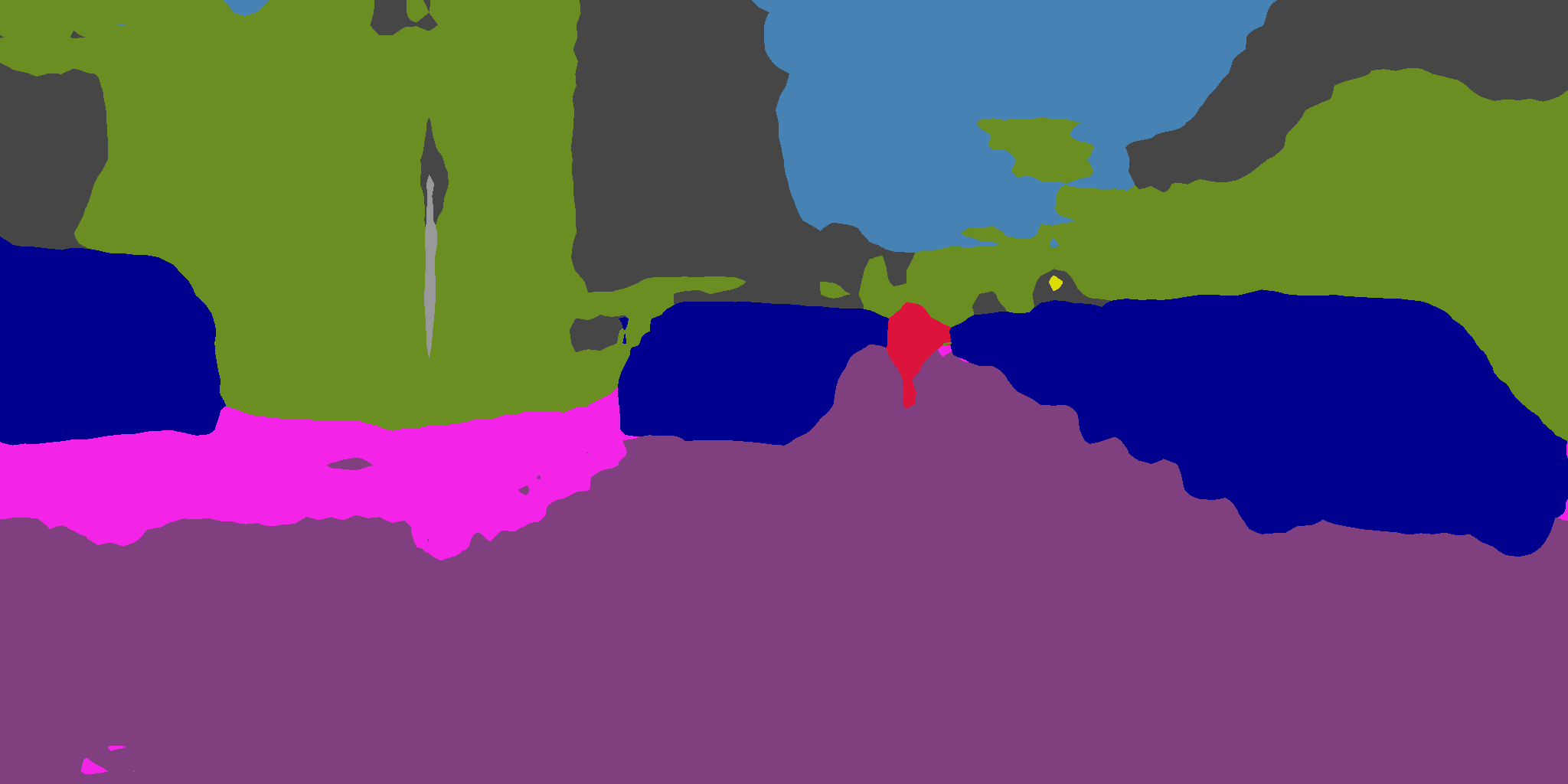}
\end{minipage}
\vspace{2pt}
\centering
\begin{minipage}[c]{0.03\linewidth}
\centering {\rotatebox{90}{\textbf{TPS~(Ours)}}} 
\end{minipage}
\begin{minipage}[c]{0.31\linewidth}
\centering\includegraphics[width=.99\linewidth]{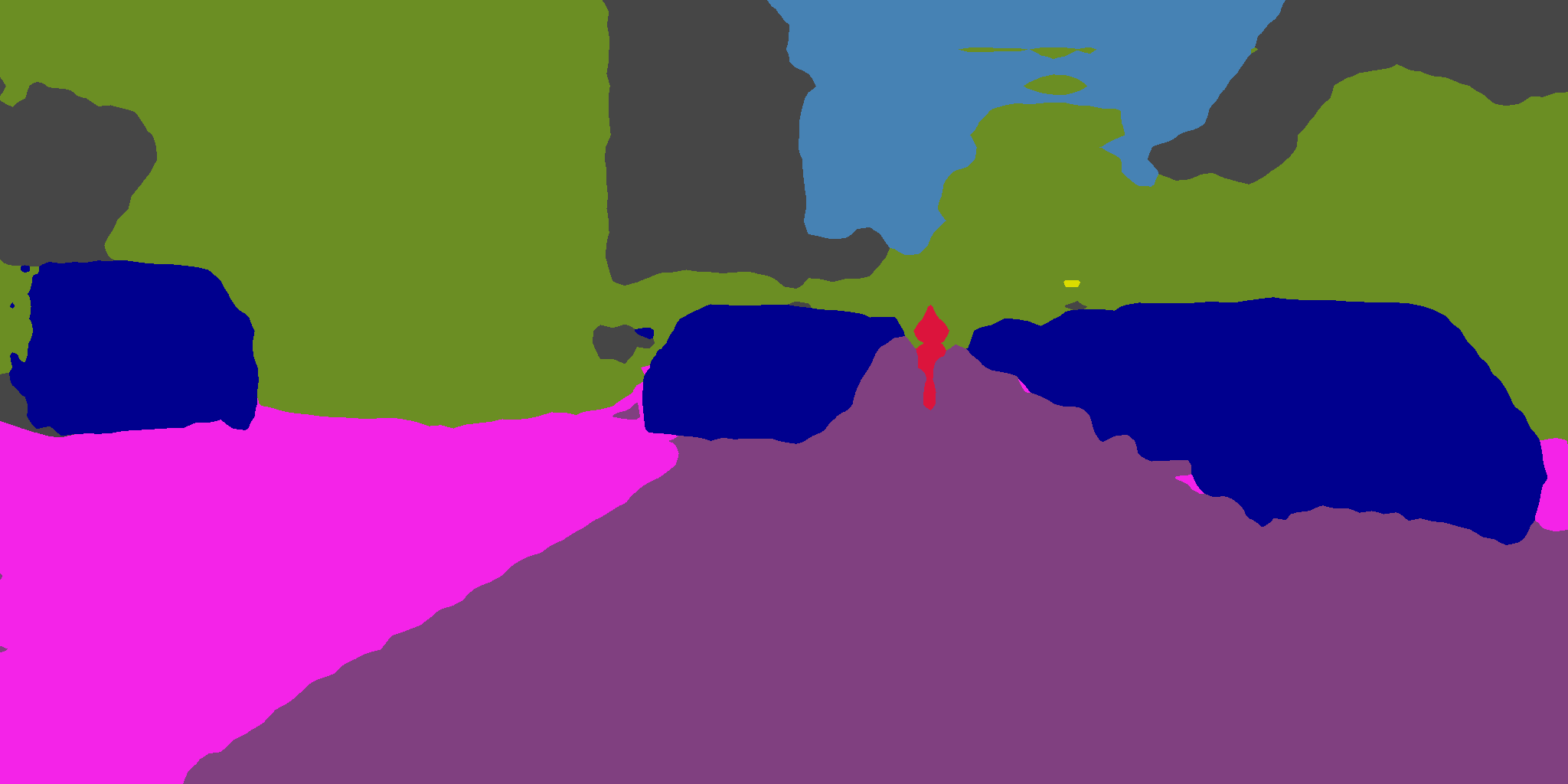}
\end{minipage}
\begin{minipage}[c]{0.31\linewidth}
\centering\includegraphics[width=.99\linewidth]{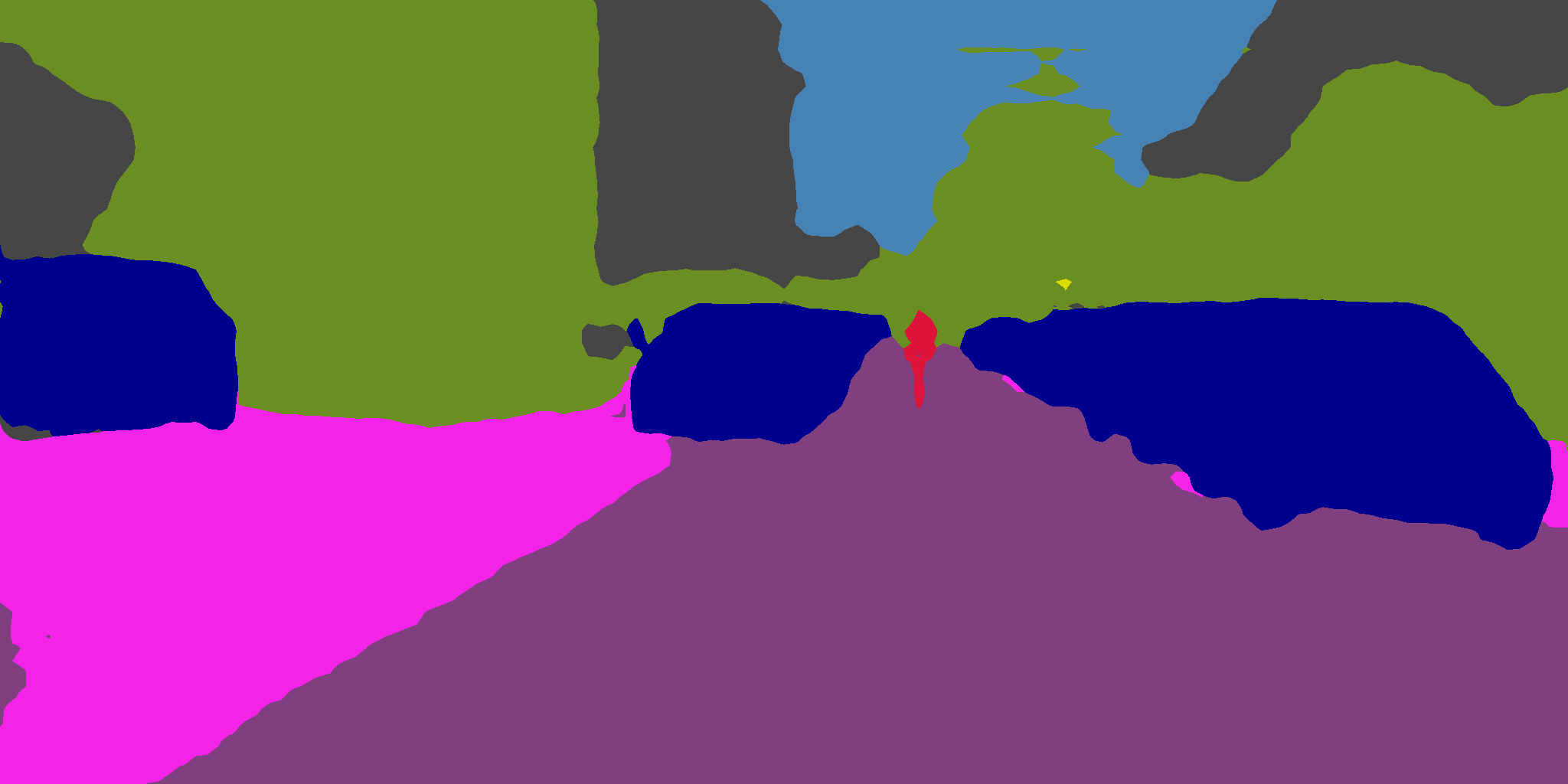}
\end{minipage}
\begin{minipage}[c]{0.31\linewidth}
\centering\includegraphics[width=.99\linewidth]{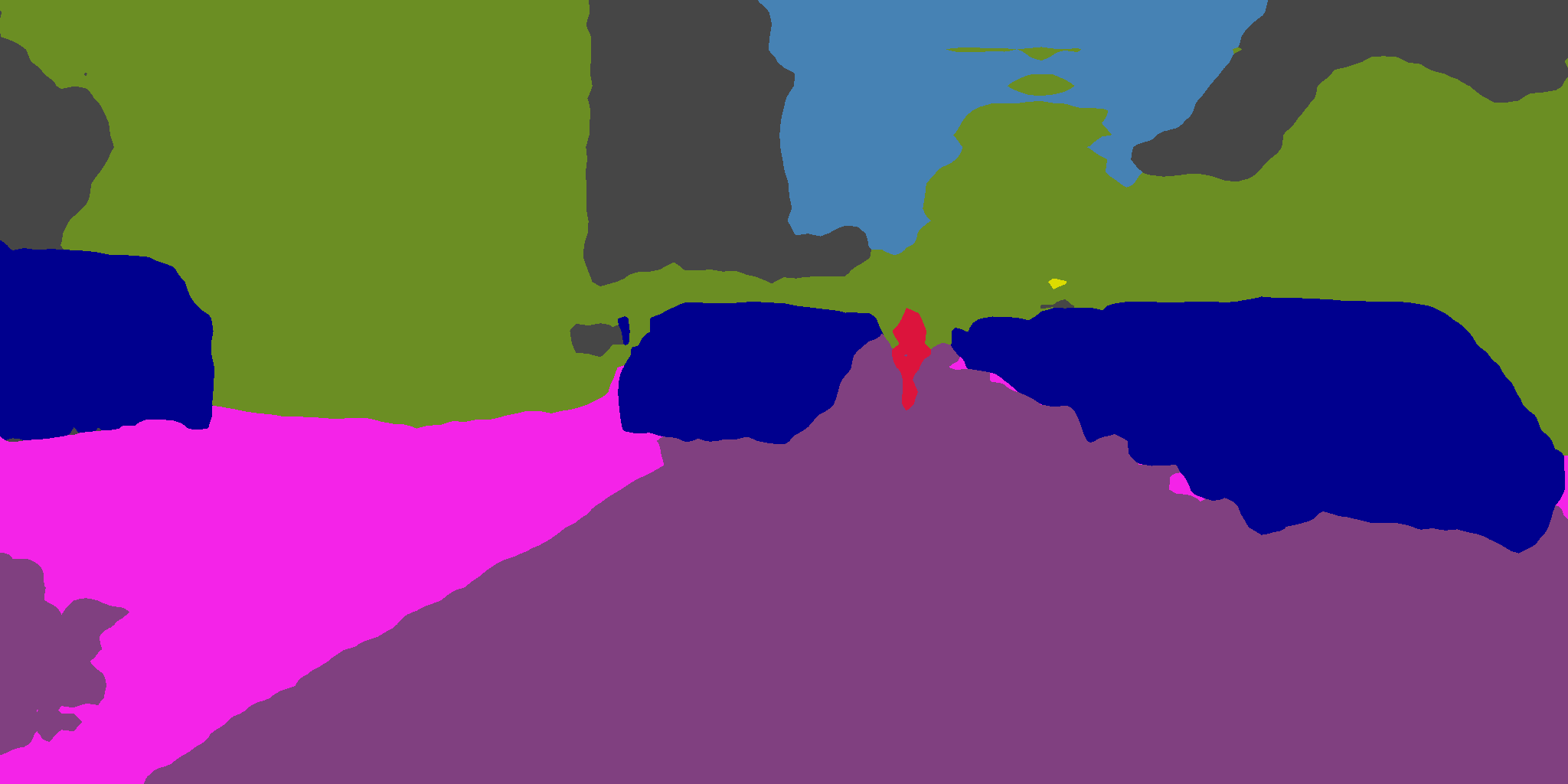}
\end{minipage}
\centering
\caption{Qualitative comparison of TPS with the state-of-the-art over domain adaptive video segmentation benchmark \enquote{SYNTHIA-Seq~$\rightarrow$~Cityscapes-Seq}: TPS produces much more accurate segmentation as compared to ``source only'', indicating the effectiveness of our approach on addressing domain adaptation issue. Moreover, TPS generates better segmentation than DA-VSN~\cite{guan2021domain} and PixMatch~\cite{melas2021pixmatch} as shown in rows 4-5, which is consistent with our quantitative result. Best viewed in color.}
\label{fig:appendix_synthia}
\end{figure}

\vspace{10pt}
\noindent \textbf{C. More Qualitative Comparisons}
\vspace{10pt}

We qualitatively compare the proposed TPS with two best-performing baselines \textit{DA-VSN}~\cite{guan2021domain} and \textit{Pixmatch}~\cite{melas2021pixmatch} over two domain adaptive video segmentation benchmarks. Figs. 1 and 2 show the comparisons, where three consecutive video frames are shown in each figure. It can be observed that the proposed TPS outperforms both DA-VSN and PixMatch clearly and consistently. 

For further evaluation, we compare our method with the state-of-the-arts on real-scene long video sequence from Cityscapes. Instead of directly using test data that only contains short sequences (30 consecutive frames), we evaluate our method on the Cityscapes video demo that lasts much longer (hundreds of frames each sequence, 3 sequences in total).\footnote{https://www.cityscapes-dataset.com/file-handling/?packageID=12/} We pick one sequence for each benchmark and make further comparisons on both benchmarks (i.e. SYNTHIA-Seq$\rightarrow$Cityscapes-Seq and VIPER$\rightarrow$Cityscapes-Seq). The complete record is provided in \url{https://github.com/xing0047/TPS/releases/tag/demo}.

\begin{figure}[!ht]
\vspace{50pt}
\centering
\begin{minipage}[c]{0.03\linewidth}
\centering {\rotatebox{90}{Frames}} 
\end{minipage}
\vspace{2pt}
\begin{minipage}[c]{0.31\linewidth}
\centering\includegraphics[width=.99\linewidth]{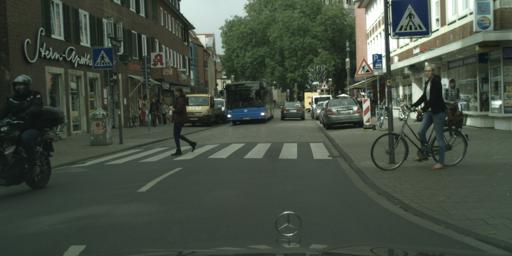}
\end{minipage}
\begin{minipage}[c]{0.31\linewidth}
\centering\includegraphics[width=.99\linewidth]{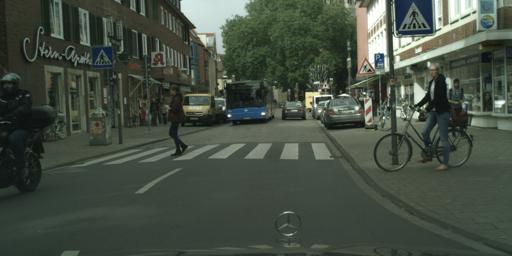}
\end{minipage}
\begin{minipage}[c]{0.31\linewidth}
\centering\includegraphics[width=.99\linewidth]{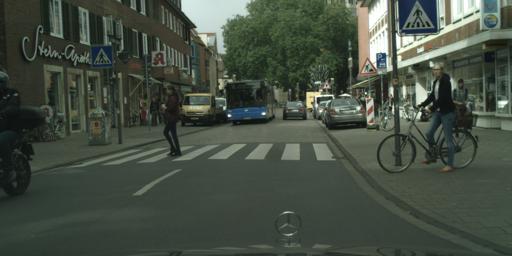}
\end{minipage}
\vspace{2pt}
\centering
\begin{minipage}[c]{0.03\linewidth}
\centering {\rotatebox{90}{GT}} 
\end{minipage}
\begin{minipage}[c]{0.31\linewidth}
\centering\includegraphics[width=.99\linewidth]{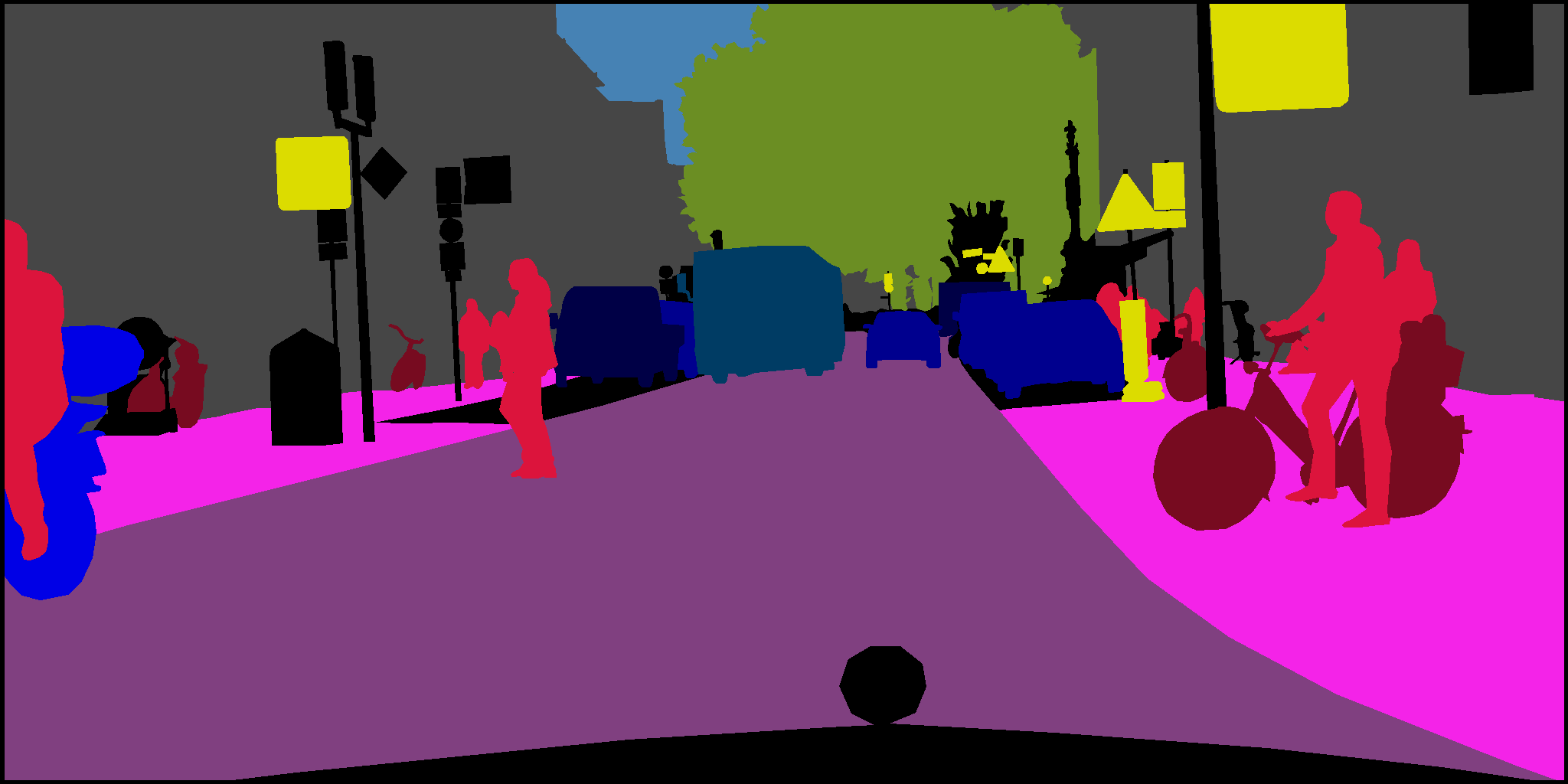}
\end{minipage}
\begin{minipage}[c]{0.31\linewidth}
\centering\includegraphics[width=.99\linewidth]{images/results_viper/munster_000040_000019/munster_000040_000019_gt.png}
\end{minipage}
\begin{minipage}[c]{0.31\linewidth}
\centering\includegraphics[width=.99\linewidth]{images/results_viper/munster_000040_000019/munster_000040_000019_gt.png}
\end{minipage}
\vspace{2pt}
\centering
\begin{minipage}[c]{0.03\linewidth}
\centering {\rotatebox{90}{Source Only}}
\end{minipage}
\begin{minipage}[c]{0.31\linewidth}
\centering\includegraphics[width=.99\linewidth]{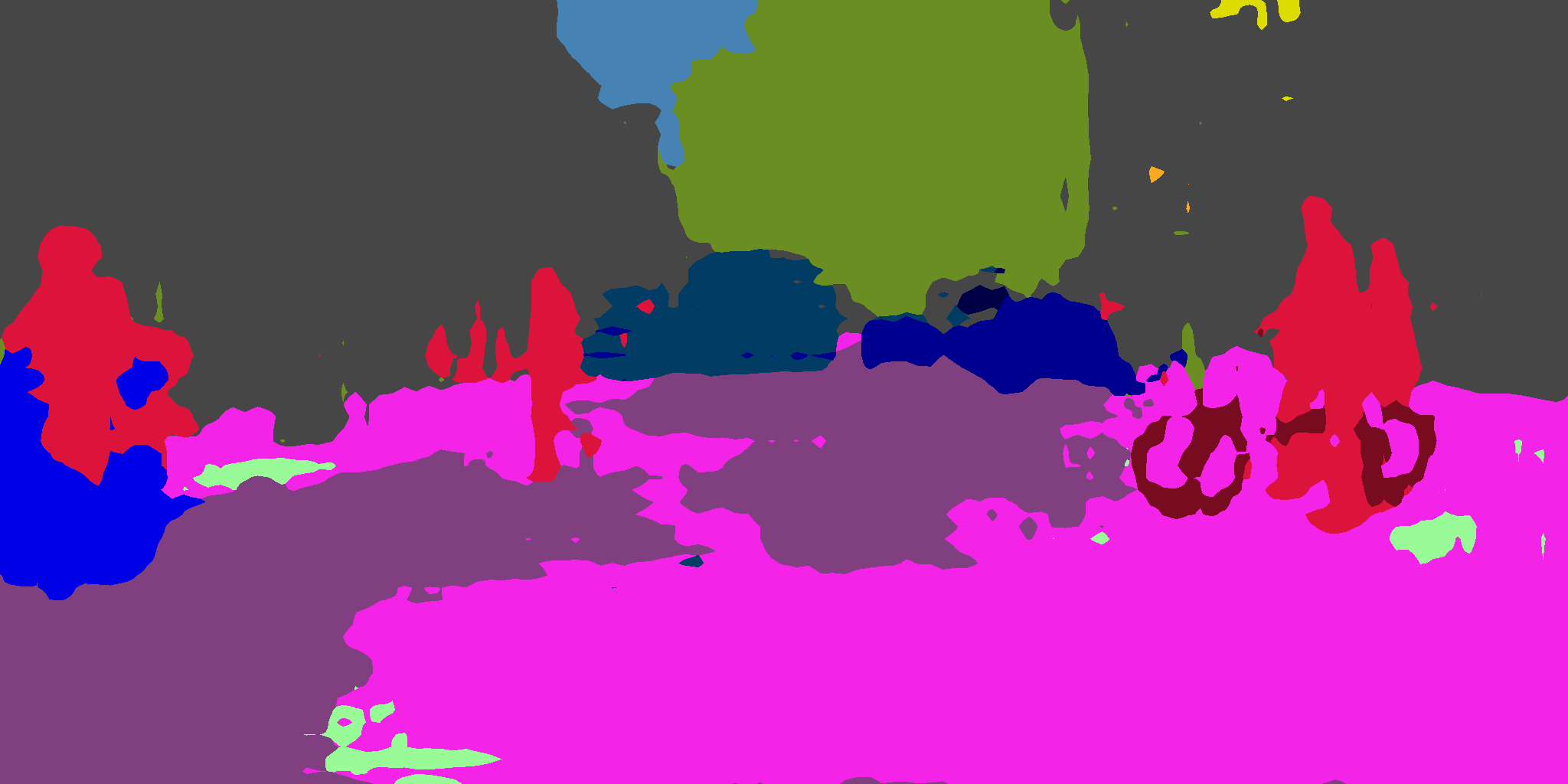}
\end{minipage}
\begin{minipage}[c]{0.31\linewidth}
\centering\includegraphics[width=.99\linewidth]{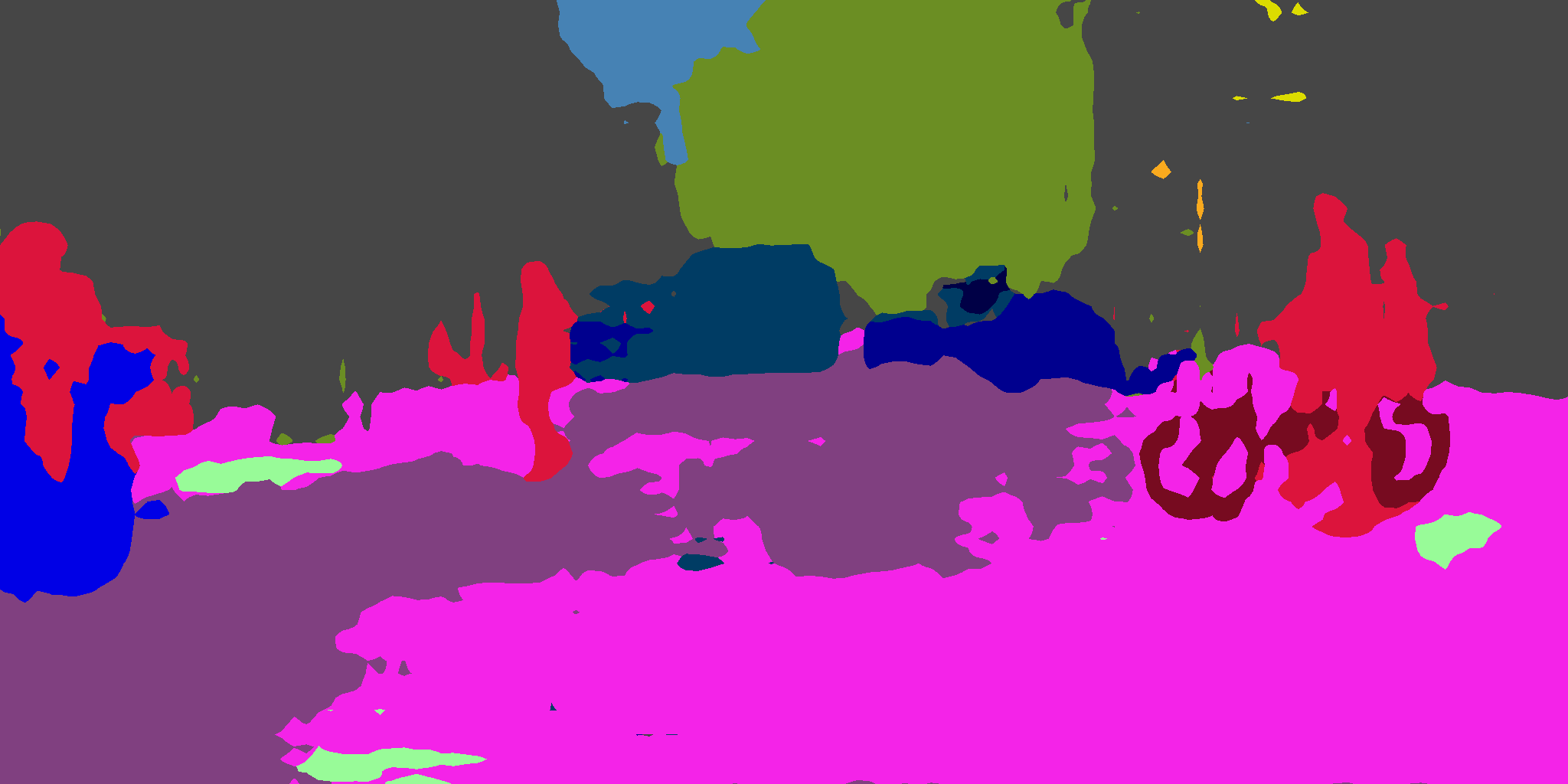}
\end{minipage}
\begin{minipage}[c]{0.31\linewidth}
\centering\includegraphics[width=.99\linewidth]{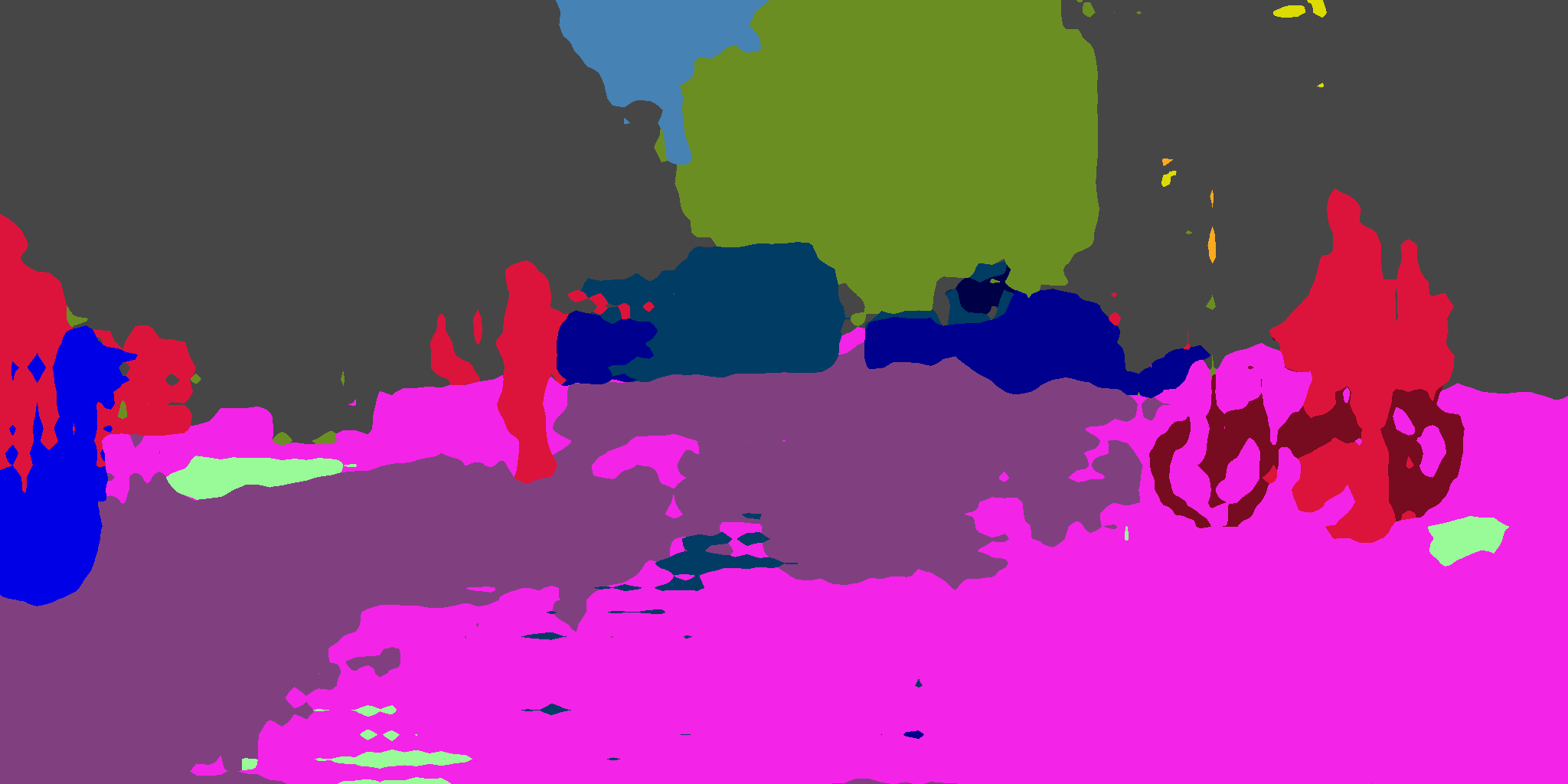}
\end{minipage}
\vspace{2pt}
\centering
\begin{minipage}[c]{0.03\linewidth}
\centering {\rotatebox{90}{DA-VSN}} 
\end{minipage}
\begin{minipage}[c]{0.31\linewidth}
\centering\includegraphics[width=.99\linewidth]{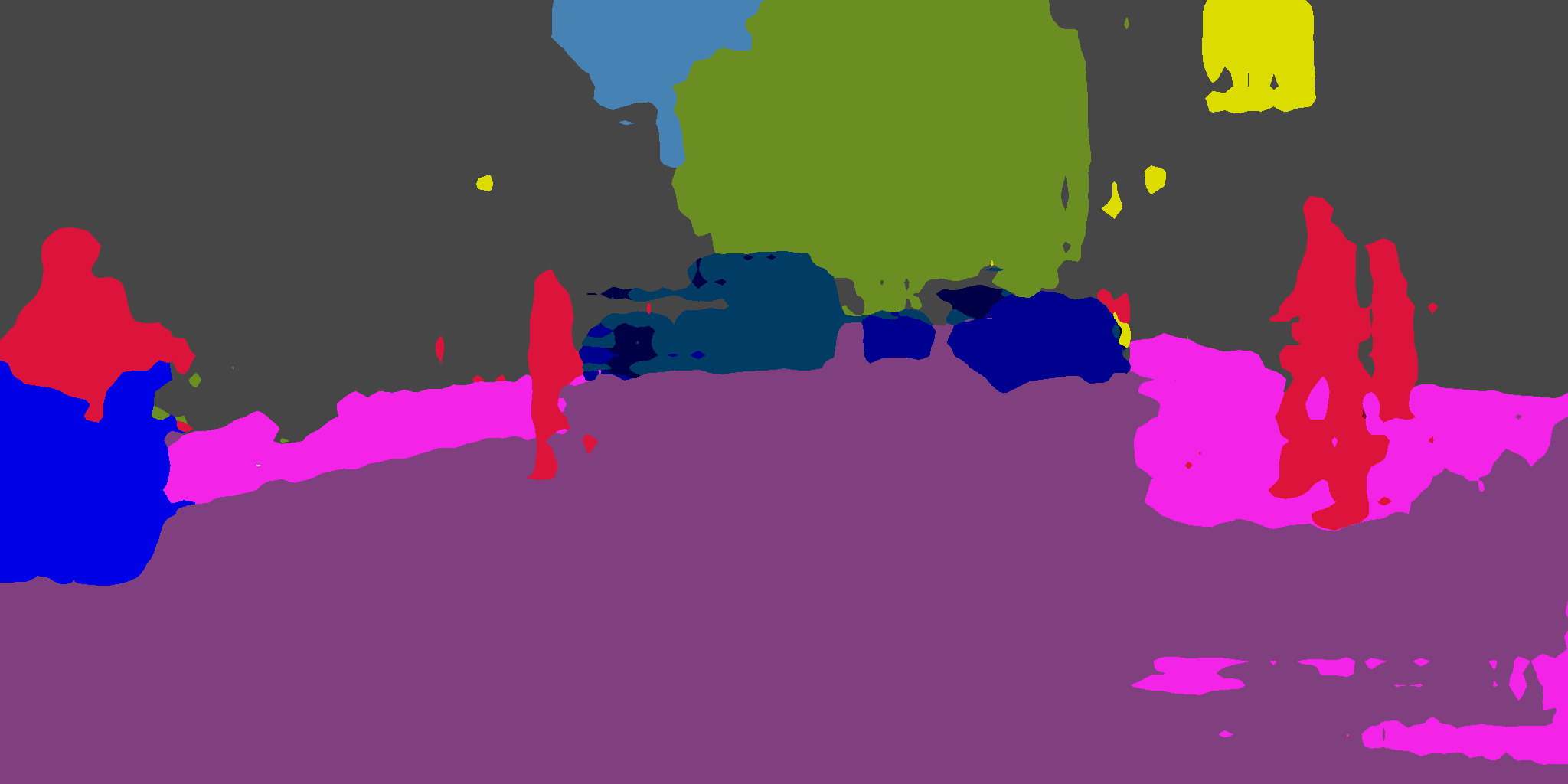}
\end{minipage}
\begin{minipage}[c]{0.31\linewidth}
\centering\includegraphics[width=.99\linewidth]{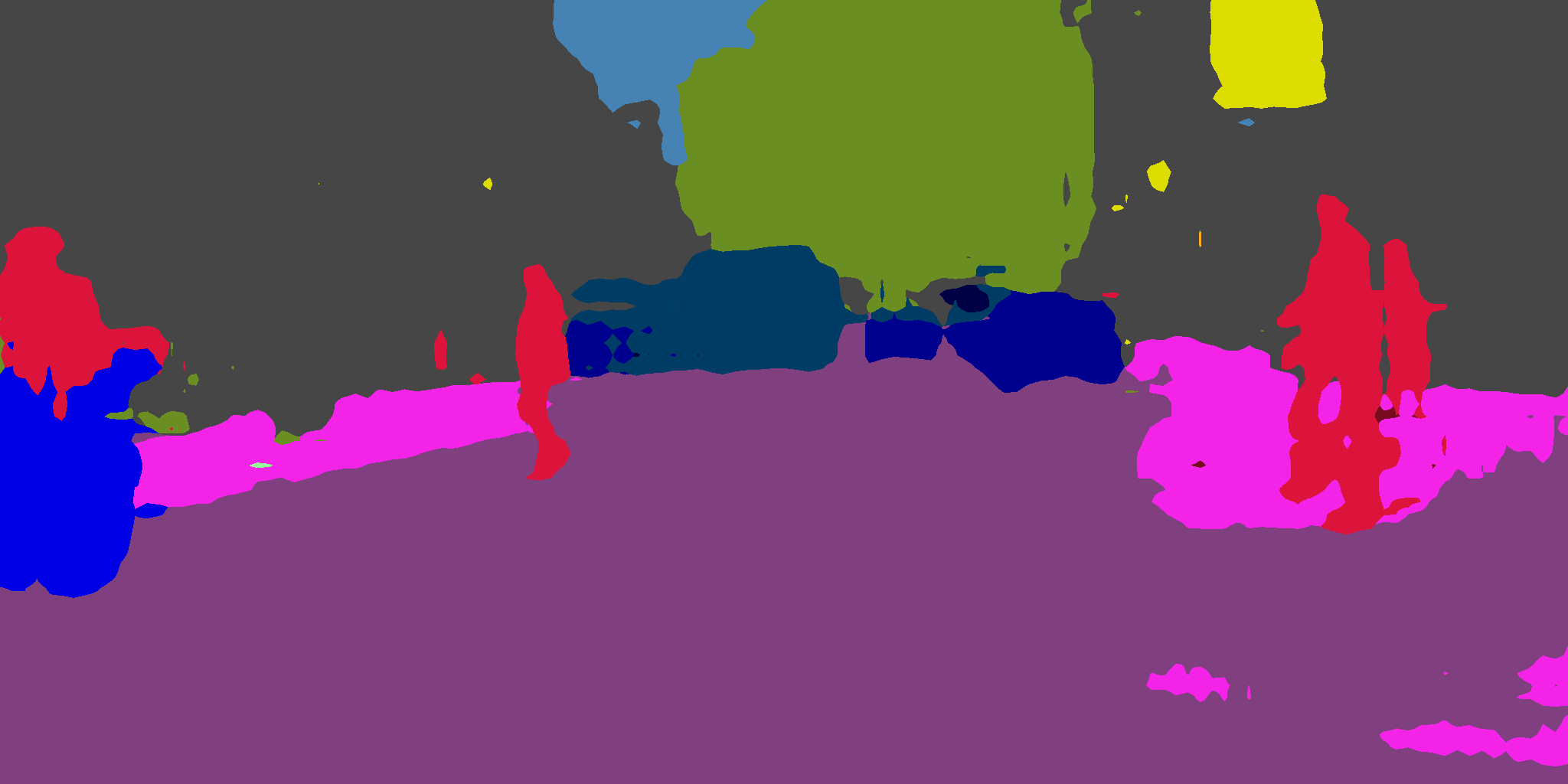}
\end{minipage}
\begin{minipage}[c]{0.31\linewidth}
\centering\includegraphics[width=.99\linewidth]{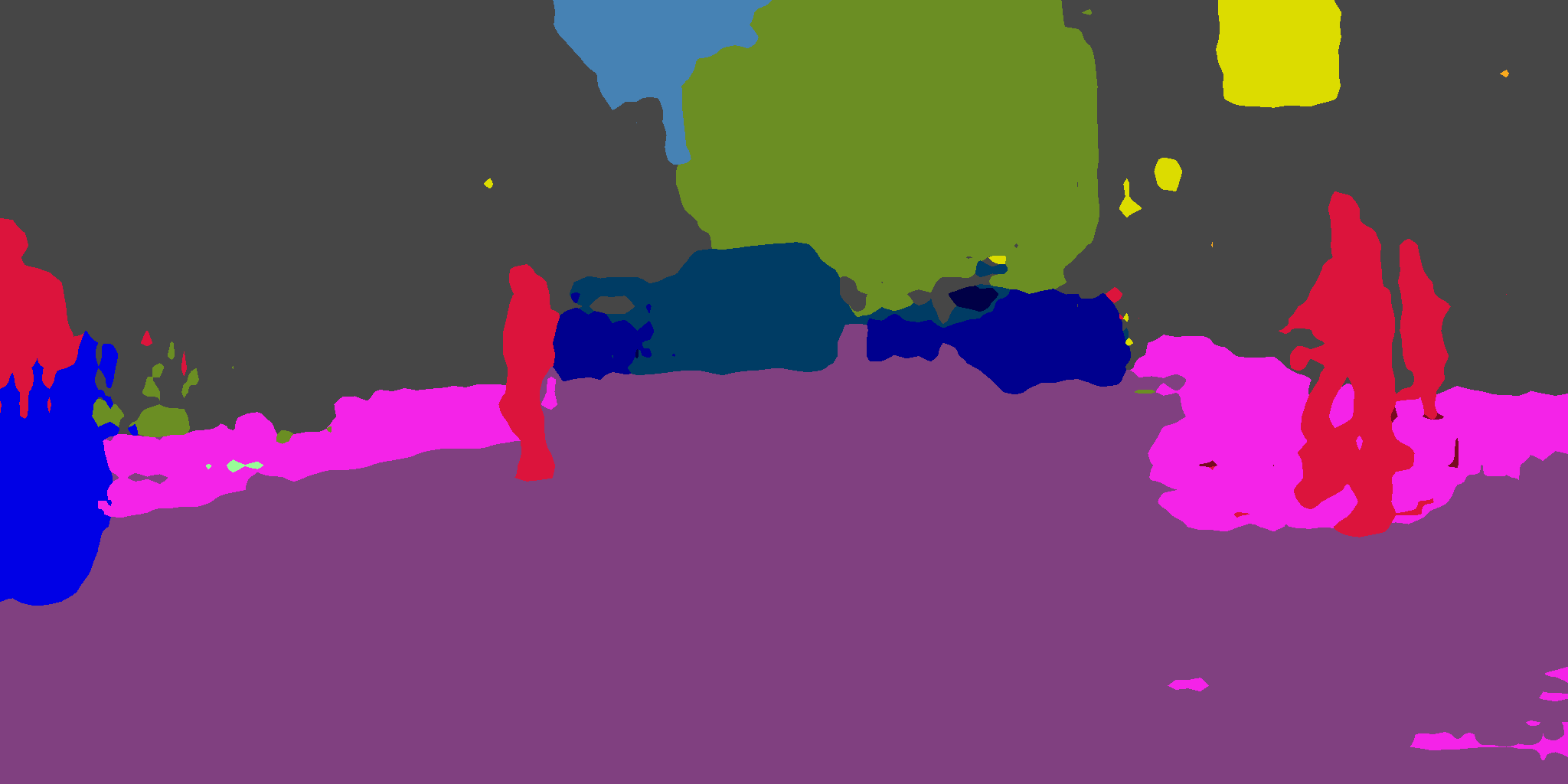}
\end{minipage}
\vspace{2pt}
\centering
\begin{minipage}[c]{0.03\linewidth}
\centering {\rotatebox{90}{PixMatch}} 
\end{minipage}
\begin{minipage}[c]{0.31\linewidth}
\centering\includegraphics[width=.99\linewidth]{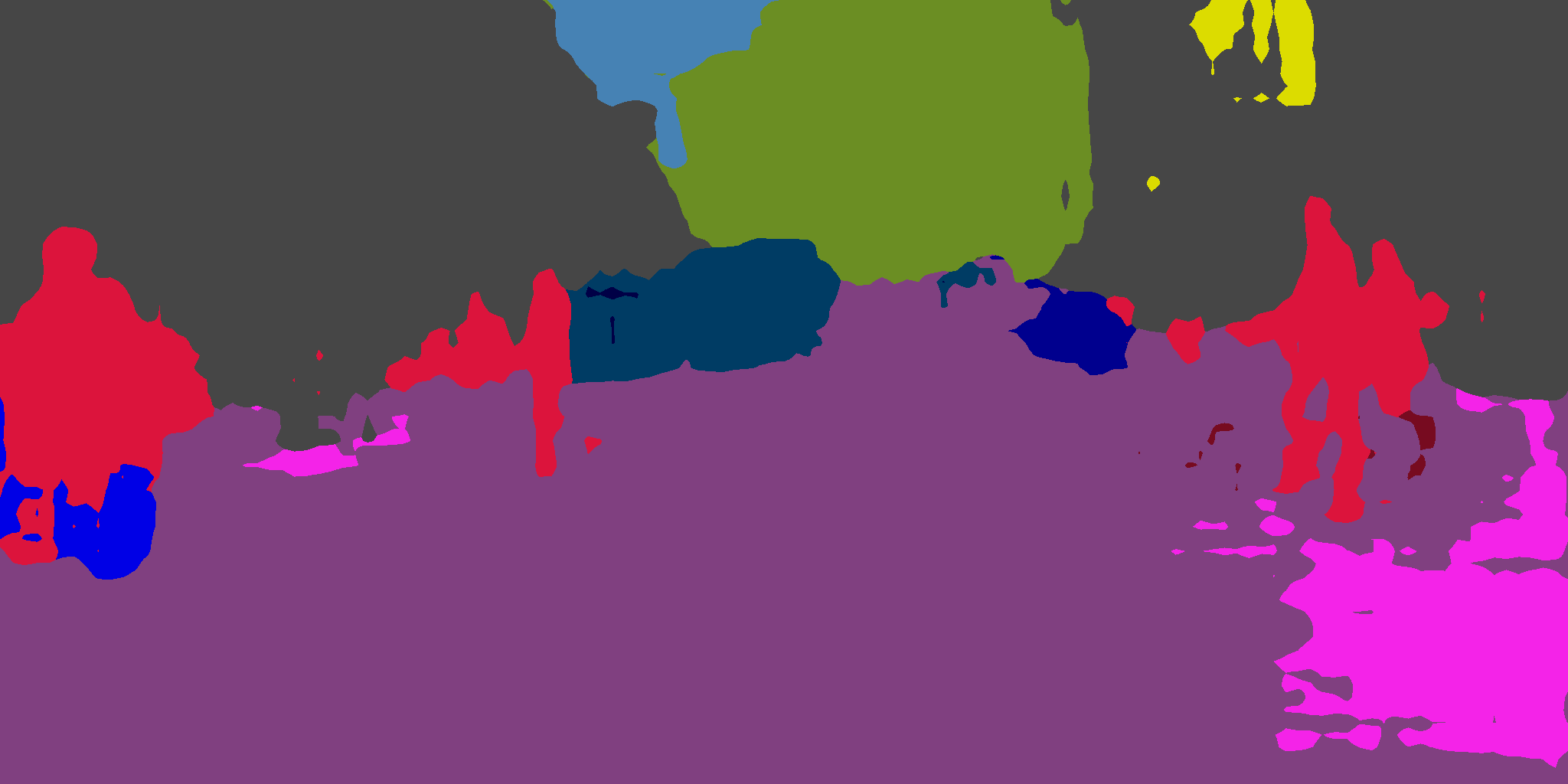}
\end{minipage}
\begin{minipage}[c]{0.31\linewidth}
\centering\includegraphics[width=.99\linewidth]{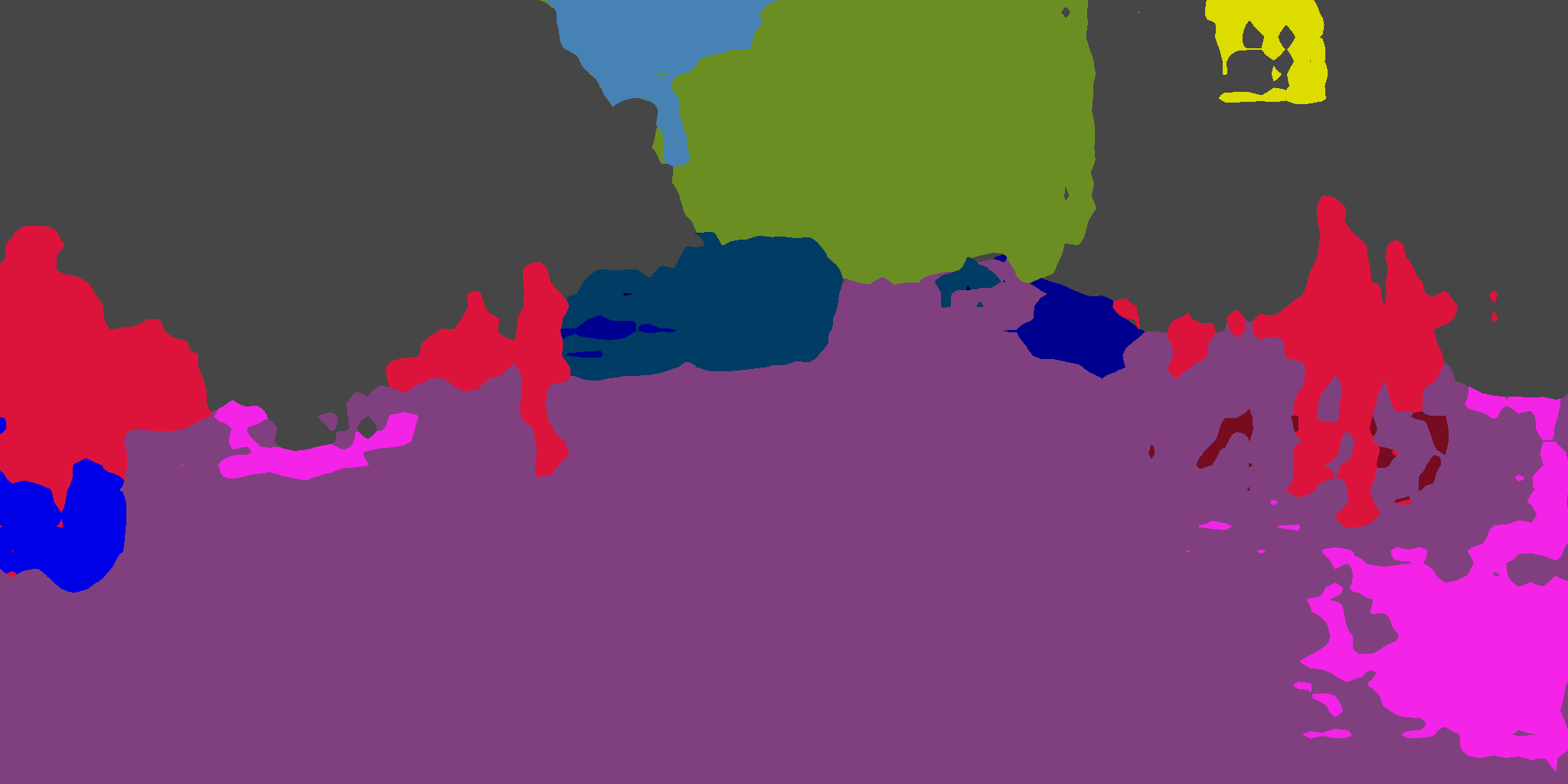}
\end{minipage}
\begin{minipage}[c]{0.31\linewidth}
\centering\includegraphics[width=.99\linewidth]{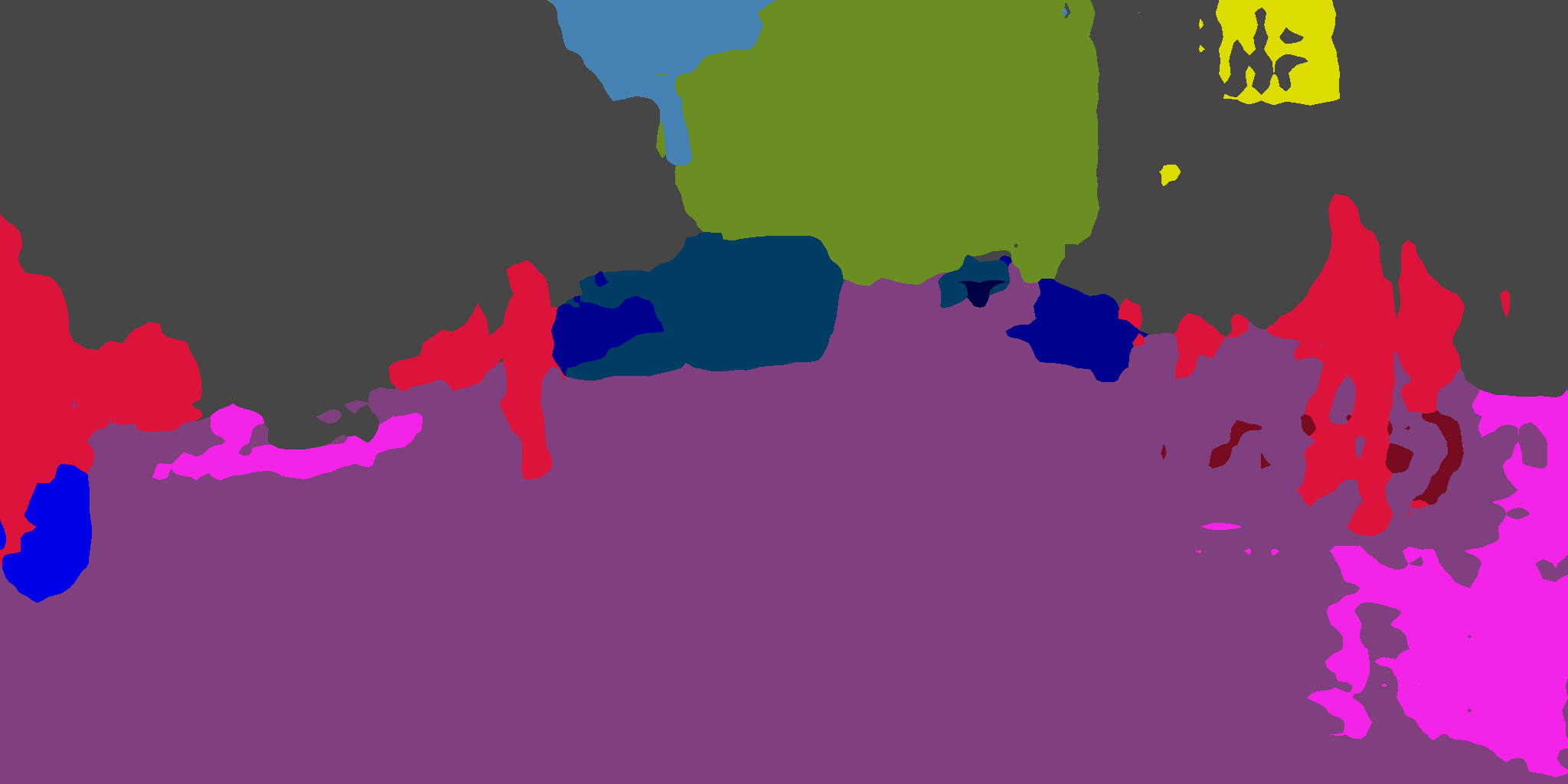}
\end{minipage}
\vspace{2pt}
\centering
\begin{minipage}[c]{0.03\linewidth}
\centering {\rotatebox{90}{\textbf{TPS~(Ours)}}} 
\end{minipage}
\begin{minipage}[c]{0.31\linewidth}
\centering\includegraphics[width=.99\linewidth]{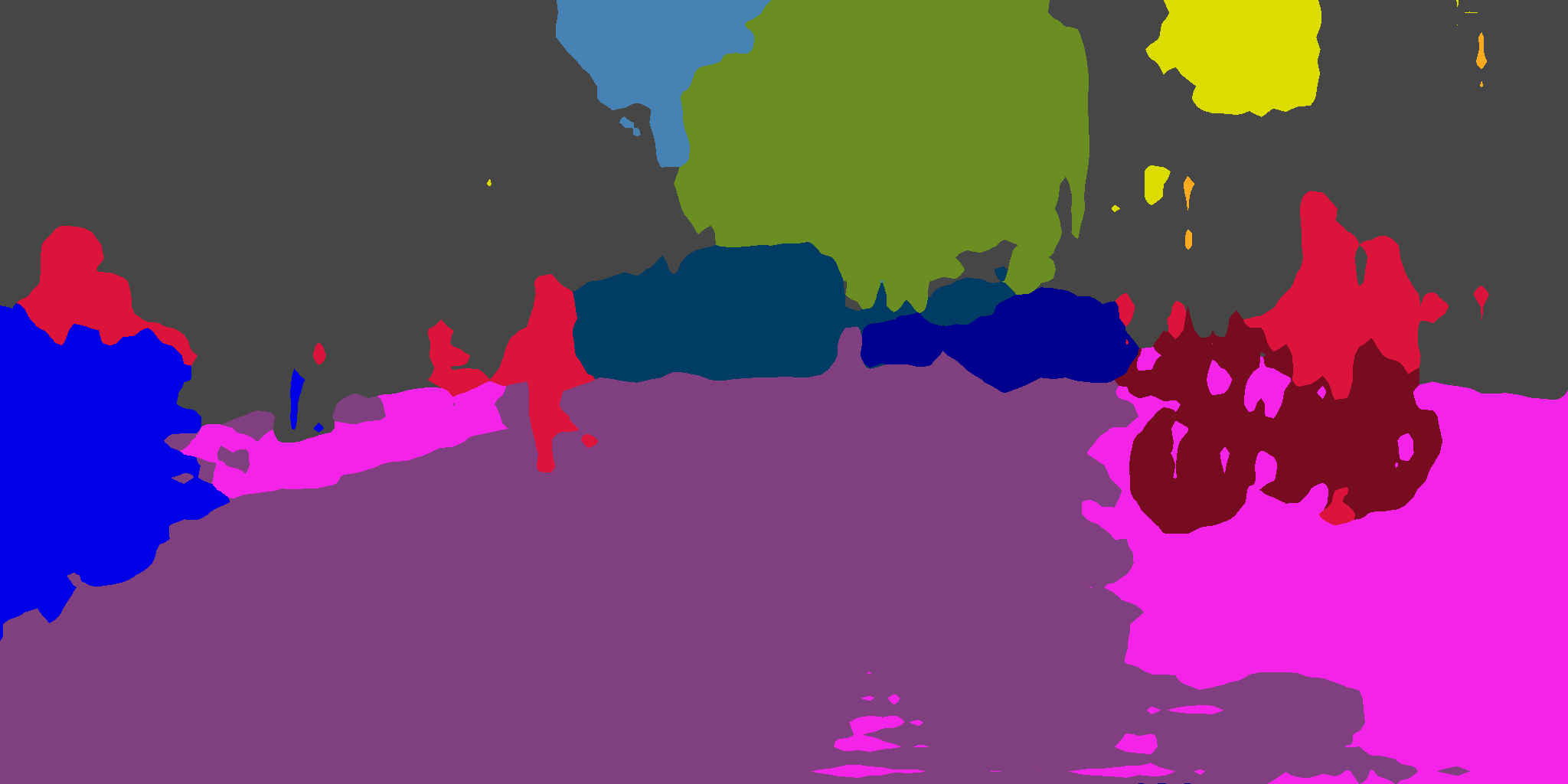}
\end{minipage}
\begin{minipage}[c]{0.31\linewidth}
\centering\includegraphics[width=.99\linewidth]{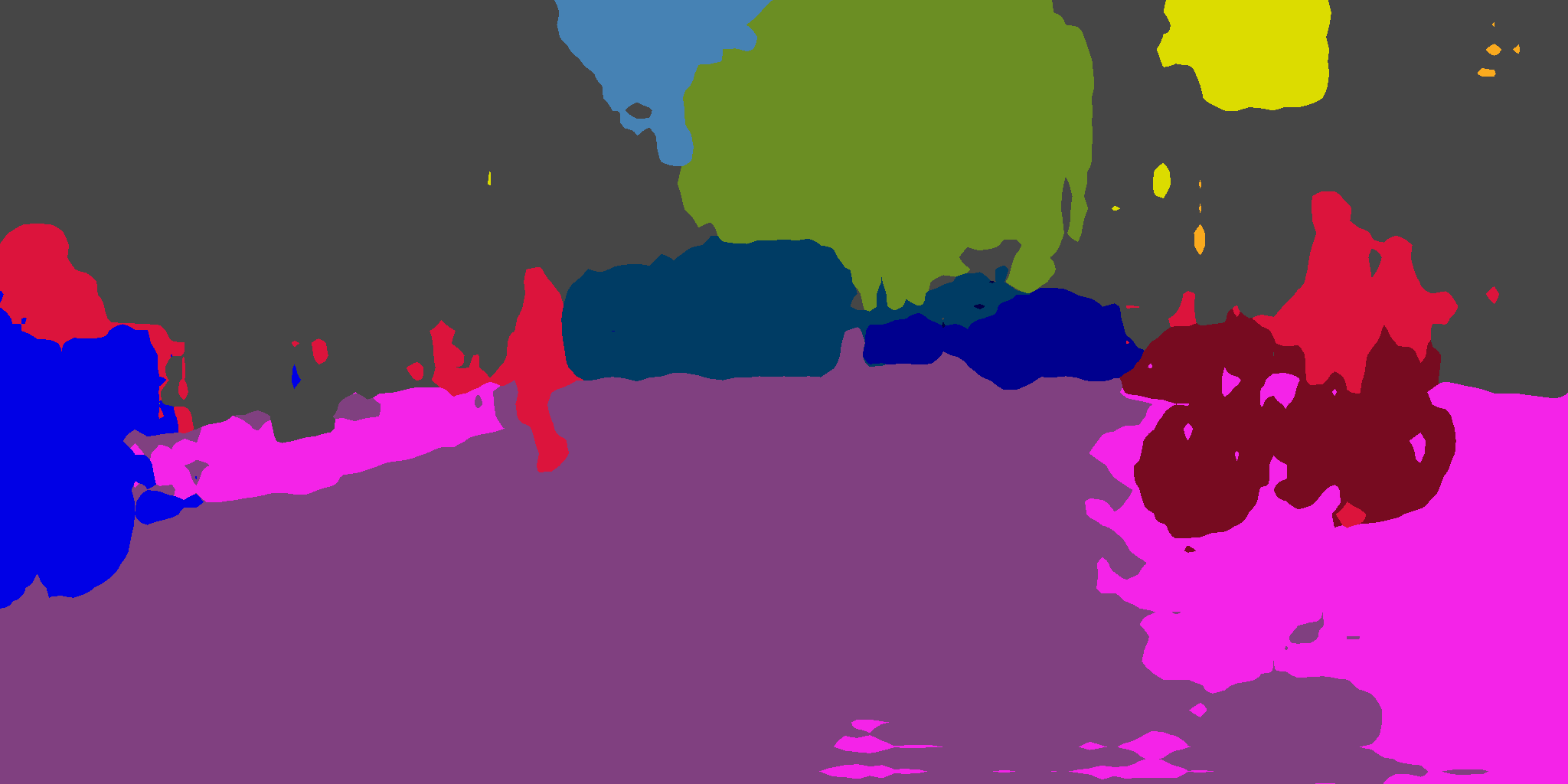}
\end{minipage}
\begin{minipage}[c]{0.31\linewidth}
\centering\includegraphics[width=.99\linewidth]{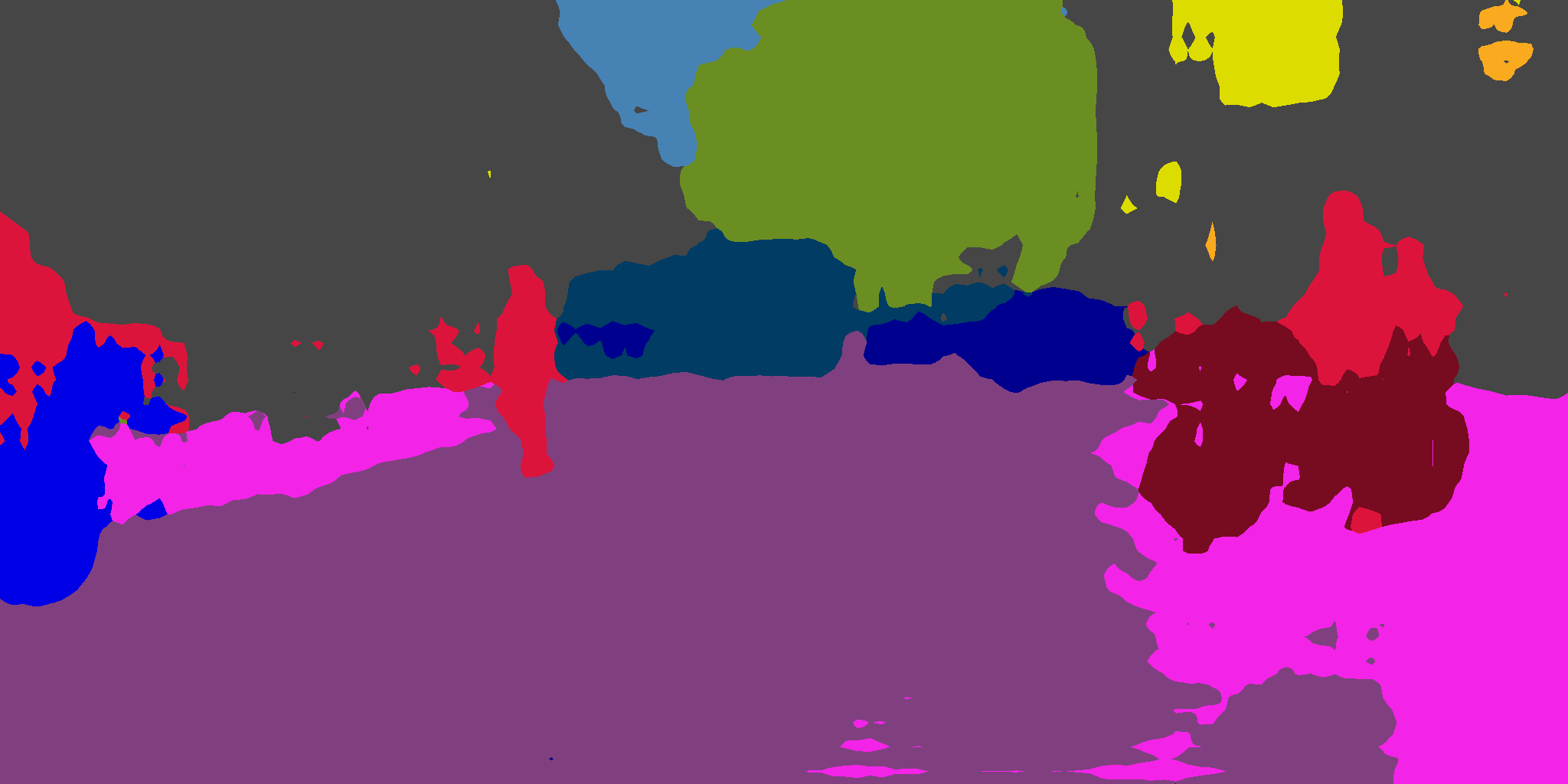}
\end{minipage}
\centering
\caption{Qualitative comparison of TPS with the state-of-the-art over domain adaptive video segmentation benchmark \enquote{VIPER~$\rightarrow$~Cityscapes-Seq}: TPS produces much more accurate segmentation as compared to ``source only'', indicating the effectiveness of our approach on addressing domain adaptation issue. Moreover, TPS generates better segmentation than DA-VSN~\cite{guan2021domain} and PixMatch~\cite{melas2021pixmatch} as shown in rows 4-5, which is consistent with our quantitative result. Best viewed in color.}
\label{fig:appendix_viper}
\end{figure}

\noindent \textbf{D. More Quantitative Comparisons with Consistency-training-based Methods}
\vspace{10pt}

In the Section 4.2, we compared the proposed TPS with the state-of-the-art method on domain adaptive image segmentation using consistency training (the same learning scheme as in this work).
We further reproduce recent consistency-training-based approaches SAC~\cite{araslanov2021self} and DACS~\cite{tranheden2021dacs} for domain adaptive image segmentation task and evaluate on both video adaptive semantic segmentation benchmarks. We note that TPS outperforms all the consistency-training-based methods in
Tabs.~8~and~9, which demonstrates the superiority of our approach.

\renewcommand\arraystretch{1.36}
\begin{table}[!ht]
\label{tab:appendix_synthia}
\centering
\caption{Quantitative comparisons over the benchmark of SYNTHIA-Seq~$\rightarrow$~Cityscapes-Seq: TPS outperforms multiple consistency-training-based domain adaptation methods~\cite{melas2021pixmatch,araslanov2021self,tranheden2021dacs} by large margins. Note that ``Source only'' denotes the network trained with source-domain data solely. Abbreviations for 'sidewalk', 'building', 'vegetation' and 'person' are noted as 'side.', 'buil.', 'vege.' and 'pers.' for simplicity
}
\begin{scriptsize}
\begin{tabular}{p{2cm}|*{11}{p{0.7cm}}|p{0.7cm}}
 \toprule
 \multicolumn{13}{c}{\textbf{SYNTHIA-Seq~$\rightarrow$~Cityscapes-Seq}} \\
 \midrule
 Methods  &{road} &{side.} &{buil.} &{{pole}} &{{light}} &{{sign}} &{vege.} &{sky} &{{pers.}} &{{rider}} &{{car}} &mIoU  \\
 \midrule
 Source only &56.3 &26.6 &\textbf{75.6} &25.5 &5.7 &15.6 &71.0 &58.5 &41.7 &17.1 &27.9 &38.3 \\
 \midrule
 SAC~\cite{araslanov2021self} &87.0& 41.1& 64.0& 20.4& 12.1& 32.8& 38.2& 47.6& 53.1& 19.3& 81.1 &48.9 \\
 DACS~\cite{tranheden2021dacs} &86.4& 40.0& 74.0& \textbf{27.8}& 9.5& 28.2& \textbf{71.6}& \textbf{72.0}& 55.6& 20.0& 76.4&51.0 \\
 PixMatch~\cite{melas2021pixmatch} &90.2& 49.9& 75.1& 23.1& 17.4& 34.2& 67.1& 49.9& 55.8& 14.0& 84.3&51.0 \\
 \textbf{TPS (Ours)} &\textbf{91.2}& \textbf{53.7}& 74.9& 24.6& \textbf{17.9}& \textbf{39.3}& 68.1& 59.7& \textbf{57.2}& \textbf{20.3}& \textbf{84.5}& \textbf{53.8} \\
\bottomrule
\end{tabular}
\end{scriptsize}
\end{table}

\renewcommand\arraystretch{1.36}
\begin{table}[!ht]
\label{tab:appendix_viper}
\centering
\caption{Quantitative comparisons over the benchmark of VIPER~$\rightarrow$~Cityscapes-Seq: TPS outperforms multiple consistency-training-based domain adaptation methods~\cite{melas2021pixmatch,araslanov2021self,tranheden2021dacs} by large margins. Abbreviations for 'sidewalk', 'building', 'vegetation', 'terrain', 'person' and 'motor' are noted as 'side.', 'buil.', 'vege.', 'terr.', 'pers.' and 'mot.' correspondingly}
\begin{scriptsize}
\begin{tabular}{p{1.8cm}|*{15}{p{0.54cm}}|p{0.6cm}}
 \toprule
 \multicolumn{17}{c}{\textbf{VIPER~$\rightarrow$~Cityscapes-Seq}} \\
 \midrule
 Methods  &{road} &{side.} &{buil.} &{fence} &{light} &{sign} &{vege.} &{terr.} &{sky} &{pers.} &{car} &{truck} &{bus} &{mot.} &{bike} &mIoU \\
 \midrule
 Source only &56.7 &18.7 &78.7 &6.0 &22.0 &15.6 &81.6 &18.3 &80.4 &59.9 &66.3 &4.5 &16.8 &20.4 &10.3 &37.1 \\
  \midrule
 DACS~\cite{tranheden2021dacs} &69.6& 24.1& 76.9& 9.1& 16.1& 15.3& 74.1& 20.3& 76.5& 59.4& 74.8& 38.6& 43.1& 7.7& 1.9& 40.5\\
 SAC~\cite{araslanov2021self} &52.2& 19.6& 73.4& 3.7& 23.1& 25.2& 73.9& 17.3& 78.1& 56.9& \textbf{80.3}& 38.3& \textbf{48.2}& 17.8& 14.1 &41.5 \\
 PixMatch~\cite{melas2021pixmatch} &79.4& 26.1& \textbf{84.6}& \textbf{16.6}& \textbf{28.7}& 23.0& \textbf{85.0}& \textbf{30.1}& \textbf{83.7}& 58.6& 75.8& 34.2& 45.7& 16.6& 12.4 &46.7 \\
 \textbf{TPS (Ours)} & \textbf{82.4}& \textbf{36.9}& 79.5& 9.0& 26.3& \textbf{29.4}& 78.5& 28.2& 81.8& \textbf{61.2}& 80.2& \textbf{39.8}& 40.3& \textbf{28.5}& \textbf{31.7}& \textbf{48.9} \\
\bottomrule
\end{tabular}
\end{scriptsize}
\end{table}

\end{document}